\RequirePackage{fix-cm}
\documentclass[twocolumn,twoside]{svjour3} 
\newcommand{\makeheadbox}{}
\usepackage[final]{changes}
\usepackage[switch]{lineno}
\usepackage{hyperref}
\usepackage{graphicx}
\usepackage{tabularx}
\usepackage{comment}
\usepackage{amsmath,amssymb,amsfonts}
\usepackage[x11names, table]{xcolor}
\usepackage[dvipsnames]{xcolor}
\usepackage{mathtools}
\usepackage{pifont}
\usepackage{color}
\usepackage{multirow}
\usepackage{makecell}
\usepackage{tikz}
\usepackage{fbox}
\usepackage{booktabs}
\usepackage{wrapfig}
\usepackage{enumerate}
\usepackage{enumitem}
\usepackage{rotating}
\usepackage{diagbox}
\usepackage{cite}
\usepackage{marginnote}
\usepackage{microtype}
\usepackage[linesnumbered,ruled,vlined]{algorithm2e}
\usepackage[misc]{ifsym}
\usepackage{circledsteps}
\usepackage{longtable}
\usepackage[normalem]{ulem}
\usepackage{array}
\usepackage{siunitx}
\usepackage{arydshln}
\usepackage{tcolorbox}  

\definecolor{userbg}{HTML}{F7FAFE}
\definecolor{sysbg}{HTML}{FDF3E8}

\newtcolorbox{userbox}{colback=userbg, colframe=userbg!85!black, arc=1.5mm, boxrule=0.5pt, left=2mm, right=2mm, top=1.5mm, bottom=1.5mm, fontupper=\small}
\newtcolorbox{sysbox}{colback=sysbg, colframe=sysbg!85!black, arc=1.5mm, boxrule=0.5pt, left=2mm, right=2mm, top=1.5mm, bottom=1.5mm, fontupper=\small}

\newcommand{\blackcircled}[1]{%
\tikz[baseline=(char.base)]{
  \node[shape=circle, fill=black, inner sep=1pt] (char)
  {\textcolor{white}{#1}};
}}

\graphicspath{{fig/}}

\newcommand{\cmark}{\ding{51}}%
\newcommand{\xmark}{\ding{55}}%

\renewcommand{\arraystretch}{1.2}
\definecolor{RoyalBlue}{RGB}{0, 0, 160}
\definecolor{OliveGreen}{RGB}{0, 128, 0}
\definecolor{CustomOrange}{RGB}{255, 128, 0}

\setaddedmarkup{\textcolor{blue}{#1}}
\setdeletedmarkup{\textcolor{red}{\sout{#1}}}

\newlist{tabitemize}{itemize}{1}
\setlist[tabitemize]{
    label=-, 
    leftmargin=*, 
    nosep, 
    before=\vspace{0.25\baselineskip}\scriptsize\raggedright,
}

\usepackage[dvipsnames]{xcolor}
\definecolor{customblue}{rgb}{0.21,0.49,0.74}
\hypersetup{
    colorlinks=true,
    linkcolor=red,
    citecolor=customblue,
    urlcolor=RubineRed
}

\journalname{International Journal of Computer Vision}

\begin{document}

\title{\raisebox{-0.28\height}{
\hspace{-0.4cm}
  \includegraphics[
    height=1.3em,
    trim=0 0 0 0,
    clip
  ]{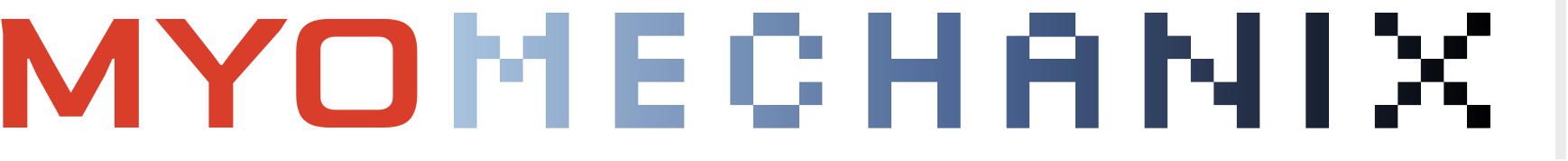}
}
\hspace{-0.2cm}Biomechanically-Grounded Compositional Skilled Activity Understanding and Coaching
}

\author{\rm Hao Yin\textsuperscript{\rm 1, \rm 2*}\thanks{*These authors contributed equally to this work.}\!\and
        \!Paritosh Parmar\textsuperscript{\rm 3*}\!\\
        \!Lijun Gu\textsuperscript{\rm 1, \rm 2}\! \and
        \!Lin Xu\textsuperscript{\rm4}\!\and
        \!Tianxiao Guo\textsuperscript{\rm5}\!\and
        \!Xiujin Liu\textsuperscript{\rm6}\! \and
        \!Tianyou Zheng\textsuperscript{\rm 2}\!\\ 
        \!Yang Zhang\textsuperscript{\rm 2 \textrm{\Letter}}\!\and 
        \!Weiwei Fu\textsuperscript{\rm 1, \rm 2 \textrm{\Letter}}\!
}

\institute{
\textrm{\Letter} Weiwei Fu \at \email{fuww@sibet.ac.cn}\\
\textrm{\Letter} Yang Zhang \at \email{zhangyang@sibet.ac.cn}\\
\textsuperscript{\rm 1} School of Biomedical Engineering (Suzhou), Division of Life Sciences and Medicine, University of Science and Technology of China, Hefei, Anhui, China\\
\textsuperscript{\rm 2} Suzhou Institute of Biomedical Engineering and Technology, Chinese Academy of Sciences, Suzhou, Jiangsu, China\\
\textsuperscript{\rm 3} Arexeni Research and Technologies Inc.\\
\textsuperscript{\rm 4} School of Psychology, Beijing Sports University, Beijing, China\\
\textsuperscript{\rm 5} School of Competitive Sports, Beijing Sports University, Beijing, China\\
\textsuperscript{\rm 6} Department of Robotics, University of Michigan, Ann Arbor, Michigan, USA\\
}

\titlerunning{MyoMechanix Ecosystem}
\authorrunning{Hao Yin and Paritosh Parmar et al.}
\date{}
\maketitle

\vspace{-1cm}

\begin{abstract}
    Existing skilled activity understanding/action quality assessment (AQA) datasets \& methods face two key limitations: they rely primarily on explicit visual inputs (e.g., RGB \& Pose data), \textit{overlooking implicit physiological dynamics such as muscle mechanics}, \& they model actions as monolithic, non-compositional patterns, \textit{limiting semantic richness}. These constraints hinder \textit{fine-grained, biomechanically grounded feedback}. To address this, we pioneer a paradigm that integrates multimodal sensing, \textit{structured} representations, \& \textit{compositional} reasoning. We introduce \textbf{MyoMechanix}, the first of its kind biomechanically grounded multimodal ecosystem for weight-loaded actions with strong internal–external alignment, enabling faithful coupling between motion \& muscle activity. \textit{Expert-annotated}, it comprises 7,500+ samples of 20 actions from 38 subjects, with synchronized multiview RGB video, 3D pose, sEMG, \& additional physiological signals, forming the largest multimodal AQA benchmark to date. Building upon this sensing foundation, we construct the \textbf{Fitness Knowledge Graph} (FKG), which organizes annotations from field experts into structured relationships linking actions, phases, key steps, error types, \& corrective feedback. This representation supports a compositional scoring \& reasoning framework that enables interpretable \& fine-grained quality assessment. To capitalize on the multimodal richness \& structured representations, we develop a new modeling paradigm, \textbf{CUBIST} (Compositional Ontological Reasoning Engine), a framework that performs decomposition–analysis–recomposition over structured actions to enable fine-grained error attribution and feedback generation. We further establish the \textbf{MyoMechanix-AQA} benchmark \& \textbf{MyoMechanix-VideoQA}, supporting tasks ranging from action quality assessment to error diagnosis \& cross-modal reasoning, including a novel \textbf{MyoMechanix-Video2EMG} task. Three key findings emerge from extensive experiments: 1) multimodal sensing with structured representations improves performance, interpretability, \& error attribution, with CUBIST achieving state-of-the-art results; 2) VideoQA enhances language-grounded, fine-grained action understanding in vision–language models; 3) Video2EMG results suggest cheap video-based alternatives to expensive EMG sensors. Nonetheless, MyoMechanix remains challenging, motivating further research. This work advances skilled activity understanding from visual recognition to biomechanically grounded, multimodal, \& compositional reasoning, providing a foundation for next-generation Physical AI with applications in fitness, rehabilitation, healthcare, \& core machine learning research, particularly, representation learning.\\
    Project page: \href{https://haoyin116.github.io/MyoMechanix/}{MyoMechanix Ecosystem}.

\end{abstract}

\begin{figure*}[ht]
    \centering
    \includegraphics[width=\linewidth]{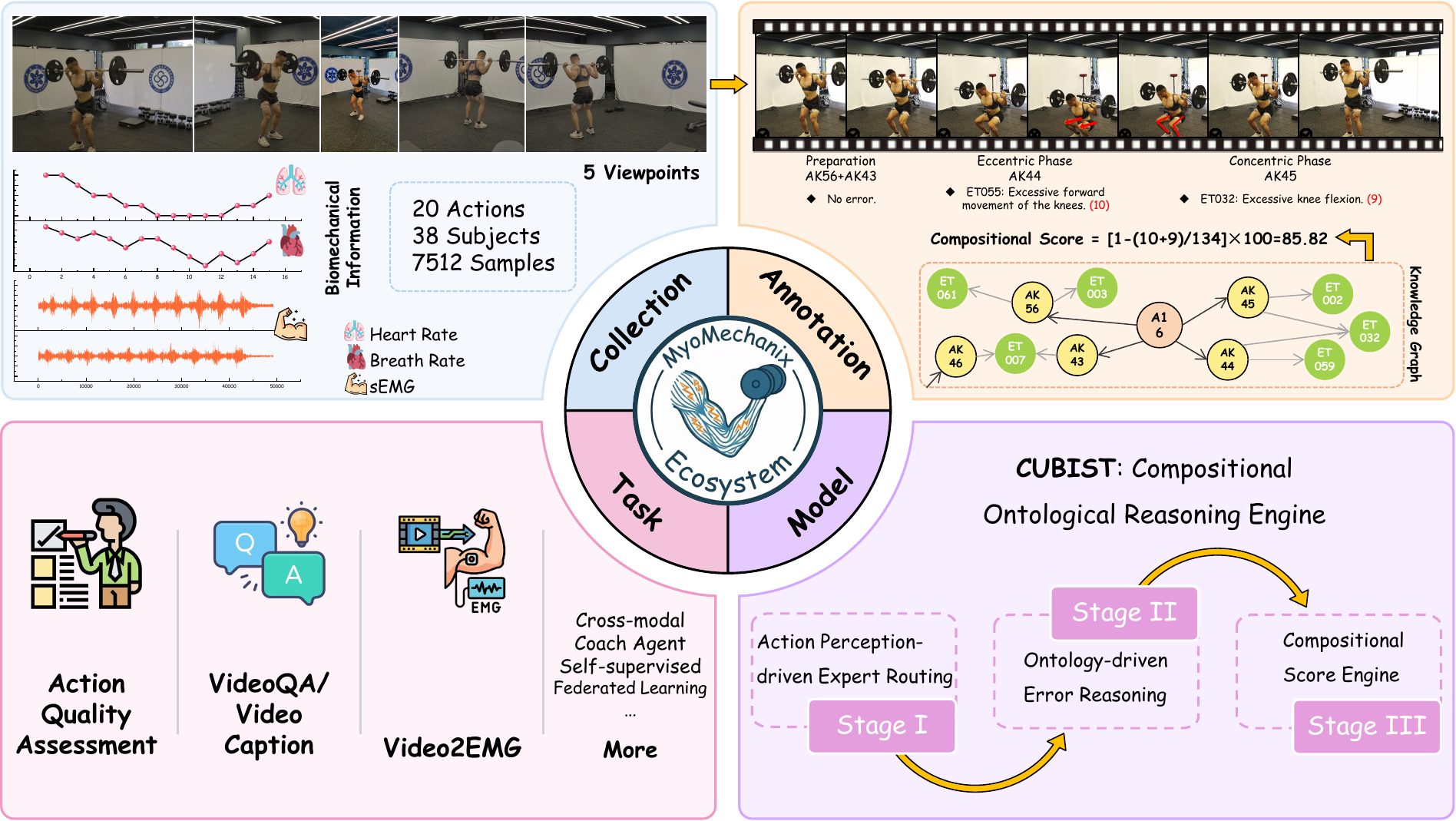}
    \caption{\textbf{An overview of our MyoMechanix ecosystem.} 
    We introduce the MyoMechanix ecosystem, the first  biomechanically grounded multimodal framework for skilled activity understanding and coaching. MyoMechanix advances fitness activity understanding along two complementary dimensions: physical fidelity and the semantic richness of actions. For data collection, we use high-precision motion capture to record 7,512 samples from 38 subjects spanning three ability levels and performing 20 fitness actions, making it the largest multimodal fitness AQA benchmark. The dataset includes synchronized recordings from five RGB camera views, 3D poses, sEMG, and physiological signals, and is annotated by multiple fitness experts. To support structured reasoning, MyoMechanix incorporates a fitness knowledge graph (FKG) that formalizes relationships among action key steps (AK), error types (ET), and action feedback. FKG recasts fitness actions as structured procedures, enabling fine-grained error reasoning, feedback generation, and interpretable scoring. This framework shifts the paradigm from subjective overall scoring to objective, fine-grained error annotation and establishes a computational scoring system. Building on these resources, we develop benchmark datasets for AQA, VideoQA, and Video2EMG, providing a versatile foundation for future multidisciplinary research. Furthermore, we propose CUBIST, a Compositional Ontological Reasoning Engine. Through a three-stage progression from action perception to scoring, CUBIST enables precise quantitative evaluation with high interpretability at both the intermediate reasoning and outcome levels.
    }
    \label{fig_intro_overview}
\end{figure*}

\section{Introduction}
\label{Sec1-Intro}

Last spring, Christina, a fitness enthusiast attempting a personal-best barbell squat, appeared to exhibit correct form when assessed visually—both through self-observation and by an AI-based coaching system. However, subtle misalignments in load distribution and improper muscle activation ultimately led to a lower-back injury. Crucially, neither the system nor the athlete was able to identify which specific phase of the movement caused the failure, nor to uncover its underlying physiological origin. This example highlights a fundamental limitation of current action analysis approaches: their reliance on external visual cues, lack of access to internal physiological signals, and inability to decompose actions into interpretable components for fine-grained reasoning.

Human motion is fundamentally governed by the neuromuscular system, in which observable kinematics emerge from complex coordination patterns of muscle activation. Due to the redundancy of the musculoskeletal system, visually similar movements can arise from substantially different internal control strategies—some of which may compromise safety and performance. Consequently, errors in execution often manifest not as obvious deviations in motion trajectories, but as latent inconsistencies in muscle coordination and force generation. Effective assessment, therefore, requires both physiological grounding and structured representations that capture the compositional nature of human actions.

\begin{figure}[t]
    \centering
    \includegraphics[width=\linewidth]{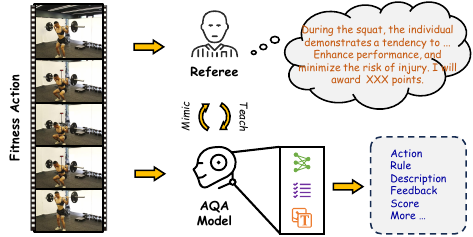}
    \caption{\textbf{Illustration of AQA task.} Given an action video, the AQA model plays the role of a referee to evaluate performance and score skill levels.}
    \label{fig_intro_aqa}
\end{figure}

Existing research on Action Quality Assessment (AQA)\footnote{The task of computational assessment of how well an action is performed has been described using various terms, including Skilled Activity Understanding, AQA, Skills Assessment, \& Proficiency Estimation. Consequently, this study treats these terms as conceptually equivalent \& uses them interchangeably.} has achieved significant progress in modeling spatiotemporal dynamics using convolutional neural networks~\cite{parmar2017learning}, Transformers~\cite{bai2022action}, and graph-based architectures \cite{yan2018spatial,zhou2023hierarchical} (see \autoref{fig_intro_aqa}). However, these approaches remain in several critical aspects. \textit{Firstly}, they primarily concentrate on exteroceptive visual inputs and explicit kinematic features such as joint positions, limb configurations, and temporal patterns \cite{yin2026decade}. As a result, they struggle to capture crucial information regarding implicit physiological states—including muscle activation, fatigue, and coordination strategies~\cite{alcan2023current}. \textit{Secondly}, they tend to treat actions as monolithic, non-compositional patterns rather than structured compositions. In addition, widely used AQA datasets are often derived from web-crawled videos, which suffer from inconsistent quality, limited diversity, and coarse-grained annotations lacking semantic structure and causal interpretability. Collectively, these limitations hinder the development of models capable of fine-grained, physiologically informed reasoning about complex human actions.

To address these challenges, we introduce a new paradigm for action understanding that integrates: \textbf{1)} \textit{sensorimotor integration}, \textbf{2)} \textit{structured representation}, and \textbf{3)} \textit{compositional reasoning}. Concretely, we propose to model human actions as biomechanically grounded, structured processes, in which observable motion is coupled with underlying physiological dynamics and organized into semantically meaningful components. Our main contributions can be summarized as follows.

\textbf{First}, we introduce MyoMechanix, a multimodal data acquisition and analysis ecosystem for weight-loaded exercises. In contrast to prior datasets, MyoMechanix combines synchronized multi-view video (including ubiquitous phone camera), high-precision motion capture, and surface electromyography (sEMG) \cite{fastmove, fastmove_gait}, enabling direct measurement of muscle mechanisms/activity alongside external motion. The inclusion of sEMG provides access to internal neuromuscular signals, offering a crucial complementary perspective to visual observation and enabling a fundamentally more comprehensive characterization of action quality. The resulting dataset comprises over 7,500 samples across 20 exercise types performed by 38 subjects, totaling more than 40 hours of multimodal recordings (see \autoref{fig_intro_overview}).

\textbf{Second}, we propose the Fitness Knowledge Graph (FKG), which formalizes human actions as structured action representations. Specifically, each action is decomposed into hierarchical phases (e.g., preparation, concentric, eccentric) and further subdivided into ordered action key steps (AKs) associated with domain-specific error types (ETs) and corrective feedback (FBs). This representation introduces an explicit semantic structure that captures both the procedural organization of actions and the causal relationships between execution errors and their consequences. Building upon this ontology, we design a compositional, penalty-based annotation protocol that assigns weights to errors according to their biomechanical impact, enabling interpretable and fine-grained evaluation.

\textbf{Third}, we introduce a new modeling paradigm---CUBIST (Compositional Ontological Reasoning Engine), a framework for compositional reasoning over human actions that leverages the multimodal richness and structured representations of FKG. Inspired by the principle of decomposition–analysis–recomposition, CUBIST operates by (i) decomposing actions into structured components, (ii) analyzing each component through multimodal inputs and specialized processing modules, and (iii) recomposing the results to produce coherent assessments, including error attribution and corrective feedback. This design enables the model to move beyond holistic scoring toward interpretable, step-wise reasoning that more closely reflects expert analysis.

Based on multimodal data and structured annotations of this ecosystem, we establish the MyoMechanix-AQA benchmark for structured action quality assessment, as well as MyoMechanix-VideoQA, a dataset supporting tasks ranging from action recognition to fine-grained error diagnosis and feedback generation. Importantly, we introduce Video2EMG, a novel cross-modal task that aims to infer muscle activation patterns from visual input.

Extensive experiments demonstrate that firstly, the integration of multimodal sensing and structured representations significantly improves performance over existing SOTA baselines, while also enabling richer interpretability and more precise error attribution. Our CUBIST achieves SOTA performance, significantly outperforming existing AQA models. Secondly, the VideoQA dataset helps instill deeper action understanding in vision-language models, supporting the development of foundation models for language-grounded, query-conditioned, fine-grained action understanding. Thirdly, we demonstrate promising results on the cross-modal task---Video2EMG, paving the way for video-based alternatives to expensive, specialized EMG sensors. Overall, despite these advances, MyoMechanix remains a challenging benchmark with considerable room for improvement, motivating future research in physiologically grounded and compositional action understanding.

In summary, this work advances action understanding from purely visual pattern recognition toward biomechanically grounded, multimodal, and compositional reasoning. We believe this paradigm provides a foundation for future research in physically grounded AI, with broad implications for applications in fitness training, rehabilitation, and smart healthcare systems.

The remainder of our paper is organized as follows:
\begin{itemize}
    \item Section \ref{Sec2-Related} provides a comprehensive literature review of relevant datasets and research methodologies. 
    \item Section \ref{Sec3-Dataset} details the construction process and protocols of the MyoMechanix ecosystem. 
    \item Section \ref{Sec4-Model} elaborates on the architectural design and algorithms of the proposed CUBIST model. 
    \item Section \ref{Sec5-Exp} presents comprehensive experimental results and in-depth analyses of various baseline models on the proposed benchmarks. 
    \item Finally, Section \ref{Sec6-Con} concludes the paper and discusses potential directions for future research. 
\end{itemize}
\section{Related Work}
\label{Sec2-Related}

We comprehensively compare our work with prior art in related areas, especially action quality assessment datasets and models, video question answering datasets, electromyography datasets, and muscle state estimation.

\subsection{Action Quality Assessment Datasets}
\label{Sec21-AQAD}

\begin{table*}
\centering
\caption{\textbf{Comparison of the MyoMechanix dataset with representative AQA datasets in terms of sample size, action types, views, modalities, annotations, skill level coverage \& sources.} V: Video, Sk: Skeleton, sE: sEMG, P: Physiological Info., S: Score, G: Grade, A: Action, F: Formation, D: Description, E: Error, Fb: Feedback. N: Novice, Am: Amateur, Ex: Expert, Nor: Normal, Abn: Abnormal, Rehab: Rehabilitation.}
\begin{tabular}{@{}>{\raggedright\arraybackslash}p{2.7cm}>{\centering\arraybackslash}p{1cm}>{\centering\arraybackslash}p{0.9cm}>{\centering\arraybackslash}p{0.9cm}>{\raggedright\arraybackslash}p{2cm}>{\raggedright\arraybackslash}p{2.3cm}>{\raggedright\arraybackslash}p{1.8cm}>{\centering\arraybackslash}p{1.2cm}>{\centering\arraybackslash}p{1.2cm}X@{}}
    \toprule[1.5pt]
    \textbf{Dataset} & \textbf{Sample} & \textbf{Type} & \textbf{View} & \textbf{Modality} & \textbf{Annotation} & \textbf{Skill Levels} & \textbf{Domain} & \textbf{Source} \\ \midrule [1.0pt]
    MIT-Dive \cite{pirsiavash2014assessing}      & 159  & 1  & 1 & V            & S         & Ex          & Sport & Web  \\
    MIT-Skate \cite{pirsiavash2014assessing}     & 150  & 1  & 1 & V            & S         & Ex          & Sport & Web  \\
    UNLV-Dive \cite{parmar2017learning}          & 370  & 1  & 1 & V            & S         & Ex          & Sport & Web  \\
    UNLV-Vault \cite{parmar2017learning}         & 176  & 1  & 1 & V            & S         & Ex          & Sport & Web  \\
    AQA-7 \cite{parmar2019action}                & 1189 & 7  & 1 & V            & S, A      & Ex          & Sport & Web  \\
    MTL-AQA \cite{parmar2019and}                 & 1412 & 52  & 1 & V           & S, A, D   & Ex          & Sport & Web  \\
    Waseda-Squat \cite{ogata2019temporal}        & 2001 & 1  & 1 & V, Sk        & E         & N           &Fitness& Camera\\
    Fis-V \cite{xu2019learning}                  & 500  & 1  & 1 & V            & S         & Ex          & Sport & Web  \\
    TASD-2 \cite{gao2020asymmetric}              & 606  & 2  & 1 & V            & S, A      & Ex          & Sport & Web  \\
    Rhy.Gym. \cite{zeng2020hybrid}               & 1000 & 4  & 1 & V            & S, A      & Ex          & Sport & Web  \\
    QMAR \cite{sardari2020vi}                    & 306  & 2  & 6 & V, Sk        & S, A      & Nor, Abn    & Rehab & Camera\\
    FR-FS \cite{wang2021tsa}                     & 417  & 1  & 1 & V            & G, A      & Ex          & Sport & Web  \\
    Fitness-AQA \cite{parmar2022domain}          &13049 & 3  & 1 & V            & G, A      & Am          &Fitness& Web  \\
    FineDiving \cite{xu2022finediving}           & 3000 & 52 & 1 & V            & S, A      & Ex          & Sport & Web  \\
    LOGO \cite{zhang2023logo}                    & 200  & 12 & 1 & V            & S, A, F   & Ex          & Sport & Web  \\
    FineFS \cite{ji2023localization}             & 1167 & 1  & 1 & V            & S, A      & Ex          & Sport & Web  \\
    PaSk \cite{gao2023automatic}                 & 1018 & 1  & 1 & V            & S         & Ex          & Sport & Web  \\
    EgoExo-Fitness \cite{li2024egoexo}           & 6131 & 12 & 6 & V            & S, A, D   & N, Am       &Fitness& Camera  \\
    LucidAction \cite{dong2024lucidaction}       & 6702 & 8  & 8 & V, Sk        & S, A      & Am, Ex      & Sport & MoCap  \\ 
    FitAQA \cite{zheng2026fitaqa}                & 5512 & 30 & 1 & V            & S, A, D   & N, Am       &Fitness& Web  \\ \midrule [1.0pt]
    \rowcolor[HTML]{9AFF99}
    \textbf{MyoMechanix}                           & \textbf{7512} & \textbf{20} & \textbf{5} & \textbf{V, Sk, sE, P}  & \textbf{S, A, D, E, Fb} & \textbf{N, Am, Ex} &\textbf{Fitness}& \textbf{MoCap}  \\ \bottomrule [1.5pt]
    \end{tabular}
\label{table_related_dataset}
\end{table*}

The AQA field has advanced significantly over the past decade, with high-quality datasets (e.g., \cite{pirsiavash2014assessing, gao2014jhu, sardari2020vi, chen2024gaia, sener2022assembly101, doughty2018s, doughty2019pros}) playing a central role in this progress. However, despite this growth, existing datasets remain fundamentally limited in capturing the full complexity of human movement, particularly with respect to internal physiological processes, skill diversity, and structured reasoning. Our dataset aims to address this as follows.

\paragraph{Lack of physiological signals such as muscle mechanics.} A wide variety of AQA datasets are available, containing various modalities such as Video \cite{pirsiavash2014assessing, gao2014jhu, xu2022finediving,feng2025you, zheng2026fitaqa}, Pose \cite{ogata2019temporal, sardari2020vi, dong2024lucidaction, ji20263d}, Audio \cite{zeng2020hybrid}, and Language\cite{parmar2019and}. However, to the best of our knowledge, incorporating physiological signals such as electromyography (EMG), heart rate, etc., remains to be accomplished. Signals such as EMG can be very useful in AQA because they are probes of muscles---actuators of human movement---that tell how the muscles are being fired/activated and whether they are performing optimally. These are important insights because, often, a person's posture on the surface may look normal, but underneath, muscles are not being activated properly, are producing imbalanced forces, or are overcompensating for other muscles, etc., resulting in injury or the development of bad form in the long term. Importantly, a lack of muscle-level information hinders the ability to provide precise coaching recommendations. MyoMechanix bridges this crucial gap by providing multimodal information, including synchronized multiview RGB video, 3D pose, sEMG signals, heart signals, and breathing signals, across over 7.5K samples for 20 different exercise procedural actions that involve multiple joints and are highly prone to injury \autoref{table_data_fitness}.

\paragraph{Lack of diversity in ability levels.} A majority of the datasets in the AQA field have been collected by web crawling (see \autoref{table_related_dataset}). This collection process, although fast and scalable, suffers from an imbalance in terms of skill level as summarized in \autoref{table_data_al}. Recent works such as EgoExo-4D \cite{grauman2024ego}, EgoExo-Fitness \cite{li2024egoexo}, LucidAction \cite{dong2024lucidaction}, and Struggle \cite{feng2025you} mark a shift in data collection from rough web crawling to physical acquisition in controlled environments. While this paradigm effectively mitigates limitations such as poor data quality and action homogenization, these emerging datasets still fail to fully accommodate subjects' diverse skill levels and overlook the research gap in modeling implicit physiological states. Specifically, existing works often restrict their focus to self-loaded fitness actions, completely ignoring the highly challenging and complex domain of weight training. To overcome this limitation, we make a major contribution by meticulously collecting the full, large-scale MyoMechanix dataset in-house. This is a significantly more effortful option, but it results in a much more desirable, balanced representation of skill level, allowing for the mitigation of skill-level bias.

\paragraph{Lack of intermediate structured reasoning.} Another limitation stemming from web-crawled datasets is that they lack an intermediate structured reasoning process behind the scores predicted by the models \cite{pirsiavash2014assessing, parmar2017learning, xu2022finediving}. This lack of intermediate reasoning naturally hinders the development of causal and interpretable models. To bridge this gap, we construct the Fitness Knowledge Graph (FKG), which organizes annotations from field experts into structured relationships linking actions, phases, key steps, error types, and corrective feedback \cite{standard2020,bsu2013,HRDSport_Toolkit,joe}. This representation supports a compositional reasoning and scoring framework that enables interpretable and fine-grained quality assessment.

\paragraph{Scientific blind spot in load-based movement modeling and analysis.} AQA datasets cover a wide range of domains, including fitness. However, current fitness AQA datasets have two limitations: 1) they largely focus on self-loaded or low-load exercises/actions (e.g., \cite{ogata2019temporal, grauman2024ego, li2024egoexo})---failing to capture critical factors unique to weightlifting, such as load magnitude, joint torque distribution, and deep muscle activation; and 2) they are limited in terms of exercises they cover \cite{ogata2019temporal, parmar2022domain}. Weight-loaded training introduces significantly greater biomechanical complexity and injury risk than self-loaded movements, requiring precise coordination among multiple joints and muscle groups under increased external forces. Weight-loaded training also has unique movement and error patterns that are not covered in self-loaded exercises. This creates a scientific blind spot in modeling and understanding human movement under load. As a result, existing approaches are insufficient to accurately assess performance quality or capture the true physiological and neuromuscular demands of weight-loaded exercises. Our MyoMechanix overcomes this inherent limitation by covering 20 diverse weight-loaded exercises. These weight-loaded exercises are also compound exercises that involve multiple body parts and joints and carry a higher risk of injury.

\subsection{Action Quality Assessment Models}
\label{Sec22-AQA_Method}

Action Quality Assessment (AQA) focuses on automatically evaluating the quality of human actions from video data, with important applications in sports analytics and intelligent coaching. Recent research has achieved substantial performance gains through advances in representation learning, evaluation strategies, and multimodal modeling. In this section, we organize existing methods into four key directions: fine-grained spatiotemporal representation, objective evaluation under subjective uncertainty, multimodal and vision-language modeling, and interpretability. This taxonomy provides a structured view of current progress while highlighting remaining challenges, which our proposed MyoMechanix dataset and CUBIST modeling paradigm aim to address.

\paragraph{Fine-grained action representation.} Recent AQA research has shifted from coarse global feature learning toward fine-grained spatiotemporal modeling. To capture the internal evolution of complex actions, mainstream methods employ temporal segmentation mechanisms or phase-aware attention modules to decompose continuous action sequences into semantically meaningful sub-phases. This decomposition enables precise local alignment and fine-grained feature aggregation. Representative methods include PSL \cite{zhang2023labe}, GDLT \cite{xu2022likert}, HGCN \cite{zhou2023hierarchical}, MCoRe \cite{an2024multi}, T2CR \cite{ke2024two}, and FineParser \cite{xu2024fineparser}. In parallel, spatial modeling has evolved toward actor-object-centric representations, which highlight the action subject while suppressing irrelevant background noise, thereby improving sensitivity to subtle motion variations. Representative methods include JR-GCN \cite{pan2019action}, ACTION-NET \cite{zeng2020hybrid}, Sport-Cap \cite{chen2021sportscap}, and NS-AQA \cite{okamoto2024hierarchical}. Some works also explore self-supervised learning to acquire robust pose-motion representations \cite{roditakis2021towards, parmar2022domain, parmar2024learning}.

\paragraph{Objective evaluation strategies.} To address the inherent subjectivity and label ambiguity in manual scoring, one line of work introduces uncertainty modeling to explicitly capture cognitive ambiguity by learning the distribution of human scores rather than predicting a single deterministic value. Representative methods include USDL \cite{tang2020uncertainty}, UD-AQA \cite{zhou2022uncertainty}, and DAE \cite{zhang2024auto}. Another line of work adopts contrastive regression, which leverages expert demonstrations as reference anchors and estimates relative quality by measuring spatiotemporal differences between the evaluated sample and benchmark exemplars. Representative methods include CoRe \cite{yu2021group} and TPT \cite{bai2022action}. In addition, NS-AQA \cite{okamoto2024hierarchical} proposes a causal framework that decomposes scores into base-level components, thereby reducing subjectivity. Notably, such approaches can operate without relying on potentially biased ground-truth labels, which are influenced by human judgment (e.g., bias toward splash size in Olympic diving).

\paragraph{Multimodal and vision-language modeling.} Beyond representation and evaluation, recent advances enhance information richness and interaction by incorporating multi-source modalities and vision-language alignment. Given the limitations of a single visual modality in capturing complex kinematic patterns, multimodal approaches integrate heterogeneous signals such as RGB, skeleton, audio, and physiological data to improve robustness. Representative methods include PISA \cite{parmar2021piano}, MS-GCN \cite{lei2023multi}, PAMFN \cite{zeng2024multimodal}, and SkillSight \cite{wu2025skillsight}. Furthermore, vision-language frameworks introduce prior knowledge from large language models, enabling AQA systems to move beyond scalar score prediction toward natural language feedback with fine-grained error localization and corrective suggestions. Representative methods include MTL-AQA \cite{parmar2019and}, NAE \cite{zhang2024narrative}, VATP-Net \cite{gedamu2024visual}, and VidDiff \cite{burges2025video}.

\paragraph{Interpretability.} Despite these advances in representation, evaluation, and multimodal understanding, most existing approaches rely on end-to-end deep architectures and thus remain inherently black-box, limiting the interpretability of their predictions, particularly in fitness-related domains. Interpretability, however, is critical for bridging the gap between theoretical research and real-world deployment. Models equipped with reasoning and explanatory capabilities can significantly enhance transparency, reliability, and debuggability in intelligent coaching systems. Moreover, they facilitate the generation of detailed performance analysis reports, providing actionable and practical corrective feedback for practitioners \cite{okamoto2024hierarchical, matsuyama2023iris, majeedi2024rica}. Nevertheless, interpretable approaches for procedural fitness activity understanding remain underexplored.

\paragraph{Lack of detailed, fine-grained mechanism-level reasoning.} Despite the growing emphasis on interpretability, interpretable modeling remains largely underexplored in existing AQA research. A major reason is that mainstream AQA datasets are predominantly constructed from web-scraped sports competition videos \cite{pirsiavash2014assessing,parmar2017learning,parmar2019action,xu2022finediving}, and their annotation schemes typically provide only a final single score or multiple referee scores. As a result, these datasets largely lack fine-grained and compositional breakdowns of performance scores, as well as intermediate reasoning steps that explain how different aspects of the action execution contribute to the final assessment. Although some recent studies attempt to incorporate textual priors or perform sub-action decomposition, native, authoritative, and structured process-level annotations remain largely absent. This data-level limitation further constrains progress on the modeling side, as models have limited supervision for learning interpretable, mechanism-aware assessment processes rather than relying primarily on outcome-level score prediction.

To address this critical data and knowledge bottleneck, we introduce the MyoMechanix dataset, which leverages the \textbf{Fitness Knowledge Graph (FKG)} as a domain-specific ontology to reformulate traditional subjective scoring into objective, rule-based, fine-grained error annotations meticulously provided by multiple domain experts throughout the action execution procedure (see \autoref{Sec33-Annotation}). Building upon this design, the dataset establishes a compositional scoring system based on error weights. As a result, MyoMechanix is the first benchmark in AQA to simultaneously provide final macroscopic scores, microscopic execution parameters, and highly interpretable intermediate reasoning step annotations. 

Furthermore, we introduce a new modeling paradigm -- \textbf{CUBIST} (Compositional Ontological Reasoning Engine)---a framework for compositional reasoning over human actions that leverages the multimodal richness and structured representations of FKG. Inspired by the principle of decomposition–analysis–recomposition, CUBIST operates by (i) decomposing actions into structured components, (ii) analyzing each component through multimodal inputs and specialized processing modules, and (iii) recomposing the results to produce coherent assessments, including error attribution and corrective feedback. This design enables the model to move beyond holistic scoring toward interpretable, step-wise reasoning that more closely reflects expert analysis.

\subsection{Video Question Answering Datasets}
\label{Sec23-VQAD}

The Video Question Answering (VideoQA) task involves answering questions related to corresponding videos. VideoQA datasets have evolved from simple short-video understanding to more advanced multimodal and spatiotemporal reasoning. Early datasets focused on high-level semantic questions about short clips \cite{xu2017msrvttqa, jang2017tgifqa}, while later datasets \cite{yu2019activitynet, lei2019tvqa+} introduced longer videos and improved grounding for cross-modal alignment. Recent work \cite{parmar2024causalchaos} further explores complex reasoning tasks such as causal inference and human behavior analysis. However, despite these advances, existing datasets still represent human actions at a coarse level, lacking fine-grained annotations of posture, movement quality, and execution details.

There is a lack of datasets that provide corrective feedback, limiting the ability to answer questions about how to improve movement. Most existing AQA datasets only include simple supervision signals, such as numerical scores or phase labels \cite{parmar2017learning, parmar2019action, xu2022finediving, zhang2023logo}. Although some efforts have introduced text annotations, these are often noisy, unstructured, or incomplete—particularly lacking coverage of weight-based fitness actions \cite{grauman2024ego, li2024egoexo}. Moreover, their reliability is uncertain due to the lack of expert validation.

To address these limitations, we introduce MyoMech-anix-VideoQA, a fine-grained and feedback-oriented Vid-eoQA benchmark for weight-based fitness actions. Unlike existing datasets that represent human actions at a coarse level, lacking fine-grained annotations of posture, movement quality, and execution details, MyoMechanix-Video-QA enables language-conditioned reasoning over posture, movement phases, key steps, execution errors, muscle involvement, causal relations, and corrective strategies. Built from the expert-grounded Fitness Knowledge Graph (FKG), it converts structured action knowledge into graph-grounded question–answer pairs, providing reliable supervision for both diagnostic understanding and actionable feedback generation. The dataset supports training vision-language models to reason not only about what action is performed, but also how well it is executed, why specific errors occur, and how they can be corrected, thereby addressing the lack of fine-grained annotations, instructional feedback, and expert validation in existing VideoQA datasets. More broadly, MyoMechanix-VideoQA provides a flexible framework for developing models capable of interactive and interpretable reasoning, with applications ranging from AI coaching systems to general-purpose foundation models for fine-grained human action understanding.

\subsection{Electromyography Datasets}
\label{Sec24-EMGD}
Electromyography (EMG) captures the electrophysiological signals generated during skeletal muscle contraction. It effectively quantifies muscle activation levels, evaluates motor coordination across multiple muscle groups, and provides a physiological basis for detecting anomalous action. Regarding acquisition techniques, EMG is primarily categorized into surface electromyography (sEMG) using non-invasive skin patch electrodes and needle electromyography (nEMG) relying on micro-probes implanted into muscle tissues \cite{alcan2023current}. Due to the non-invasive nature and high portability of the former, existing EMG datasets widely adopt sEMG technology. This technology sees extensive application in advanced computational fields, including biomechanical analysis and gesture intention estimation.

However, a review of current research indicates that the vast majority of datasets treat sEMG signals merely as auxiliary features for coarse-grained action classification \cite{zhang2017extracting,hug2019individuals,jarque2020large,khan2020semg,dimitrov2023hig,wang2023wearable,atzori2014electromyography,liu2021neuropose,ozdemir2022dataset,salter2024emg2pose}. They overlook the deep underlying relationship between electromyography signals and fine-grained action quality. In complex fitness scenarios, precise neuromuscular control and the functional state of deep muscles provide insight into underlying neuromuscular dynamics, including muscle activation patterns and coordination efficiency. These factors are some of the key determinants of whether an exercise is performed correctly. Since visual modalities alone often fail to capture such internal activation patterns, incorporating EMG signals can enhance the physiological fidelity of human models and support more objective assessments of movement quality.

To address this gap, MyoMechanix introduces the first benchmark dataset that systematically incorporates real electromyography signals into action quality assessment. The dataset is collected using a sixteen-channel sEMG system operating at a high sampling rate of 2 KHz \cite{fastmove}, which enables continuous recording of muscle activation patterns during weight-loaded fitness exercises. Unlike vision-only datasets, MyoMechanix captures both external movement trajectories and internal neuromuscular responses, providing a more physiologically grounded representation of exercise performance. In addition, strict hardware-level synchronization between the sEMG and MoCap systems ensures precise temporal alignment between physiological and kinematic signals. This design supports fine-grained cross-modal analysis and enables more objective assessment of movement quality with improved physiological fidelity. 

Beyond action quality assessment, the synchronized multimodal design of MyoMechanix enables emerging cross-modal tasks such as Video-to-EMG prediction, where models learn to estimate neuromuscular activation patterns from visual observations. This capability opens a pathway toward video-based surrogate sensing of muscle activity, reducing dependence on expensive and specialized EMG hardware. With promising validation, MyoMechanix further supports translation beyond fitness scenarios, offering a foundation for low-cost healthcare sensing applications such as rehabilitation monitoring, motor function assessment, remote neuromuscular analysis, and clinically informed evaluation of movement disorders.

\subsection{Muscle Activity Estimation}
\label{Sec25-Muscle}

In muscle state estimation and cross-modal analysis, some existing works infer active muscles from visual features \cite{kunyu_acmmm}. While these methods establish useful associations between visual actions and muscular involvement, they often formulate muscle estimation as a categorical prediction problem, similar to action recognition. In this setting, the model predicts muscle names rather than measured activation states. Consequently, the continuous and subject-specific neuromuscular dynamics reflected in real EMG signals remain insufficiently explored.

To address this limitation, this work introduces the Video2EMG task, which aims to estimate continuous muscle activation patterns from visual input. Compared with traditional active muscle detection methods, Video2EMG represents a fundamentally different formulation in terms of task objective, supervision signal, and evaluation granularity. Instead of predicting discrete muscle names or binary activation states, it uses real sEMG signals as physiological supervision to model fine-grained neuromuscular activity across consecutive video frames. By aligning external visual motion with internal muscle activation, Video2EMG encourages models to learn physiologically grounded representations and provides a pathway from coarse action-level recognition toward fine-grained assessment of movement quality.

More broadly, Video2EMG offers a promising direction toward video-based surrogate sensing of muscle activity, which may reduce reliance on expensive and specialized EMG hardware in scalable sensing scenarios. This capability further extends the relevance of MyoMechanix beyond fitness assessment, providing a foundation for clinically relevant healthcare applications such as rehabilitation monitoring, motor function assessment, remote neuromuscular analysis, and clinically informed evaluation of movement disorders.
\section{MyoMechanix Dataset}
\label{Sec3-Dataset} 

\subsection{Motivation}

Motivated by the limitations identified in existing benchmarks related to \textit{AQA}, \textit{VideoQA}, and EMG, we introduce \textbf{MyoMechanix}, the largest and first-of-its-kind multimodal dataset for weight-loaded action assessment and coaching. As discussed in the previous section, current datasets remain limited in four key aspects: 1) they lack physiological signals that reflect implicit muscle mechanics, 2) they provide insufficient diversity across performer ability levels, 3) they offer limited support for intermediate structured reasoning, and 4) they largely overlook load-dependent movement modeling and analysis.

MyoMechanix is constructed to address these gaps through in-house data acquisition, synchronized multi-view and multimodal recordings, and physiological sensing, including sEMG, heart rate, and respiratory rate. By bridging explicit visual movement features with implicit muscle mechanics, MyoMechanix provides a foundational testbed for biomechanically grounded Physical AI. Each action is meticulously annotated by multiple fitness experts with comprehensive error labels and detailed actionable feedback.

To support structured and interpretable reasoning, we construct a comprehensive \textit{Fitness Knowledge Graph} (FKG) that provides an ontology-like representation of fitness actions. The FKG captures the procedural structure of actions and the relationships between execution errors and corrective guidance by encoding actions, action key steps (AKs), fine-grained error types (ETs), and prescriptive feedback messages (FBs).

Unlike existing fitness datasets that primarily focus on self-loaded or low-load actions, MyoMechanix centers on weight-training actions, where external loads introduce greater biomechanical complexity, novel movement and error patterns, neuromuscular coordination demands, and injury-related assessment challenges.

This section presents the design of MyoMechanix, including the fitness action scope, data collection and annotation procedures, quality-control mechanisms, Fitness Knowledge Graph construction, compositional scoring, and benchmark design.

\subsection{MyoMechanix Design}

Fitness actions are complex, skilled procedural activities consisting of temporally ordered movement steps that require precise posture, coordination, force control, and joint alignment. Consequently, they provide a structured setting for linking explicit visual motion patterns with implicit physiological and biomechanical mechanisms. While existing fitness-related benchmarks—such as Waseda-Squat \cite{ogata2019temporal}, Fitness-AQA \cite{parmar2022domain}, and EgoExo-Fitness \cite{li2024egoexo}—have established a foundation for action assessment, they focus almost exclusively on self-loaded or low-load exercises. Although these movements are accessible and safe for data collection, they fail to capture the unique biomechanical complexity introduced by external resistance.

Weight-loaded exercises differ fundamentally from bodyweight movements, as external loads (such as barbells or dumbbells) significantly alter the human movement profile. The introduction of resistance increases joint torque, demands deeper muscle activation, and requires more intricate neuromuscular coordination. Furthermore, the heightened injury risk associated with improper technique under load necessitates a more granular level of postural assessment than bodyweight datasets currently provide. This discrepancy creates a significant scientific gap: existing models cannot sufficiently account for the specific movement patterns and error modes unique to strength training.

Current action quality assessment (AQA) datasets remain insufficient for bridging this gap due to two primary limitations. First, by emphasizing low-load exercises, they overlook critical factors such as load magnitude and joint torque distribution under resistance. Second, the diversity of these datasets is often restricted; for instance, the Waseda-Squat dataset \cite{ogata2019temporal} is confined to a single exercise performed by a single subject with simulated errors rather than realistic weight-loaded conditions. As a result, there is a clear need for benchmarks that represent the physiological and biomechanical realities of professional and recreational strength training.

To address these limitations, MyoMechanix focuses on weight-loaded compound exercises. These exercises involve multiple joints and muscle groups, impose higher biomechanical demands, and require stricter form control for safe and effective execution. Given the practical constraints of dataset collection, we select 20 representative weight-loaded actions according to four principles:
\textbf{1)} covering both upper- and lower-body muscle groups;
\textbf{2)} involving complex and injury-prone joints, such as the shoulders, knees, and hips;
\textbf{3)} including free-weight actions that require greater stabilization and carry elevated injury risk;
and \textbf{4)} focusing on widely practiced exercises to ensure broad relevance to common strength-training regimens.

Through this design, MyoMechanix complements existing fitness benchmarks by capturing diverse, realistic, and biomechanically challenging weight-loaded actions. This makes the benchmark intrinsically more challenging, as weight-loaded action assessment requires reasoning beyond visible pose geometry. External resistance changes movement dynamics, introduces equipment occlusions, amplifies subtle but safety-critical error modes, and requires models to account for implicit biomechanical factors such as joint torque, load distribution, stabilization, fatigue-induced tremors, and neuromuscular control. By focusing on these challenging movements, MyoMechanix advances fitness AQA beyond coarse geometric evaluation toward fine-grained assessment of biomechanical integrity, performance quality, and injury risk. A detailed comparison between the proposed dataset and existing benchmarks is provided in \autoref{table_data_fitness}. \\

\begin{table}[t]
\caption{\textbf{Comparison between MyoMechanix and existing SOTA fitness datasets.} EWL: equipment (barbell, dumbbell, etc.)-based weight loading; RI: level of risk of injury.}
\label{table_data_fitness}
\centering
    \begin{tabular}{@{}>{\raggedright\arraybackslash}p{2.7cm}>{\centering\arraybackslash}p{1.8cm}>{\centering\arraybackslash}p{1.5cm}>{\centering\arraybackslash}p{1.2cm}@{}}
    \toprule [1.5pt]
    \textbf{Dataset}                    & \textbf{Exercises} & \textbf{EWL} & \textbf{RI} \\ \midrule
    Fitness-AQA \cite{parmar2022domain} & 3                  & \cmark       & high\\
    EgoExo-Fitness \cite{li2024egoexo}  & 12                 & \xmark       & low\\ 
    FitAQA \cite{zheng2026fitaqa}       & 30                 & \xmark       & low\\
    \midrule[1.0pt]
    \rowcolor[HTML]{9AFF99} 
    \textbf{MyoMechanix}                  & \textbf{20}        & \cmark       & \textbf{high} \\ \bottomrule [1.5pt]
    \end{tabular}
\end{table}

\begin{figure}[!t]
    \centering
    \includegraphics[width=\linewidth]{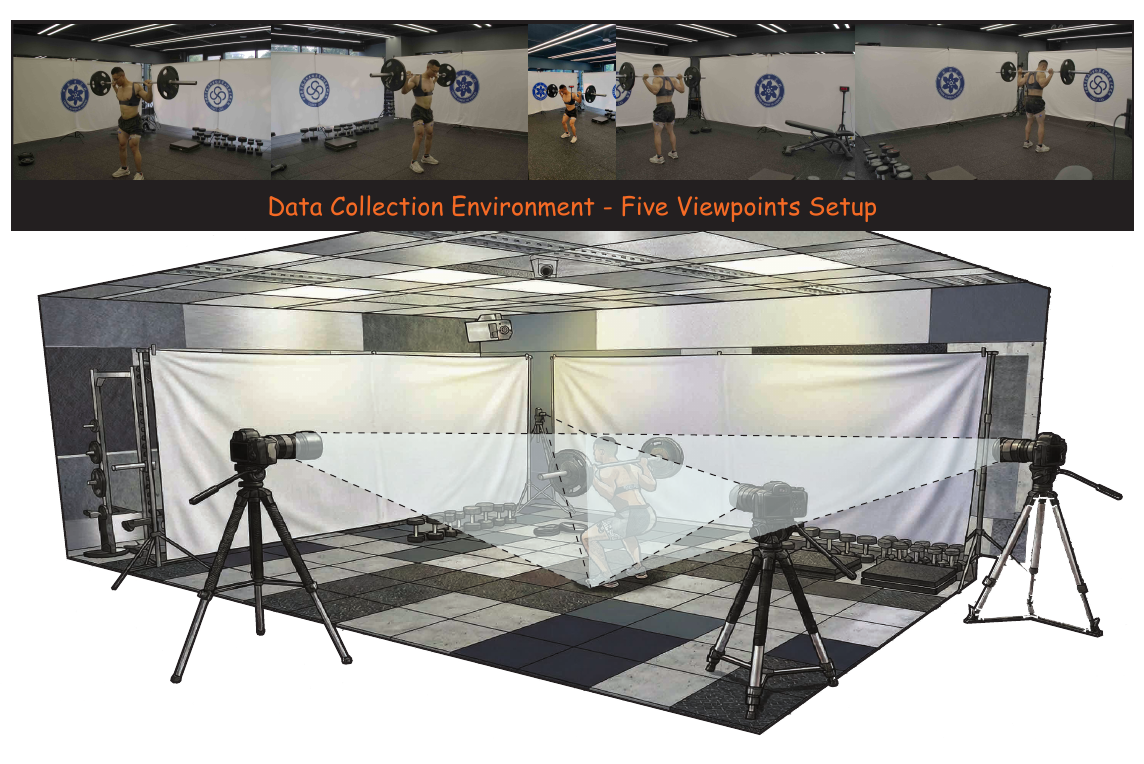}
    \caption{\textbf{The Data Acquisition Environment.} We equipped four cinema cameras fixed at the four corners of the collection area, along with one smartphone camera included to simulate typical user-recording scenarios and improve in-the-wild generalizability. Video, sEMG, heart rate, and respiratory rate signals were recorded synchronously during exercise performance by subjects.}
    \label{fig_dataset_collect}
\end{figure}

\noindent\textbf{Multiview Capture.}
The biomechanical and visual complexity of weight-loaded actions directly motivates our multiview capture design. Since action quality in strength training depends on fine-grained joint alignment, limb trajectories, and equipment-body interactions, single-view observations are often insufficient for reliable visual assessment. This limitation is further exacerbated by frequent self-occlusions and equipment occlusions caused by barbells, dumbbells, and compact lifting postures. Multiview capture is therefore essential for robust visual human modeling and sensing, as complementary viewpoints reduce occlusion, improve 3D pose reconstruction, and enable more reliable assessment of subtle postural and movement errors.
To obtain comprehensive ground-truth motion representations, including multi-view videos and 3D skeletal points, we use a high-precision markerless motion capture system \cite{fastmove_gait}. This setup uses four synchronized ZCAM E2M4 cinema cameras \cite{zcam} with Olympus 14--150 mm lenses fixed at 14 mm \cite{lens}, capturing at 4K resolution (3840$\times$2160 pixels) and 120 FPS. We also include a standard smartphone camera, OnePlus 7 \cite{phone}, recording at 1080P resolution (1920$\times$1080 pixels) and 60 FPS. Setup visualized in \autoref{fig_dataset_collect}. Adding this smartphone viewpoint replicates typical user-recording scenarios, improving the dataset's practical utility for real-world fitness applications while maintaining precise multimodal synchronization.\\

\noindent\textbf{Multimodal Information.}
Beyond visual complexity, weight-loaded action assessment also involves latent physiological and biomechanical factors that cannot be fully observed from videos or skeletal points alone. While multiview capture improves visual human modeling and 3D motion reconstruction, it does not directly measure muscular activation, exertion, stabilization effort, or breathing control, all of which are closely tied to action quality and safety in strength training. This motivates a multimodal sensing design that complements visual observations with physiological signals, enabling more comprehensive assessment of both external movement patterns and internal bodily responses. Existing AQA datasets primarily focus on videos \cite{pirsiavash2014assessing}, text \cite{parmar2019and}, skeletal points \cite{capecci2019kimore}, or audio \cite{parmar2021piano}, with limited attention to physiological data. EMG, in particular, captures muscle activation patterns that are closely tied to force production, stabilization, and movement quality, providing critical information for fitness AQA and fine-grained feedback. Accordingly, the MyoMechanix dataset records three key physiological signals:
\textbf{1)} EMG: we use a 16-channel FastMove sEMG device \cite{fastmove} operating at a sampling rate of 2000 Hz to capture the target muscle groups' activity;
\textbf{2)} heart rate: collected with a customized wearable vest to provide insight into exercise intensity and physiological exertion;
and \textbf{3)} respiratory rate: also recorded via the vest, since proper breathing is an important procedural component of safe and effective weight-loaded exercise. Accordingly, our action procedures and rules incorporate breathing as an AQA criterion.\\

\noindent\textbf{Skill-Level Diversity for Robust Modeling.}
In addition to multimodal sensing, a diverse subject pool is essential for capturing the variability of weight-loaded action quality. Since strength-training performance depends strongly on proficiency, strength capacity, motor control, and exercise experience, datasets constructed from narrow subject populations may fail to represent the full range of movement patterns and error modes encountered in real-world fitness scenarios. 
To address the uneven proficiency distributions prevalent in existing AQA benchmarks, we recruit participants across three proficiency levels: novice, amateur, and expert.
This proficiency-balanced design reduces skill-level bias and supports computational modeling of a wide spectrum of movement patterns, error modes, and performance qualities across novice, amateur, and expert subjects.
Participants were recruited through two channels: research institutions and local gymnasiums. During the preliminary selection phase, over 60 applications were received from students and professional coaches. To ensure reliable proficiency stratification, all applicants underwent a standardized screening protocol consisting of online questionnaires, on-site athletic performance tests, and 3D body-composition analysis. This procedure enabled a quantitative evaluation of their strength capacity, weight-bearing ability, and motor skill level. Following previous work \cite{li2024egoexo}, we selected 38 subjects in total: 10 experts (IDs P01 to P10), 8 amateurs (IDs A01 to A10, excluding A06 and A09), and 20 novices (IDs N01 to N20). Unlike many existing datasets with limited proficiency coverage, MyoMechanix spans a broad spectrum of skill levels, as illustrated in \autoref{table_data_al}. This proficiency-balanced design helps reduce skill-level bias and supports more robust evaluation across diverse movement patterns, error modes, and performance qualities.

\subsection{Data Acquisition Protocol}

\noindent\textbf{Human subject safety.}
To ensure participant safety during the collection of risk-prone weight-loaded actions, subjects were instructed to perform each action at 80\% of their one-repetition maximum. In addition, all subjects were provided with sports injury insurance, and on-site safety supervisors and emergency medical kits were available throughout the experiments. The data collection protocol was reviewed and approved by the Institutional Review Board (IRB) of Beijing Sport University (Approval No.: 2023313H). All participants provided written informed consent prior to participation.\\

\noindent\textbf{Execution Procedure.}
At the beginning of data acquisition, each subject was informed of the specific actions to be performed. Staff assisted subjects in wearing the customized vest correctly and securely attaching the sEMG sensors to the target muscles, ensuring stable skin contact to reduce signal noise and acquire high-quality physiological signals. During data collection, physiological signals were monitored to capture internal exertion and fatigue-related responses, providing important references for subsequent fine-grained annotation.

Unlike prior in-house datasets such as FLAG3D \cite{tang2023flag3d} and EgoExo-Fitness \cite{li2024egoexo}, subjects entered the collection area individually to avoid observing others or receiving detailed instructions, thereby minimizing external influence. No textual descriptions were provided. Instead, a brief demonstration was given before each session to ensure that subjects performed the actions based primarily on their own experience. After sensor attachment, each subject faced the curtain between Viewpoints 1 and 2 and performed the action upon the staff's signal. The staff then recorded the corresponding weight data, and the entire process was supervised by the authors.

Furthermore, to support the modeling of the temporal evolution of action quality, each subject was required to complete 10 continuous repetitions for each action. Such continuous repetition sequences are largely absent from existing datasets.  

\begin{table}[t]
\caption{\textbf{Comparison of ability/skill level diversity of the subjects in AQA datasets.} N: Novice, A: Amateur, P: Professional.}
\label{table_data_al}
\centering
\begin{tabular}{@{}>{\raggedright\arraybackslash}p{3.5cm}*{3}{>{\centering\arraybackslash}p{1.2cm}}@{}}
    \toprule [1.5pt]
    \textbf{Dataset} & \textbf{N} & \textbf{A} & \textbf{P} \\ \midrule [1.0pt]
    Sport Datasets \cite{pirsiavash2014assessing, parmar2017learning, xu2022finediving} & & & \cmark \\
    Waseda-Squat \cite{ogata2019temporal}   & \cmark &           & \\
    Fitness-AQA \cite{parmar2022domain}     &           & \cmark & \\
    EgoExo-Fitness \cite{li2024egoexo}      & \cmark & \cmark & \\
    LucidaAction \cite{dong2024lucidaction} &           & \cmark & \cmark \\
    FitAQA \cite{zheng2026fitaqa}           & \cmark & \cmark & \\
    \midrule[1.0pt]
    \rowcolor[HTML]{9AFF99}
    \textbf{MyoMechanix} & \textbf{\cmark} & \textbf{\cmark} & \textbf{\cmark} \\
    \bottomrule[1.5pt]
\end{tabular}
\end{table}

\subsection{Expert-Guided Data Annotation Framework}
\label{Sec33-Annotation}

\begin{figure*}[h]
    \centering
    \includegraphics[width=\linewidth]{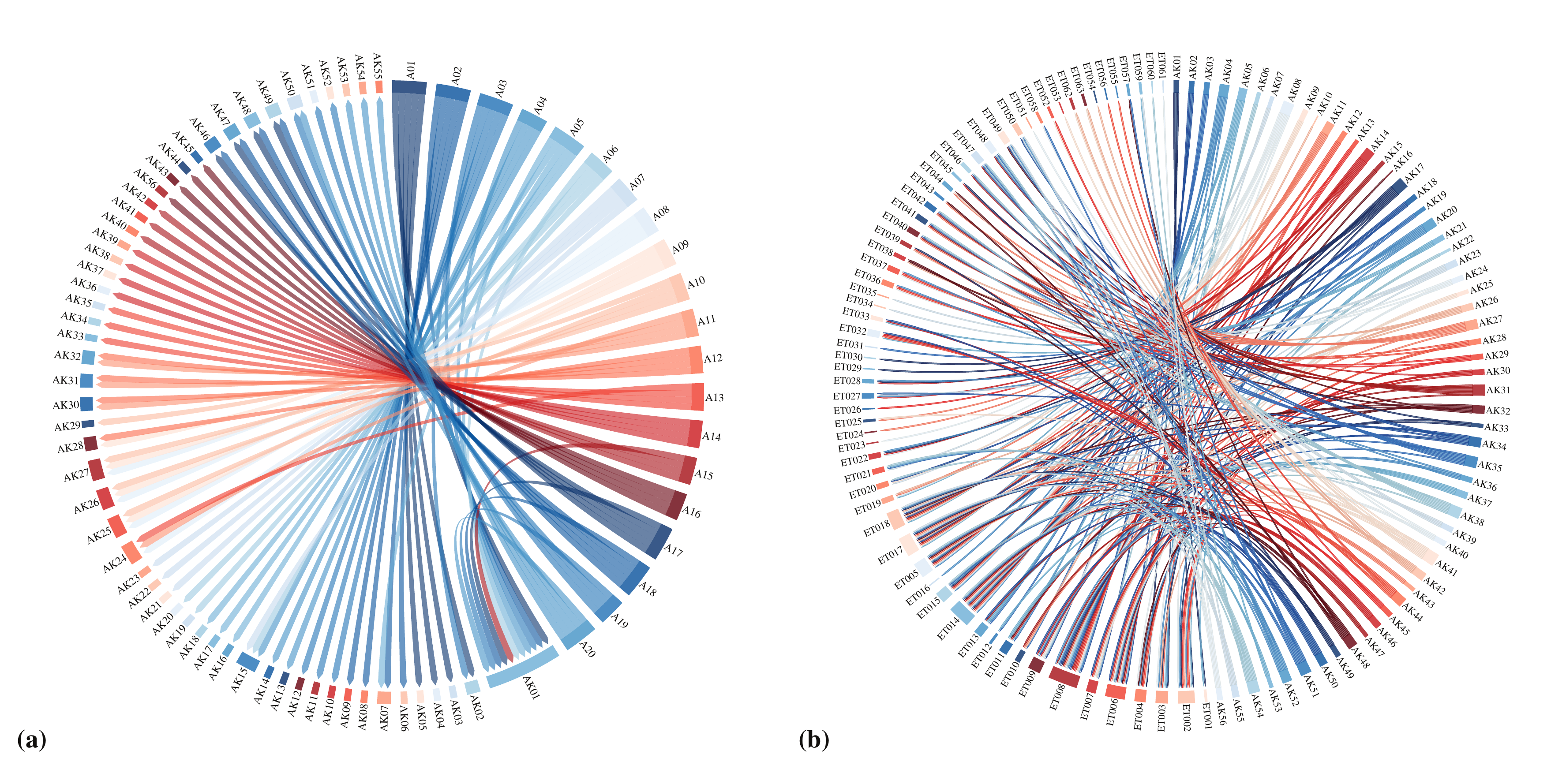}
    \caption{\textbf{Visualization of the entity relationships within the Fitness Knowledge Graph.} \textbf{(a)} Mapping between actions and action keysteps. \textbf{(b)} Mapping between action keysteps and error types.}
    \label{fig_dataset_annot_kg}
\end{figure*}

Accurate annotation is essential for transforming raw multiview and multimodal recordings into a benchmark for fine-grained fitness action quality assessment. For weight-loaded exercises, annotation is particularly challenging because action quality depends on phase-specific movement execution, localized postural errors, biomechanical safety, and physiological factors such as breathing and fatigue. To ensure consistent and interpretable labels, we develop an expert-guided annotation framework that integrates standardized fitness guidelines, biomechanical phase decomposition, structured error taxonomy, a Fitness Knowledge Graph (FKG), and a compositional scoring system. This section first describes how the annotation guidelines are constructed, then introduces the structured annotation framework, and finally details the expert annotation workflow.

\subsubsection{Annotation Guideline Construction}
Unlike competitive sports with strictly defined action objectives and scoring criteria, such as diving or gymnastics, fitness actions target diverse goals such as muscle growth, fat loss, rehabilitation, and general physical conditioning. Consequently, evaluation standards for a single action may vary across individuals, training contexts, and coaching practices. To ensure high-quality and consistent annotations for MyoMechanix, we collected existing fitness-action evaluation standards from multiple sources, including national standards, fitness associations, professional literature, fitness applications, and domain experts. By synthesizing these sources, we established an expert-guided annotation system for weight-loaded fitness actions, which can also be adopted by future work.\\

\noindent\textbf{Rule formulation for Fitness Skill Evaluation via multisource knowledge integration.}
Specifically, our annotation rules drew primarily on the "National Occupational Skill Standards—Social Sports Instructor (Occupation Code: 4-13-04-01)", jointly formulated by the Ministry of Human Resources and Social Security of the People's Republic of China and the General Administration of Sport of China \cite{standard2020}. Additionally, we also reference the "Fitness and Bodybuilding Tutorial" from Beijing Sport University \cite{bsu2013}, "Occupational Competency Training Textbook for Social Sports Instructors—Fitness Coaches" published by the Human Resources Development Center of the General Administration of Sport of China \cite{HRDSport_Toolkit}, and "Joe Weider's Bodybuilding System" \cite{joe}. 

While several of these primary texts are localized national standards, they adhere to global training principles and biomechanics. These domestic guidelines are rigorously benchmarked against frameworks from the American College of Sports Medicine (ACSM) and the National Strength and Conditioning Association (NSCA), as well as advanced European and Australian sports science practices. Consequently, these sources provide comprehensive and broadly applicable guidance on action mechanics, posture, and safety considerations.

We thoroughly reviewed all sources and meticulously documented essential information relevant to the actions collected, emphasizing target muscles, action descriptions, and potential error types. By carefully synthesizing these details, we developed annotation guidelines that adhere to existing standards while accommodating practical training scenarios. Furthermore, these rules underwent rigorous verification by experts at Beijing Sport University, thereby ensuring theoretical soundness, consistency, and practicality in the annotation process.\\

\begin{figure*}
    \centering
    \includegraphics[width=\linewidth]{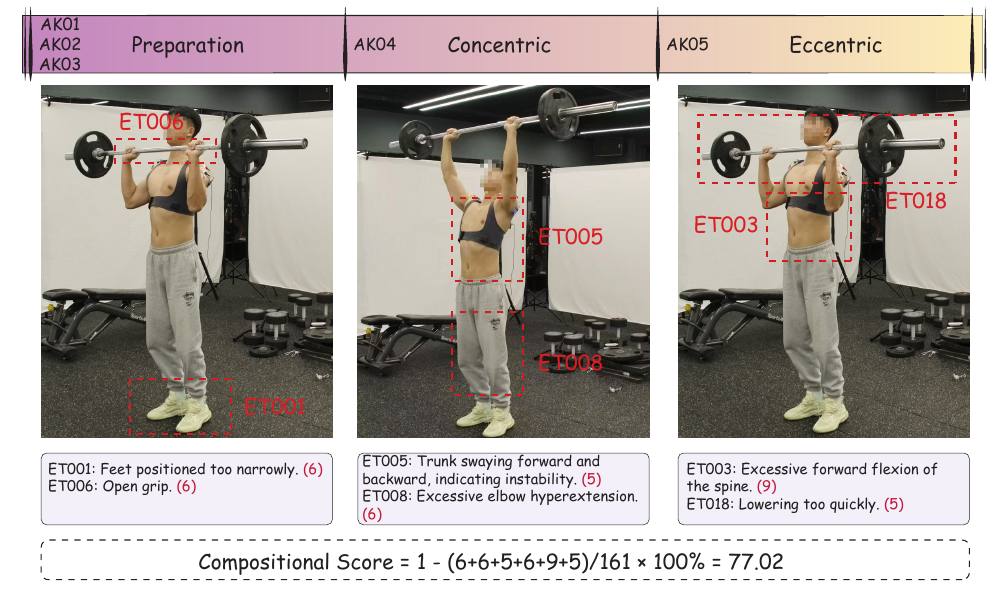}
    \caption{\textbf{Visualization of exemplary errors and compositional scoring process during one of the actions---BarBell Overhead Press.} 
    Several key errors were observed that could compromise exercise form and effectiveness, as mentioned in the following. \textbf{(a)} During the Preparation phase, 1) the stance was too narrow, and 2) the grip was open rather than closed. \textbf{(b)} During the Concentric phase, 1) trunk swaying and 2) excessive elbow hyperextension were observed. \textbf{(c)} During the Eccentric phase, 1) The barbell was also lowered too quickly, 2) accompanied by excessive forward spinal curvature. The corresponding score deductions are shown in the figure, resulting in a final action score of 77.02. (Zoom in for the best view.)}
    \label{fig_dataset_annot_score}
\end{figure*}

\subsubsection{Structured Annotation Framework}

Building on the expert-guided rules, we develop a structured annotation framework that converts fitness knowledge into fine-grained, machine-readable annotations for model training, evaluation, and interpretable reasoning. Each action is decomposed into biomechanical phases and action keysteps, with corresponding phase-specific error types and corrective feedback. These components are organized into a Fitness Knowledge Graph (FKG), enabling explicit modeling of the relationships among actions, execution steps, errors, and feedback. Finally, a compositional scoring system aggregates the annotated errors into an interpretable action quality score.\\

\noindent\textbf{Biomechanical phase decomposition.} To guide the annotation process and ensure consistency, a biomechani-cally-grounded framework was developed for structuring each action. In this framework, each collected action is decomposed into three distinct phases, based on dominant-limb movement patterns: Preparation, Concentric, and Eccentric.\\ 

\noindent\textbf{Action keysteps.} Within each phase, the primary limb actions are further subdivided into finer-grained action keysteps (AKs), each paired with integrated textual descriptions. By systematically combining these AKs, we generate rich textual representations of diverse fitness actions, supporting fine-grained action analysis and structured representation learning. Building on this framework, we next introduce extensions for error identification, structured knowledge representation, and annotation efficiency.\\

\noindent\textbf{Error Taxonomy paired with Corrective Feedback.} We cataloged fine-grained error types (ETs) that characterize localized joint deviations within each major limb action, pairing them with specific prescriptive corrective feedback. To systematically quantify the severity of these deviations, we established a hierarchical criterion for assigning penalty weights to each ET. These weights are determined based on their potential impact on physical safety, core stability, and movement efficacy, structured as follows:
\begin{itemize}[label=$\bullet$, labelwidth=!, labelindent=0pt, noitemsep,topsep=0pt]
\item Errors that directly compromise joint safety (e.g., spine, knee, hip, shoulder) receive the highest weight (9–10).
\item Errors that undermine core stability, disrupt movement trajectory, or alter force generation (e.g., relying on momentum) receive high weight (7–9).
\item Errors in foundational posture (e.g., initial stance, foot spacing) and insufficient movement amplitude receive moderate weight (6–7).
\item Auxiliary errors related to fine control receive a lower weight (5).
\item Breathing errors (incorrect inhalation or exhalation) receive the minimal weight (2).\\
\end{itemize}

\noindent\textbf{Fitness Knowledge Graph Construction.} To support structured and interpretable reasoning, we introduce a novel \textit{Fitness Knowledge Graph} (FKG) that formalizes weight-training exercises as biomechanically phased action representations. Each exercise is decomposed into constituent actions organized by biomechanical phases, such as preparation, eccentric, transition, and concentric phases. Within each phase, the FKG encodes ordered action key steps, phase-conditioned error types, and corresponding corrective feedback. This explicit semantic structure captures both the procedural organization of weight-training actions and the relationships between execution errors, their biomechanical consequences, and appropriate corrective guidance. Building on this ontology, we design a compositional, penalty-based annotation protocol that weights phase-specific errors according to their biomechanical impact, enabling interpretable and fine-grained action evaluation. FKG is illustrated in \autoref{fig_dataset_annot_kg}.\\

\noindent\textbf{Compositional Scoring System.} Based on our error taxonomy, we designed a compositional penalty-based scoring scheme to evaluate overall action quality. Our compositional scoring is a crucial contribution, providing a way to account for all errors in a transparent and traceable manner. Since our compositional AQA score is built in a bottom-up approach from individual errors, it enjoys full interpretability and offers full accountability/verifiability. This is in stark contrast to existing approaches and practices, which only produce a final single score that lumps together all the errors in a completely opaque and untraceable manner -- losing all interpretability and accountability/verifiability. This system separates the mathematical complexity from the annotation process, ensuring both rapid and consistent data labeling. Annotators are required to identify and select the observed ETs within a video clip, and the overall execution score is automatically computed by composing the respective penalties.

Specifically, the overall score is built up bottom-up by calculating the ratio of the sample’s cumulative error weight to the maximum possible error weight meticulously predefined for that action by experts as discussed in the previous section. For each sample, the compositional action quality score, $\mathcal{S}$, is computed using the following compositional formula:
\begin{equation}
\centering
\label{equa_score}
    \mathcal{S} = [1-\frac{\sum_{i=1}^{N}W_i}{\sum_{j=1}^{M}W_j}]\times100
\end{equation}
where $N$ represents the number of errors the sample contains, $W_i$ represents the weight of errors that the sample contains, $M$ represents the number of errors that the action type contains, and $W_j$ represents the weight of errors that the action type contains. The theoretical range of $\mathcal{S}$ is $[0, 100]$. We have visualized a detailed example of action quality assessment along with exemplary errors and the associated full compositional scoring process in \autoref{fig_dataset_annot_score}.

\begin{figure*}[h]
    \centering
    \includegraphics[width=\linewidth]{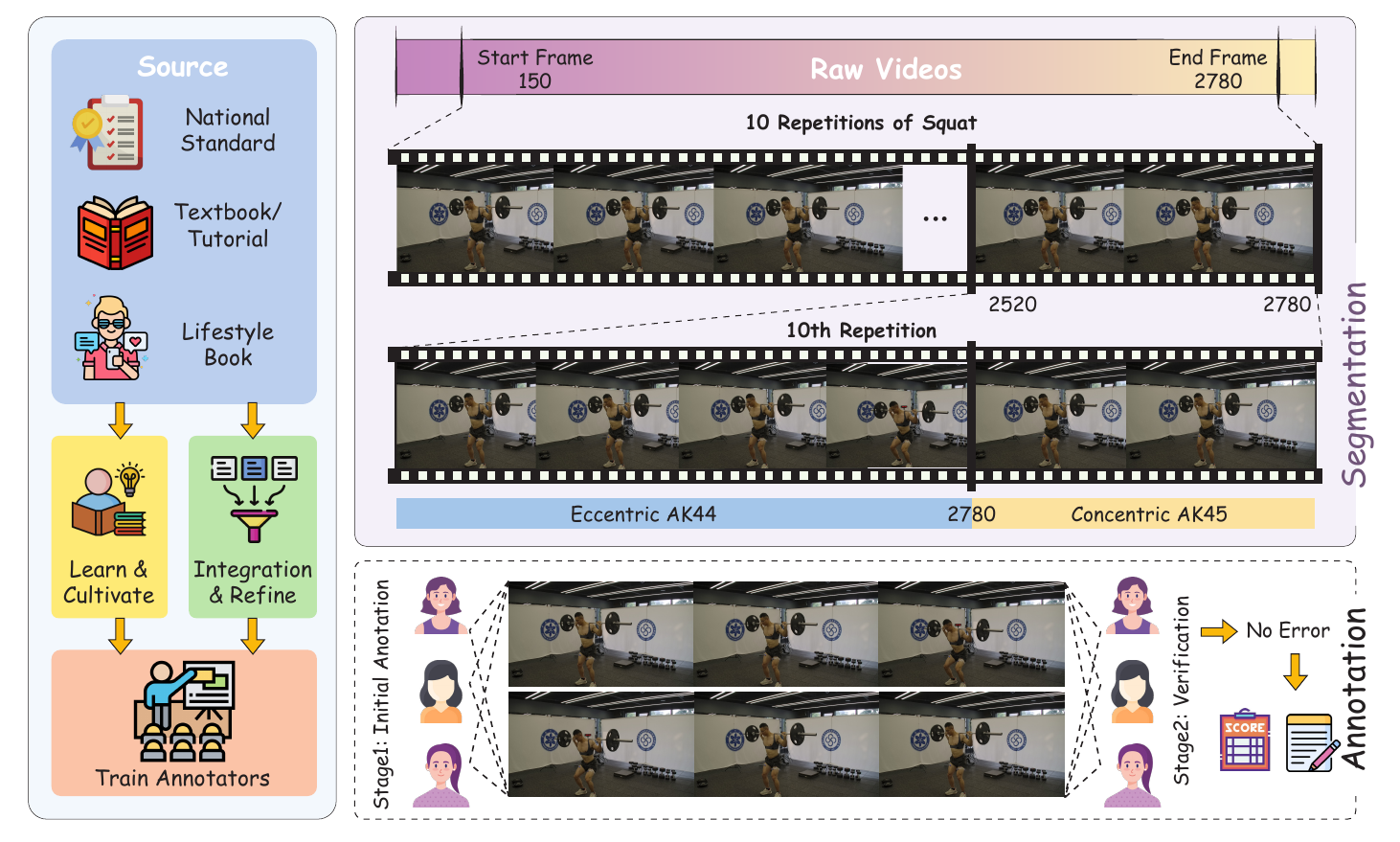}
    \caption{\textbf{Our Annotation Framework and Process.}
    (\textbf{Left}) We integrate multiple sources of knowledge, and annotators are trained on the provided guidelines, and they receive centralized instruction to ensure full understanding of the rules. (\textbf{Right Top}) Raw video data was segmented following predetermined criteria. (\textbf{Right Bottom}) We implement a two-stage expert annotation and verification process to reduce annotation errors and mitigate subjective bias.
    }
    \label{fig_dataset_annot_proc}
\end{figure*}

\subsubsection{Expert Annotation Workflow}

We annotate the following information for each sample: 
\textbf{1)} action segmentation;
\textbf{2)} action keysteps;
\textbf{3)} action errors;
and \textbf{4)} action quality assessment scores using the following systematic procedure.
The complete annotation process is visualized in \autoref{fig_dataset_annot_proc}.

\paragraph{Annotator Recruitment.}
Due to the substantial volume of the MyoMechanix dataset and the stringent requirements for annotation quality, 16 professionals, including fitness trainers and sports science graduate students, were recruited for annotation. To ensure consistency and scientific rigor throughout the annotation, we established in-depth collaboration with all annotators before project commencement, clarifying the annotation requirements, personnel qualifications, and the underlying logic of the guidelines. We then conducted a small-scale pilot annotation to validate these rules. Based on the pilot results, we refined the guidelines and finalized our annotator pool.

\paragraph{Action Segmentation.}
During action segmentation, continuous videos containing repeated actions were temporally segmented into individual samples. Annotators subsequently verified each discrete sample to ensure strict alignment with our established guidelines.

\paragraph{Annotation Task Organization.}
During the formal annotation phase, we organized the workload into distinct periods based on target muscle groups. At the onset of each period, we conducted centralized training for all 16 annotators to ensure strict adherence to the guidelines and high inter-annotator consistency.

\paragraph{Annotator Grouping.} 
Annotators were organized into 2 distinct groups based on the level of fitness experience: 1) Group 1 -- trainers with $<$ 3 years of experience and master’s students -- handled the initial annotation (Stage 1); 2) Group 2 -- trainers with $>$ 3 years of experience and doctoral students -- was responsible for annotation verification (Stage 2). 

\paragraph{Two-Stage Annotation.} 
Stage 1: Each sample was independently annotated for error types by 3 annotators from Group 1, utilizing the RGB video, 3D pose, and respiratory rate modalities. Stage 2: The initial sets of labels underwent a rigorous review by 3 senior annotators from Group 2. 

\paragraph{Verification Workflow.}
To effectively mitigate subjective bias and label noise, we implemented a progressive conflict-resolution protocol for any identified discrepancies. Initially, inconsistent annotations were resolved through a majority vote among the Stage 2 annotators. For highly ambiguous cases, the reviewers engaged in detailed discussions until a mutual agreement was reached. In exceptionally rare and complex instances, we invited collaborating domain experts to participate in the review meetings, facilitating a final consensus to rigorously determine the ground-truth annotation. Authors periodically monitored the quality of the annotated data to ensure that the dataset maintained a high degree of accuracy and reliability.

\subsection{Benchmark Design}
\label{Sec34-Design}

Building on the multimodal recordings and structured annotations in MyoMechanix, we establish a suite of benchmarks that extend beyond standard action quality assessment to support broader applications in fitness understanding, coaching, and physiological signal prediction. Specifically, we define three benchmark tasks: 
\textbf{1)} \textbf{Action Quality Assessment and Coaching}, with \textit{Vanilla}, \textit{Cross-Subject}, \textit{Cross-View}, and \textit{Mix-View} settings;
\textbf{2)} \textbf{Fine-Grained Video Question Answering};
and \textbf{3)} \textbf{Video-to-EMG Prediction}.
The following sections describe each benchmark in detail.

\subsubsection{MyoMechanix-AQA \& Coaching Dataset}
\label{Sec341-AQA_Dataset}

The MyoMechanix-AQA \& Coaching benchmark focuses on the core structured exercise assessment, including estimating execution quality, identifying form errors, explaining score deductions, and generating corrective feedback grounded in expert-defined criteria. It is designed to support both training and evaluation of models that learn our rubric-grounded knowledge about how weight-loaded actions should be assessed, how deviations from ideal execution should be penalized, and how corrective recommendations should be derived.

To support action quality assessment under varying levels of generalization difficulty, we design split protocols that evaluate both in-distribution performance and out-of-distribution generalization. Specifically, we establish a hierarchical evaluation framework with four splits: \textit{Vanilla}, \textit{Cross-Subject}, \textit{Cross-View}, and \textit{Mix-View}. For each split, samples are partitioned into training, validation, and test sets with a ratio of 60:20:20.

\begin{itemize}[label=$\bullet$,labelwidth=!, labelindent=0pt, noitemsep, topsep=0pt]

\item \textit{Vanilla.} This standard protocol follows common practice in AQA datasets and provides an in-distribution setting for estimating upper-bound model performance under minimal distribution shift. Within each action class, samples from different subjects and repetitions are aggregated and randomly partitioned into training, validation, and test sets.

\item \textit{Cross-Subject.} This protocol partitions subjects into disjoint training, validation, and test groups to evaluate model generalization across individuals.

\item \textit{Cross-View.} This protocol trains models on one camera view and tests them on another unseen view, simulating the distribution shift caused by viewpoint variation.

\item \textit{Mix-View.} This protocol trains models on three camera views and tests them on the remaining unseen view, assessing generalization to unseen viewpoints under partial multiview supervision.
\end{itemize}

We recommend a two-stage evaluation protocol: first, establishing an in-distribution baseline on the \textit{Vanilla} split, then evaluating robustness on the \textit{Cross-Subject}, \textit{Cross-View}, and \textit{Mix-View} splits. This progression enables systematic assessment of both standard AQA performance and generalization under real-world subject and viewpoint shifts.

\subsubsection{MyoMechanix-VideoQA Dataset}
\label{Sec342-VideoQA_Dataset}

\begin{figure*}[t]
    \centering
    \includegraphics[width=\linewidth]{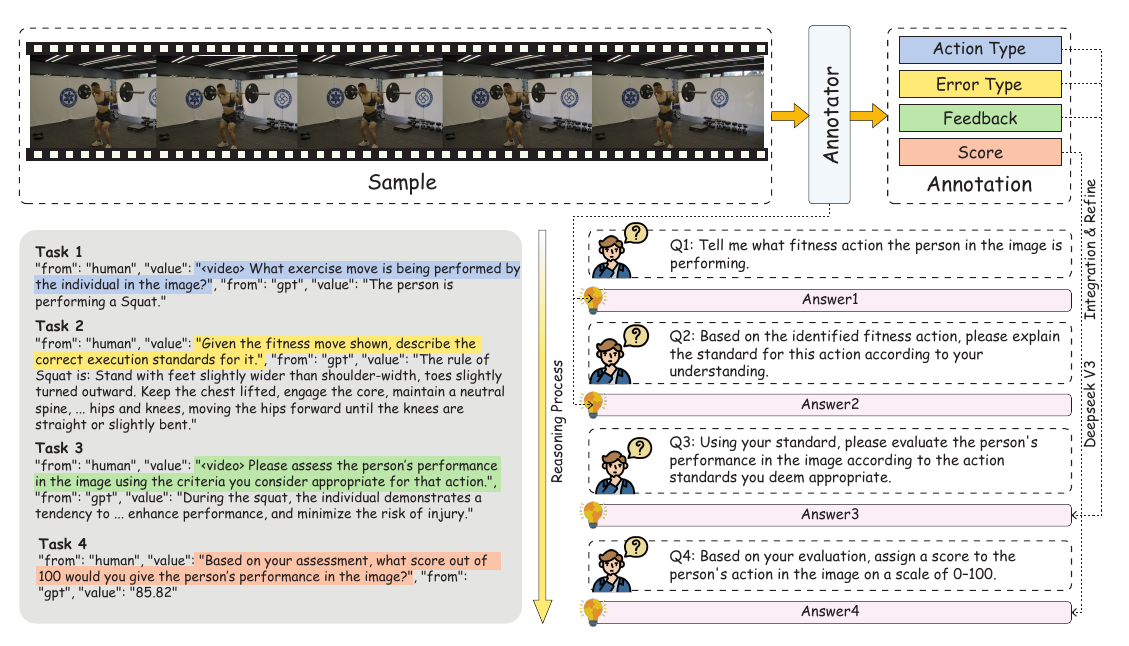}
    \caption{\textbf{The construction of the MyoMechanix-VideoQA dataset.}}
    \label{fig_dataset_vqa}
\end{figure*}

To develop foundational multimodal AI models capable of deep fine-grained understanding of complex actions, we introduce MyoMechanix-VideoQA, the first of its kind fine-grained video question–answering benchmark built directly on top of the dataset’s multimodal recordings \& the Fitness Knowledge Graph (FKG). Existing VideoQA datasets primarily focus on macroscopic scene understanding or coarse action recognition (e.g., identifying an individual's actions or forecasting subsequent movements). In contrast, MyoMechanix-VideoQA evaluates a model's capacity for structured and multimodal reasoning by targeting fine-grained biomechanical evaluation. By addressing localized postural deviations, movement mechanics, and prescriptive corrective feedback, the proposed benchmark shifts the paradigm from simply recognizing \textit{what is occurring} to providing actionable guidance/coaching on \textit{how execution can be improved}.\\

\noindent\textbf{Dataset Construction.}
Our FKG provides a structured ontology of actions, keysteps, error types, muscle groups, and corrective feedback. Leveraging this graph, we design a reasoning-oriented conversational QA pipeline that supports interactive user queries about exercise execution. The pipeline progresses through four sequential stages: action recognition, action standards, action evaluation, and action scoring, as shown in \autoref{fig_dataset_vqa}.

\noindent\textbf{Machine-Generated, Human-Verified QA Construction.}
Guided by this pipeline, a large-scale set of natural-language QA pairs was generated utilizing a mixture of rule-based templates and Large Language Models (LLMs). Specifically, complex responses for action evaluation were synthesized by DeepSeek-V3-Reasoning, which integrated video context, keysteps, errors, and feedback suggestions to ensure semantic coherence. Since every question is strictly grounded in explicit nodes of the FKG, the generated QA pairs maintain high biomechanical precision. Following the generation phase, all QA sets underwent rigorous manual verification and correction, resulting in 30,048 high-quality QA pairs.

The questions span four core reasoning categories:
\begin{itemize}[label=$\bullet$, labelwidth=!, labelindent=0pt, noitemsep, topsep=0pt]
\item \textit{Descriptive}: Inquire about observable attributes such as the action being performed or the current keystep (e.g., “Which action is shown in the video?”, “What keystep is the performer currently executing?”).
\item \textit{Relational reasoning}: Require linking multiple graph entities (e.g., “Which muscle group is primarily activated during the descent phase of a squat?” or “Which error in the setup step leads to a lower-back penalty?”).
\item \textit{Temporal}: Probe the understanding of sequential dependencies (e.g., “Which key step follows the liftoff phase?”).
\item \textit{Causal feedback}: Demand corrective advice based on observed biomechanical errors (e.g., “What feedback should be given if the knees cave inward during the ascent?”).\\
\end{itemize}

\noindent\textbf{Comprehensive multimodal reasoning platform.}
By transforming the structured annotations of the Fitness Knowledge Graph (FKG) into a large set of graph-grounded question–answer pairs, MyoMechanix-VideoQA evolves beyond a conventional AQA dataset into a comprehensive multimodal reasoning platform. It is designed to support the development of foundation vision-language models (VLMs) for language-grounded, query-conditioned, fine-grained action understanding. Rather than requiring fixed assessment outputs, VideoQA exposes models to diverse natural-language queries spanning action identity, movement phases, key steps, posture errors, muscle activation, temporal dependencies, causal relationships, and corrective strategies. This formulation enables supervised fine-tuning of VLMs that must not only recognize human actions but also reason over the hierarchical and causal structure linking actions, execution errors, underlying physiology, and feedback. Experimental results highlight the value of such structured graph supervision in enabling VLMs to learn representations of action quality that extend well beyond surface-level visual understanding. More broadly, MyoMechanix-VideoQA provides a flexible framework for training models capable of interactive and interpretable reasoning, with applications ranging from AI coaching systems to general-purpose foundation models for fine-grained human action understanding.

\begin{figure}[t]
    \centering
    \includegraphics[width=\linewidth]{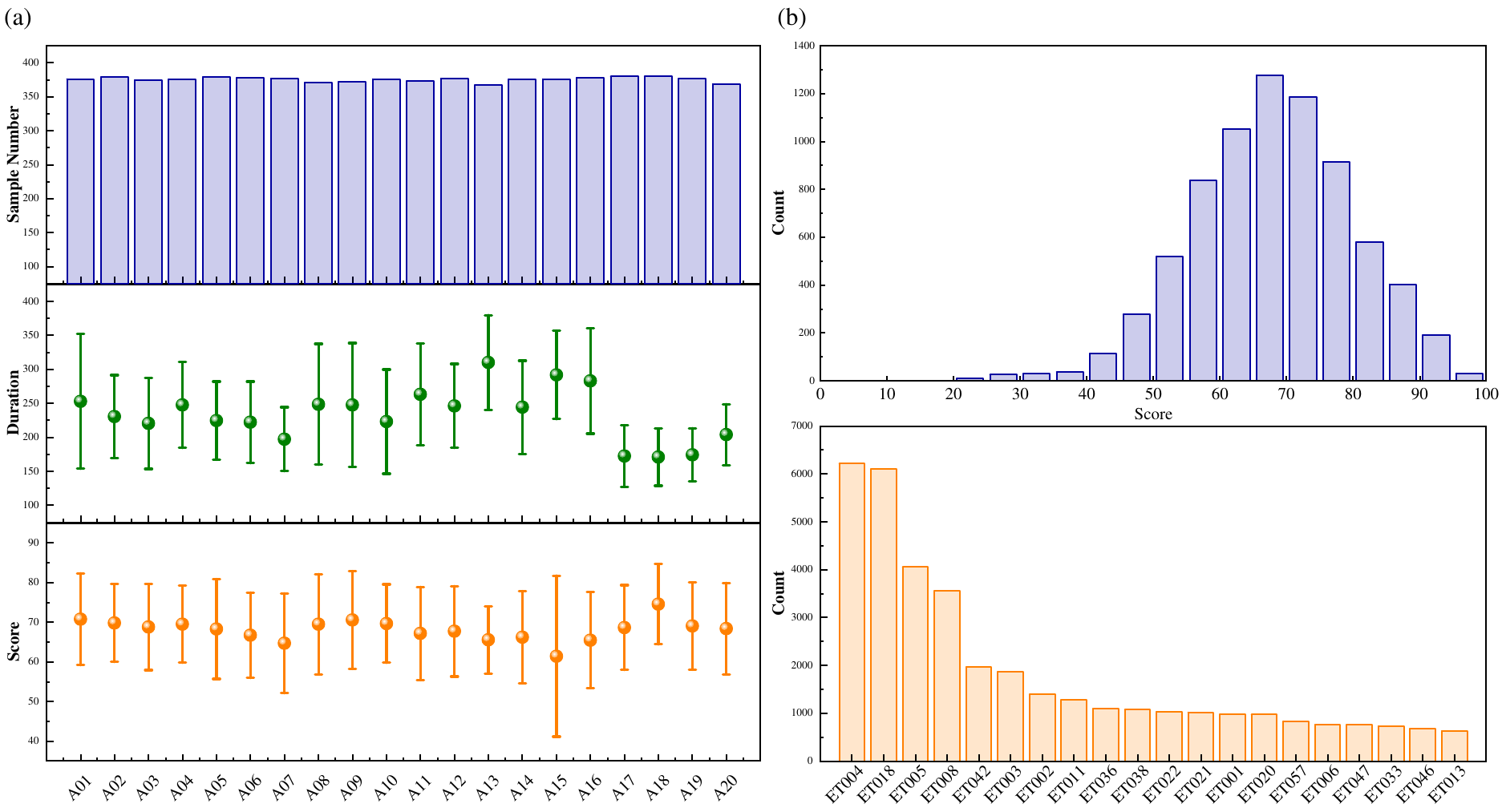}
    \caption{\textbf{Data diversity and distribution in MyoMechanix.} \textbf{(a)} Number of samples, average duration, and scores for the 20 actions. \textbf{(b)} Overall score distribution of the dataset and the 20 most frequent error types.}
    \label{fig_dataset_static_fa}
\end{figure}

\begin{figure}
    \centering
    \includegraphics[width=\linewidth]{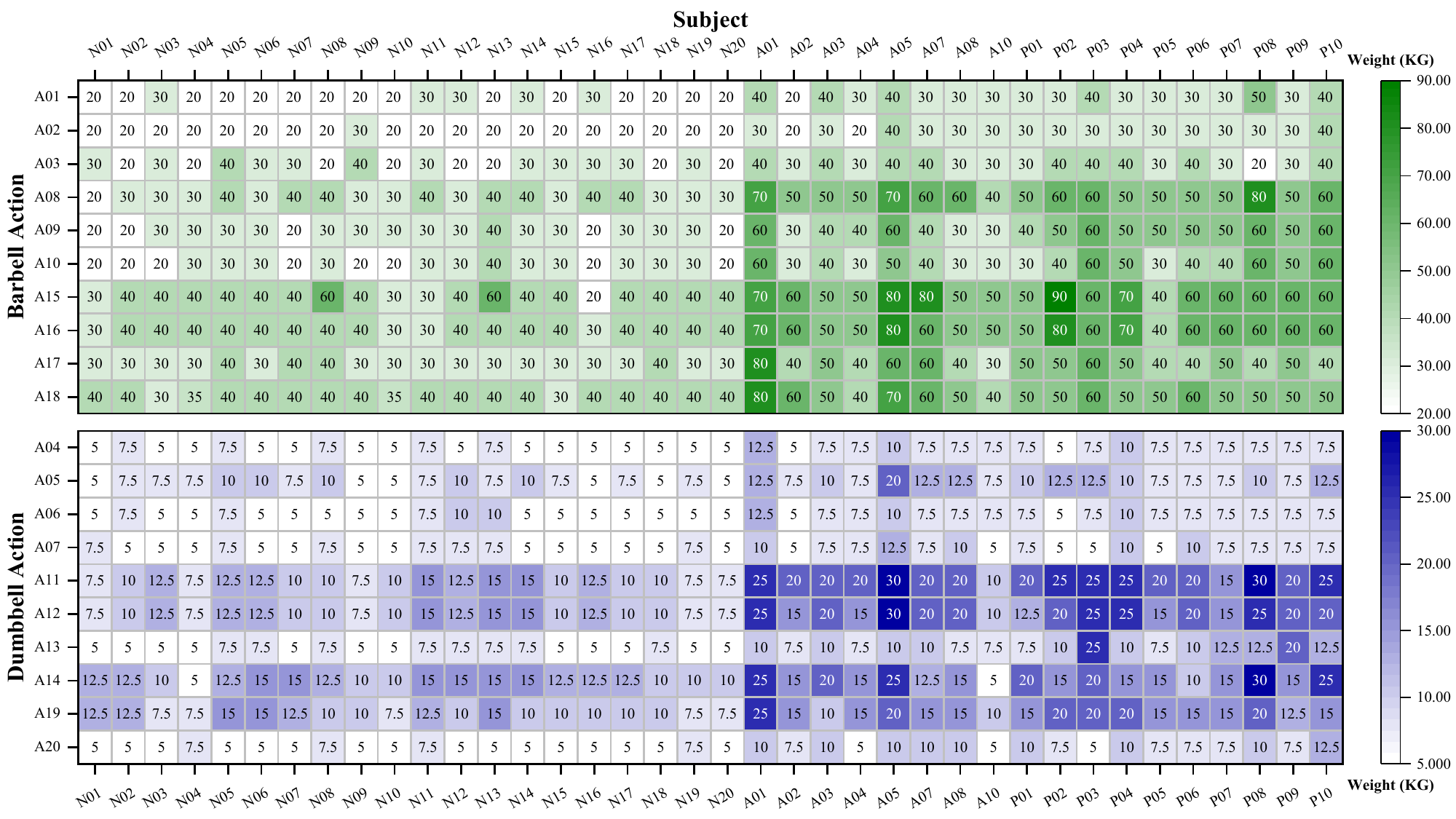}
    \caption{\textbf{Distribution of weight used by subjects across actions.} 
    MyoMechanix includes 20 weight-loaded fitness actions, evenly split between barbell and dumbbell exercises. For barbell actions, the intrinsic weight of the barbell bar, 20 kg, is included in the calculation. For dumbbell actions, the reported value corresponds to the weight of a single dumbbell, with the total bilateral load being twice that value. In the figure, the X-axis represents subjects and the Y-axis represents fitness actions. Darker colors indicate heavier weights.}
    \label{fig_dataset_static_weight}
\end{figure}

\begin{figure}
    \centering
    \includegraphics[width=\linewidth]{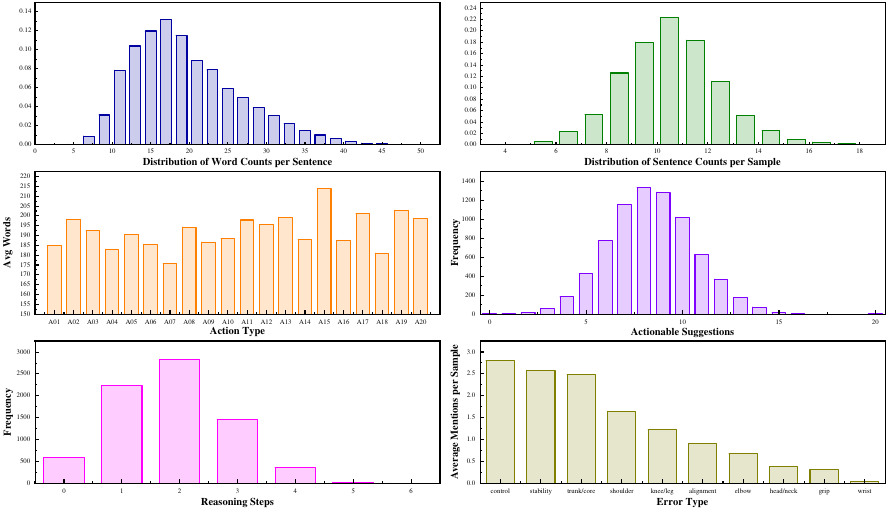}
    \caption{\textbf{Linguistic complexity and fine-grained reasoning in MyoMechanix-VideoQA.} 
    This figure characterizes the linguistic richness of the dataset through six distributions: \textbf{(1)} words per sentence, \textbf{(2)} sentences per sample, \textbf{(3)} average word counts across the 20 action categories, \textbf{(4)} frequency of actionable suggestions, \textbf{(5)} number of reasoning steps, and \textbf{(6)} average frequency of error-type keywords per sample.}
    \label{fig_dataset_static_tf}
\end{figure}

\subsubsection{MyoMechanix-Video2EMG Dataset}
\label{Sec343-Video2EMG_Dataset}
Although visual modalities, such as RGB video and 3D poses, effectively capture the macroscopic kinematic features of human action, they inherently lack information regarding the internal activation of muscles, the exertion of force, and physiological cost. To bridge this information gap and expand the boundaries of research in multimodal action analysis, we propose Video2EMG, a novel cross-modal translation benchmark. 

Focusing on the direct prediction of continuous sEMG signals from video, this benchmark leverages the strictly paired visual and physiological signals within the dataset. This approach facilitates the modeling of the complex, nonlinear mapping between external representations of motion and internal biomechanical states. This task offers significant practical value in scenarios where the deployment of wearable sensors is constrained, such as competitive sports and rehabilitation therapy, as it enables unobtrusive assessment of target muscle engagement and fatigue.

\subsection{Dataset Statistics}
\label{Sec35-Statistic}

\begin{table*}[h]
\centering
\caption{\textbf{Comparison of textual features between CoT-AFA and MyoMechanix-VideoQA (MyoMech-VQA).}}
\label{table_data_static_tf}
\begin{tabularx}{\textwidth}{@{}>{\raggedright\arraybackslash}p{4.4cm}>{\raggedright\arraybackslash}p{4.2cm}>{\centering\arraybackslash}p{2.3cm}>{\centering\arraybackslash}p{2.2cm}>{\centering\arraybackslash}X@{}}
    \toprule [1.5pt]
    \textbf{Metric Category} & \textbf{Metric} & \textbf{Statistic} & \textbf{CoT-AFA \cite{qi2025explainable}} & \textbf{MyoMech-VQA} \\ \midrule [1.0pt]
    \multirow{3}{*}{Corpus Volume} 
    & Average Words per Sample      & Word Count      & 102.19 & \textbf{192.27} \\
    & Std of Words per Sample       & Std Dev         & --     & \textbf{29.09} \\
    & Average Sentences per Sample  & Sentence Count  & 5.25   & \textbf{10.05} \\\midrule [1.0pt]
    \multirow{5}{*}{Lexical \& Syntactic Complexity} 
    & Total Vocabulary Size         & Token Count     & 3143   & \textbf{2008}\\
    & Type-Token Ratio              & Ratio           & --     & \textbf{0.0014}\\
    & Avg Sentence Length           & Words           & --     & \textbf{19.14}  \\
    & Flesch-Kincaid Grade Level    & Grade           & --     & \textbf{13.78}\\
    & Flesch Reading Ease           & Score           & --     & \textbf{30.35}\\\midrule [1.0pt]
    \multirow{2}{*}{Reasoning \& Actionability}    
    & Avg Reasoning Steps           & Per Sample      & 0.91   & \textbf{1.85} \\
    & Avg Actionable Suggestions    & Per Sample      & 0.75   & \textbf{8.41} \\
    \bottomrule [1.5pt]
\end{tabularx}
\end{table*}

\subsubsection{Fundamental Attributes}

After data cleaning and annotation verification, the MyoMechanix dataset contains 7,512 action samples collected from 38 subjects across three ability levels, covering 20 distinct weight-loaded fitness actions. We conducted statistical analyses of the per-action sample counts, sample durations, and action quality scores, as shown in \autoref{fig_dataset_static_fa}. The sample counts are well balanced across actions, ranging from 370 to 380. The average clip duration is 233.76 frames, with individual clips spanning 100 to 1000 frames. Action quality scores have a mean of 68.15 and exhibit an approximately normal distribution, with values ranging from 20 to 100. This broad score range, together with the moderate mean, indicates substantial diversity in skill levels within the dataset. We also report the 20 most frequent error types. Finally, \autoref{fig_dataset_static_weight} presents the specific weight loads used by each subject.

\subsubsection{Linguistic, Semantic and Reasoning Complexity}

In addition to quantitative statistics on dataset scale, video duration, and action scores, we conduct a systematic analysis of the structural and distributional characteristics of the MyoMechanix dataset across textual features, as summarized in \autoref{table_data_static_tf} and \autoref{fig_dataset_static_tf}. This analysis helps characterize the linguistic complexity of the dataset and the challenges it poses for multimodal reasoning and fine-grained action understanding.

In terms of \textbf{corpus volume}, we examine the word count and sentence count of the textual annotations. All text is converted to lowercase, and standard punctuation marks are removed using regular expressions before tokenization based on whitespace. The results show that the average word count across all 7,512 samples is 192.27, with a standard deviation of 29.09. In addition, segmenting the text using terminal punctuation marks, including periods, question marks, and exclamation points, shows that each sample contains an average of 10.05 sentences.

In terms of \textbf{lexical \& syntactic complexity}, we evaluate vocabulary richness and sentence-level structural characteristics. Extracting unique tokens across all samples yields a total vocabulary size of 2,008 for the entire dataset.

The resulting Type-Token Ratio (TTR) is 0.0014. Since TTR is highly sensitive to corpus size, this low value should be interpreted in conjunction with the dataset's large total word count. Together, these statistics suggest a thematically concentrated vocabulary centered on the specialized domain of fitness and exercise. Furthermore, the average sentence length across samples is 19.14 words. This relatively long sentence length reflects the dataset's tendency toward detailed descriptive expressions and may indicate increased syntactic and informational complexity.

To further characterize reading difficulty, we compute the Flesch--Kincaid Grade Level (FK Grade) and the Flesch Reading Ease (FRE) score. Both metrics are derived from weighted counts of words, sentences, and syllables. The dataset obtains an FK Grade of 13.78, corresponding approximately to the reading level of a first-year college student in the United States. Its FRE score is 30.35 on a standard scale from 0 to 100, where lower values indicate greater reading difficulty, placing the text in the ``difficult'' category. Taken together, these two metrics suggest that the textual annotations in MyoMechanix are linguistically demanding and require a relatively high level of reading comprehension.

For \textbf{reasoning \& actionability}, we adopt a keyword-matching approach to characterize the presence of causal reasoning and actionable guidance within the Chain-of-Thought (CoT) annotations. This strategy is used instead of more complex semantic analysis because the evaluation texts exhibit highly consistent expression patterns. Causal explanations frequently rely on explicit conjunctions, while actionable suggestions commonly employ imperative constructions or modal verbs. Accordingly, this keyword-based methodology provides a simple, efficient, and reproducible proxy for analyzing reasoning- and action-oriented textual patterns.

For the statistical analysis of reasoning-related expressions, we retrieve and count causal, inferential, and explanatory keywords, including ``because,'' ``due to,'' ``thus,'' ``hence,'' ``indicating,'' and ``suggesting.'' The results show that each sample contains an average of 1.85 explicit reasoning steps. These lexical cues suggest that the annotations often extend beyond surface-level description to include deeper causal reasoning. For example, the phrase ``which reduces'' in the sentence ``grip was open rather than closed, which reduces control'' explicitly links an observed movement pattern to its functional consequence, forming a causal explanatory relation.

For the statistical analysis of actionability, we calculate the frequency of directive, corrective, and effect-oriented keywords, including ``adjust,'' ``aim for,'' ``to correct,'' ``avoid,'' ``will enhance,'' and ``will reduce.'' The results show that each sample contains an average of 8.41 actionable recommendations grounded in the specific erroneous movement patterns observed in that sample. These cues indicate that the annotations provide concrete advice for performance improvement. An example of such executable guidance is the instruction to ``Keep the head firmly against the bench throughout...''.

Collectively, the dataset's causal reasoning content and actionable recommendations highlight the value of MyoMechanix for developing and evaluating models that can identify erroneous movement patterns, reason about their consequences, and generate precise, sample-grounded feedback for improvement. Overall, the dataset exhibits notable strengths across multiple dimensions, including textual scale, structural complexity, and semantic depth. These characteristics make MyoMechanix a more demanding benchmark for assessing the fine-grained multimodal understanding and reasoning capabilities of existing models.
\section{Our CUBIST Modeling Paradigm}
\label{Sec4-Model}

\begin{figure*}[ht]
    \centering
    \includegraphics[width=\linewidth]{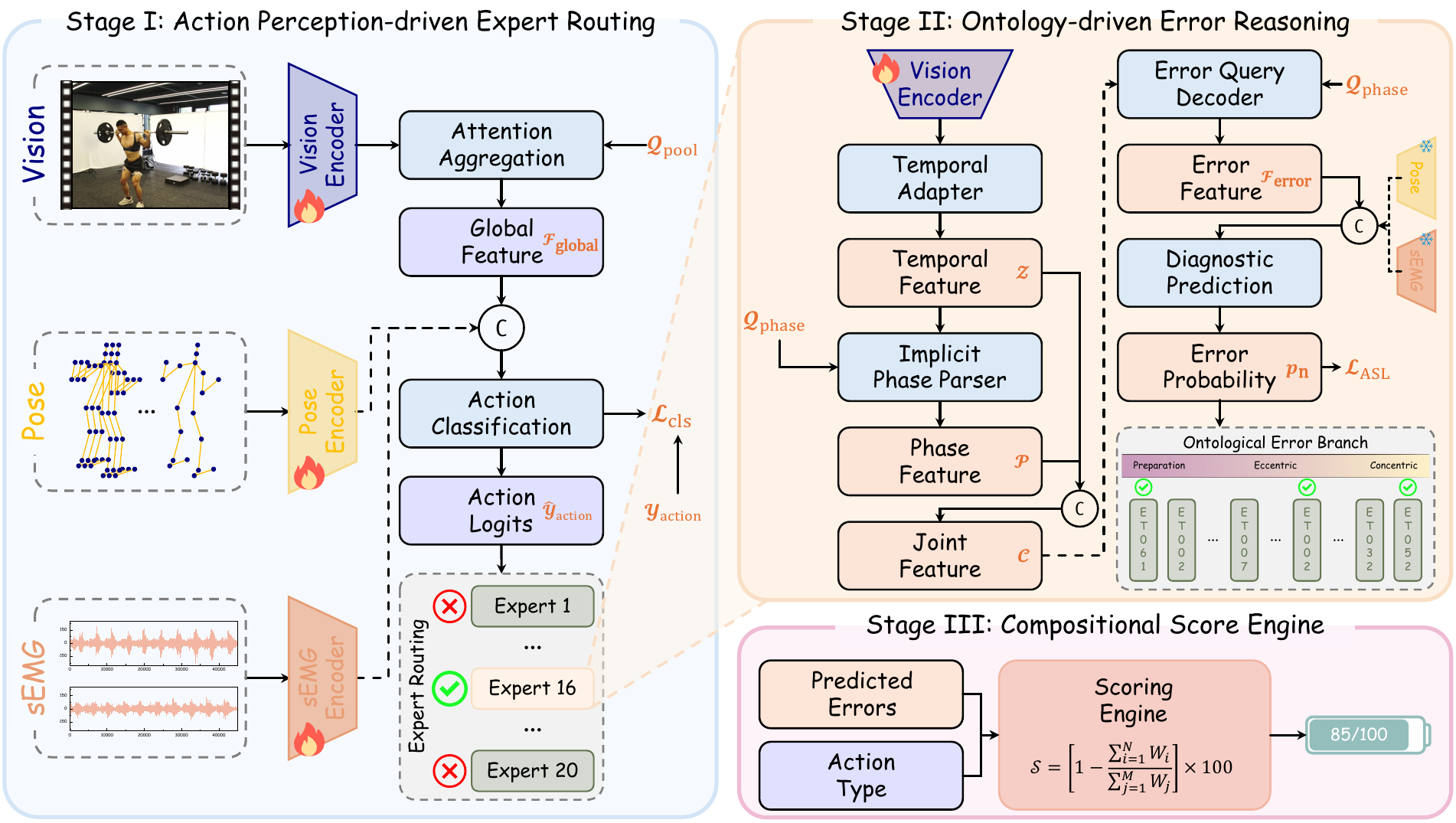}
    \caption{\textbf{Illustration of proposed CUBIST models' architectures (multimodal version) for the AQA task.}}
    \label{fig_mod_str}
\end{figure*}

Existing AQA paradigms are largely monolithic black-box models: they produce a single overall score, offering limited fine-grained error attribution or explanation. In this sense, they resemble Impressionism in art history, which emphasizes an overall impression over detailed structure. By contrast, Cubism offers a more analytical perspective: it decomposes complex objects into smaller components, examines each part from multiple viewpoints, and recombines them to reveal underlying structure and relationships.

Motivated by these principles, we propose CUBIST (Compositional Ontological Reasoning Engine), a novel modeling paradigm for compositional reasoning over human actions that leverages rich MyoMechanix-FKG and multimodal data. CUBIST moves beyond the currently prevalent single overall score paradigm toward interpretable, step-wise reasoning that closely reflects expert analysis.

CUBIST can be understood as a language for analyzing human actions, providing structured, high-resolution representations that support detailed modeling of movement, technique, and coaching processes.

Specifically, human actions are first identified and routed to activity-specific expert pathways (Stage I). Guided by the resulting activity prior and the underlying ontology, the model then performs fine-grained, phase-aware activity decomposition and error reasoning over the action sequence (Stage II). These interpretable error probabilities are subsequently synthesized through an explicit compositional rule-based scoring engine to produce the final action quality assessment report and its associated coaching interpretation (Stage III). The full CUBIST model is visualized in \autoref{fig_mod_str}.

This progressive and compositional design is intended to reflect aspects of expert reasoning and refereeing practice, where judgments are formed by breaking down complex actions, inspecting them from multiple viewpoints, and synthesizing the resulting evidence. By grounding model outputs in an underlying ontology, CUBIST enables fine-grained attribution of both intermediate reasoning and final assessments.

As a result, CUBIST provides a framework for improving interpretability in Action Quality Assessment (AQA), supporting transparency at both the reasoning and output levels. Now, in the following, we present each stage in detail.

\subsection{Stage I---Action Perception-driven Expert Routing}

In human expert action quality assessment, accurate identification of the action type constitutes the initial step of the evaluation process. Similarly, Stage I of the CUBIST model functions as a high-precision routing module that maps each input video to the appropriate action-specific expert network according to its predicted action category. This stage is built upon a spatiotemporal video encoder $\phi$, which serves as the backbone of the CUBIST model. The encoder supports reliable action-category recognition while establishing a robust feature representation for the subsequent stages of fine-grained error reasoning.

Specifically, we define the input video sequence as $\mathcal{V} \in \mathbb{R}^{B \times T \times C \times H \times W}$, where $B$ denotes the batch size, $T$ is the number of sampled frames, $C$ is the number of input channels, and $H{=}W$ denotes the spatial resolution. The backbone partitions the input video into non-overlapping spatiotemporal tubelets with temporal size $t_s{=}2$ and spatial patch size $p{=}16$. After tubelet partitioning and linear projection, the video is encoded into a one-dimensional token sequence:
\begin{equation}
\mathcal{H} = \phi(\mathcal{V}) \in \mathbb{R}^{B \times L \times D},
\end{equation}
where $L = \lfloor T / t_s \rfloor \times (H/p)^2$ is the total number of tokens, and $D$ is the embedding dimension.

Since the 3D positional encoding of the pre-trained backbone is originally defined for a 16-frame input, corresponding to a positional grid of $(8, 14, 14)$, CUBIST adopts trilinear interpolation to extend the positional encoding to $(51, 14, 14)$. This adaptation enables the backbone to process long video sequences of up to 103 frames while preserving the semantic priors encoded in the pre-trained weights as much as possible.

To aggregate the high-dimensional token sequence $\mathcal{H}$ into a compact global representation, we introduce an attention-based pooling mechanism with a learnable query. Specifically, we define a learnable query vector $\mathcal{Q}_{\text{pool}} \in \mathbb{R}^{1 \times D}$, which attends to the entire token sequence through multi-head attention:
\begin{equation}
\mathcal{F}_{\text{global}} 
= \text{Squeeze}\left(
\text{LN}\left(
\text{MHA}(\mathcal{Q}_{\text{pool}},\, \mathcal{H},\, \mathcal{H})
\right)
\right)
\in \mathbb{R}^{B \times D},
\end{equation}
where $\text{MHA}(\cdot)$ denotes an 8-head multi-head attention module, $\text{LN}(\cdot)$ denotes layer normalization, and $\text{Squeeze}(\cdot)$ removes the singleton query dimension. Compared with simple mean pooling, this attention-based pooling mechanism can adaptively assign higher aggregation weights to more informative spatiotemporal regions.

The resulting global feature $\mathcal{F}_{\text{global}}$ is then passed through a lightweight multi-layer perceptron classification head to produce action-category logits:
\begin{equation}
\hat{\mathcal{Y}}_{\text{action}} 
= \text{MLP}_{\text{cls}}(\mathcal{F}_{\text{global}}) 
\in \mathbb{R}^{B \times C_{\text{action}}},
\end{equation}
where $C_{\text{action}}$ denotes the number of action categories.

During optimization, Stage I is trained using a cross-entropy loss with label smoothing:
\begin{equation}
\mathcal{L}_{\text{cls}} 
= \text{CE}(\hat{\mathcal{Y}}_{\text{action}},\, 
\mathcal{Y}_{\text{action}};\, \epsilon{=}0.1),
\end{equation}
where $\epsilon$ is the smoothing coefficient. Label smoothing mitigates over-confidence on hard training labels and improves the generalization ability of the action classification module.

\subsection{Stage II -- Ontology-driven Error Reasoning}

Drawing on the refined analysis logic of human experts after establishing the action category, the CUBIST model uses the action category output from Stage I as a heuristic prior. It feeds the video spatiotemporal features into the ontology-driven reasoning module to decode the fine-grained error types within the action sequentially. Since the fitness knowledge graph defines highly heterogeneous error-evaluation branches, we integrate the mixture-of-experts (MoE) architecture with the ontology-driven concept. Specifically, this stage upgrades the traditional feed-forward neural network into a specific mixture of experts architecture. It instantiates and maintains an exclusive expert network module $\mathcal{E}_a$ for each action category. This module predicts and outputs the $\mathcal{N}_{\text{a}}$ dimensional binary error labels of the corresponding action branch in the fitness knowledge graph.

\subsubsection{Temporal Adapter}

The high-dimensional token sequence $\mathcal{H} \in \mathbb{R}^{B \times L \times D}$ output by the backbone exhibits entangled spatial and temporal information. To effectively recover its native spatiotemporal topology and build a compact pure temporal representation, we design a temporal adapter module. This module first reshapes the input sequence along the spatiotemporal dimensions into a 4D tensor $\mathcal{H}_{\text{grid}} \in \mathbb{R}^{B \times T_{\text{orig}} \times S \times D}$, where $T_{\text{orig}} = \lfloor T/t_s \rfloor$ represents the original number of time steps and $S = (H/p)^2$ represents the number of spatial patches contained in a single time step. Subsequently, the adapter performs a global average pooling operation on the spatial dimension to eliminate intra-frame spatial redundancy, yielding the intermediate feature representation $\tilde{\mathcal{H}}$. Building on this, it utilizes adaptive average pooling to compress the temporal dimension to a target fixed length $T^\prime$, thereby obtaining the initial temporal feature $\mathcal{Z}^{(0)}$.

To strengthen the capacity of this feature to model local temporal dependencies, this module cascades $N_{\text{conv}}$ depthwise separable convolution blocks with residual connections after the pooling layer. A single convolution block sequentially encapsulates depthwise convolution, pointwise convolution, the GELU activation function, and layer normalization to ensure efficient gradient flow. The specific feature evolution process operates as follows:
\begin{multline}
\mathcal{Z}^{(i+1)} = \text{LN}\!\left(\mathcal{Z}^{(i)} + \text{GELU}\!\left(\text{PW}\!\left(\text{DW}\!\left(\mathcal{Z}^{(i)}\right)\right)\right)\right), \\\quad i = 0, \ldots, N_{\text{conv}} - 1
\end{multline}
where $\text{DW}(\cdot)$ and $\text{PW}(\cdot)$ denote the depthwise separable convolution and pointwise convolution operations, respectively. After the aforementioned multi-level temporal modeling, the network outputs the final temporal representation denoted as $\mathcal{Z} = \mathcal{Z}^{(N_{\text{conv}})} \in \mathbb{R}^{B \times T^\prime  \times D}$.

\subsubsection{Implicit Phase Parser}

From the kinematic perspective, complex sports actions naturally consist of several continuous phases. To enable the network to autonomously discover and capture such temporal structures without manual phase boundary annotations, we introduce an implicit phase parser module. This module defines $K$ independent and learnable phase query vectors $\mathcal{Q}_{\text{phase}} \in \mathbb{R}^{K \times D}$. Then, it utilizes a cross-attention mechanism to perform selective focusing on the temporal features $\mathcal{Z}$, achieving data-driven semantic-level temporal segmentation:
\begin{equation}
\mathcal{P}_{\text{raw}},\, \mathcal{W}_{\text{phase}} = \text{MHA}(\mathcal{Q}_{\text{phase}},\, \mathcal{Z},\, \mathcal{Z})
\end{equation}
where $\mathcal{P}_{\text{raw}} \in \mathbb{R}^{B \times K \times D}$ represents the aggregated raw phase representation, and $\mathcal{W}_{\text{phase}} \in \mathbb{R}^{B \times K \times T^\prime}$ is the cross-attention weight matrix.

To further strengthen the representational capacity of the features, the raw phase representation passes through a residual connection and enters a feed-forward neural network for nonlinear refinement:
\begin{equation}
\mathcal{P} = \text{LN}\!\left(\hat{\mathcal{P}} + \text{FFN}(\hat{\mathcal{P}})\right), \quad \hat{\mathcal{P}} = \text{LN}\!\left(\mathcal{P}_{\text{raw}} + \mathcal{Q}_{\text{phase}}\right)
\end{equation}

Through this cascaded computation, the final phase feature $\mathcal{P} \in \mathbb{R}^{B \times K \times D}$ encodes the deep semantic representations of the $K$ implicit motion phases on the complete action timeline.

\subsubsection{Phase-Aware Error Query Decoder}

In the implementation of Stage II, the error query decoder instantiates an independent learnable query vector $\mathcal{Q}_{\text{err}} \in \mathbb{R}^{N_a \times D}$ for each category of fine-grained action error requiring detection. Considering the occurrence mechanism of action execution errors, their judgment relies heavily on the continuous dynamic temporal flow evolution and remains constrained by explicit motion phase semantic boundaries. Based on this physical prior, we concatenate the temporal feature $\mathcal{Z}$ and phase feature $\mathcal{P}$ along the sequence dimension. This creates a joint context matrix $\mathcal{C} = [\mathcal{Z};\, \mathcal{P}] \in \mathbb{R}^{B \times (T^\prime + K) \times D}$ that possesses phase-aware capabilities. Subsequently, each error query vector uses a cross-attention mechanism to perform adaptive retrieval on this joint context. This process autonomously anchors the temporal intervals and phase semantics that are highly correlated with the corresponding error type, finally outputting the decoded error feature representation $\mathcal{F}_{\text{error}} \in \mathbb{R}^{B \times N_a \times D}$.

Then $\mathcal{F}_{\text{error}}$ enters a feed-forward neural network with a residual connection for nonlinear refinement. A linear projection layer maps it into scalar predicted logits for various types of errors. The sigmoid activation function transforms these logits into final predicted probabilities:
\begin{multline}
p_n = \sigma\!\left(\mathcal{W}_{\text{head}}^{\top} \cdot \text{LN}\!\left(\mathcal{F}_{\text{error}} + \text{FFN}(\mathcal{F}_{\text{error}})\right)\right), \\\quad n = 1, \ldots, N_a
\end{multline}
where $\sigma(\cdot)$ denotes the sigmoid activation function and $\mathcal{W}_{\text{head}} \in \mathbb{R}^{D}$ denotes the linear projection weight.

For network optimization, since the fine-grained error detection is inherently a multi-label classification task with high-dimensional, sparse positive examples. Therefore, Stage II introduces an asymmetric loss function to guide gradient backpropagation:
\begin{multline}
\mathcal{L}_{\text{ASL}} = -\frac{1}{N_a}\sum_{n}^{N_a}\bigg[ y_n (1-p_n)^{\gamma^+}\!\log p_n \\
+ (1-y_n)\hat{p}_n^{\,\gamma^-}\!\log(1-\hat{p}_n)\bigg]
\end{multline}
where $\hat{p}_n = \max(p_n - m, 0)$ represents the predicted probability of negative samples after introducing a probability margin offset mechanism and $m$ is the truncation margin parameter. The terms $\gamma^+$ and $\gamma^-$ control the gradient focusing weights of the model on positive and negative samples, respectively. This asymmetric penalty mechanism effectively suppresses the absolute dominant role of massive simple negative samples on the global optimization gradient, significantly improving the detection performance and generalization robustness of the model for rare motion errors.

\subsection{Stage III -- Compositional Scoring Engine}

After completing fine-grained error analysis, human experts synthesize the identified errors into a final action-quality score. As the concluding component of the evaluation framework, Stage III emulates this scoring process through a deterministic neuro-symbolic module. Rather than learning a direct correlational mapping from visual features to quality scores, Stage III derives the final score through an explicit scoring rule grounded in the annotation ontology. Specifically, it receives the multi-dimensional fine-grained error probability distribution predicted by Stage II and applies the annotation-defined error weights (see \autoref{Sec33-Annotation}) together with the scoring formula in \autoref{equa_score}. Through this compositional rule-based transformation, high-dimensional error probability representations are mapped into a final quantitative score for action quality. This stage introduces no additional learnable parameters and does not require gradient-based optimization. Identical Stage II error probabilities therefore produce the same score under the fixed scoring rule. As a result, the final score is directly attributable to interpretable error concepts and their corresponding annotation-defined penalties, improving transparency at the output level.

\subsection{Multimodal Fusion}

To fully exploit the heterogeneous information collected in MyoMechanix, CUBIST integrates RGB video, 3D pose, and sEMG through a modular late-fusion strategy. Direct end-to-end optimization of heterogeneous modalities may lead to cross-modal gradient interference, potentially hindering the learning of modality-specific discriminative representations. In our pilot studies, we observed suboptimal performance that may be associated with this issue. To mitigate it, we introduce a feature-disentangled late-fusion design that allows each modality to develop task-relevant representations before their integration, while limiting additional architectural and optimization complexity. The fusion procedure is organized as follows.

\begin{itemize}
\item For \textbf{action recognition}, modality-specific encoders process the 3D pose and sEMG inputs to produce a compact non-visual multimodal feature, denoted as $\mathcal{F}_{\text{multi}}$. In parallel, the RGB stream generates a global representation $\mathcal{F}_{\text{global}}$. These modality-specific features are concatenated along the channel dimension to form a joint action representation, which is then passed to the multi-layer perceptron classification head, $\text{MLP}_{\text{cls}}$, to estimate the action-category probability $\hat{\mathcal{Y}}_{\text{action}}$.

\item For \textbf{fine-grained error detection}, multimodal fusion is performed after task-specific feature abstraction has been established. The Phase-Aware Error Query Decoder first derives an error-oriented representation $\mathcal{F}_{\text{error}}$ from the RGB stream by allowing the query vector $\mathcal{Q}_{\text{err}}$ to attend to the temporal features $\mathcal{Z}$ and phase features $\mathcal{P}$. This representation is then integrated with the synchronously extracted 3D pose and sEMG features to construct a joint error representation. The fused feature is passed through a feed-forward neural network with a residual connection, followed by the final linear prediction head $\mathcal{W}_{\text{head}}$, to estimate the probability of error occurrence.
\end{itemize}

This late-fusion mechanism avoids prematurely forcing heterogeneous modalities into a shared representation space, thereby helping to preserve their modality-specific discriminative characteristics and reducing the risk of cross-modal gradient interference. It also enables complementary reasoning across modalities. For example, RGB video captures appearance and movement context, 3D pose characterizes the spatial configuration of the body, and sEMG reflects underlying muscle activation dynamics. Their integration is particularly valuable near ambiguous action boundaries or for subtle erroneous exertion patterns that may not be fully characterized by any single modality alone. As a result, the proposed strategy leverages the complementary strengths of all three modalities while requiring only minimal architectural modification.

\subsection{Cross-Action Shared Skill Modeling for Expert Adaptation}

MyoMechanix is a large-scale dataset comprising diverse yet related action categories. This structure provides a valuable opportunity to learn skill representations that are shared across actions and transferable to downstream expert models. To capitalize on this property, we adopt a cross-action representation learning strategy that first captures robust and transferable spatiotemporal cues of action quality/skills, and then adapts them to action-specific expert models.
In the first phase, a shared backbone network is trained jointly across all action categories to learn spatiotemporal skill representations that generalize across actions. In the second phase, action-specific expert models are trained independently. To accelerate convergence and preserve the transferable knowledge acquired in the first phase, each expert backbone is initialized with the weights of the converged shared model and optimized using a progressive unfreezing strategy. Specifically, the backbone parameters are initially frozen, and only the newly introduced downstream modules are trained, including the temporal adapter, implicit phase parser, and error query decoder. After a predefined warm-up period, several upper Transformer blocks of the backbone are unfrozen and jointly fine-tuned with the downstream modules using a reduced learning rate. Both training phases use the AdamW optimizer together with a cosine-annealing learning rate scheduler.
\section{Experiments}
\label{Sec5-Exp}

To systematically validate the central claims of this work, we conduct comprehensive experiments on the three benchmark tasks introduced in \autoref{Sec34-Design}. These evaluations are designed not merely to report task-specific performance, but to examine whether the MyoMechanix ecosystem enables a richer and more challenging form of action understanding than conventional vision-centric benchmarks. In particular, we investigate three complementary questions: 
(i) whether multimodal physiological sensing improves the assessment of action quality beyond external motion cues alone; 
(ii) whether structured annotations support fine-grained, semantically grounded reasoning about execution errors and corrective feedback; and 
(iii) whether latent neuromuscular dynamics can be inferred, at least partially, from observable visual motion.

The remainder of this section is organized as follows:
\begin{itemize}
    \item First, we benchmark \textbf{Action Quality Assessment} under both standard and disjoint splits, evaluating existing methods and quantifying the contribution of multimodal fusion (\autoref{Sec51-AQA});
    \item Next, we assess structured multimodal reasoning through the \textbf{Video Question Answering} benchmark, which targets action recognition, fine-grained error diagnosis, and corrective feedback generation (\autoref{Sec52-AQU});
    \item Finally, we investigate cross-modal physiological inference through the \textbf{novel Video2EMG task}, examining the feasibility of predicting muscle activation patterns directly from visual observations (\autoref{Sec53-V2E}).
\end{itemize}

\subsection{AQA \& Coaching}
\label{Sec51-AQA}

As the cornerstone task within the MyoMechanix ecosystem, establishing performance benchmarks for AQA is paramount. To comprehensively measure the inherent challenges of the dataset and thoroughly analyze the representational capabilities of each modality, we systematically constructed a comprehensive evaluation framework that includes unimodal baseline experiments, multimodal fusion experiments, and diverse data splitting protocols. We also compare our CUBIST model against SOTA models.

\subsubsection{Baseline methods}
\noindent\textbf{Vision Branch.} Considering that vision features serve as the paramount core modality in the AQA field, we adhered to previously outlined research trends when establishing benchmarks. We selected 7 representative SOTA heterogeneous network architectures—including I3D-MLP \cite{tang2020uncertainty}, CoRe \cite{yu2021group}, TPT \cite{bai2022action}, HGCN \cite{zhou2023hierarchical}, MCoRe \cite{an2024multi}, T2CR \cite{ke2024two},  and DAE \cite{zhang2024auto} -- to ensure the comprehensive and objective evaluation. More information on these models has been provided in the Appendix.\\

\noindent\textbf{Pose Branch.} Apart from visual features, skeletal pose—as a modality that accurately captures human kinematic trajectories—plays a significant role in AQA. Therefore, we selected and incorporated multiple SOTA representative models -- including MS-G3D \cite{liu2020disentangling}, CTR-GCN \cite{chen2021channel}, ST-GCN++ \cite{duan2022pyskl}, and SkateFormer \cite{do2024skateformer} -- as evaluation baselines to systematically assess the gain value of this modality within the MyoMechanix-AQA. More information on these models has been provided in the Appendix.\\

\noindent\textbf{EMG Branch}
For the EMG modality, we designed a specialized branch to process continuous natural signals. To mitigate the inter-subject variance inherent in raw physiological data, a comparative model is incorporated to explicitly compute the relative activation contributions of distinct muscles during the execution of an action. This procedure distills the raw data into a robust representation of muscle coordination. Subsequently, an MLP is utilized to project these low-dimensional comparative features into a high-dimensional embedding space, thereby facilitating seamless integration with macroscopic visual or skeletal features.

\subsubsection{Implementation Details}
For data preparation, we segment each full action video into single-action clips based on annotated keyframes, uniformly sampling 103 frames per clip. Given the strict synchronization across modalities in the MyoMechanix dataset, skeletal and sEMG signals are cropped using the identical keyframe indices: each skeleton sample is formatted into a 103-frame sequence, while the sEMG data retains its native temporal resolution. For all multimodal baselines, we adopt a late-fusion strategy, similar to our CUBIST model (\autoref{Sec34-Design}), ensuring fairness. Each modality is processed by its respective independent branch (e.g., I3D for RGB, SkateFormer for skeleton), and the extracted representations are integrated strictly at the feature level prior to the final prediction head. All experiments were conducted on multiple NVIDIA RTX 4090 GPUs, totaling approximately 4000 GPU-hours. Further details provided in the Appendix. All the codes will be publicly released.

\subsubsection{Performance Metrics}
To maintain consistency with prior research in the AQA field, we utilize the Spearman Rank Correlation ($\rho$) and the Relative L2-Distance ($R-l2$) as the primary evaluation metrics. 

The $\rho$ metric, originally introduced by Pirsiavash et al. \cite{pirsiavash2014assessing}, is the most widely adopted performance indicator within the field of action quality assessment. This metric effectively quantifies the strength and direction of the monotonic relationship between the ground-truth and predicted rankings; however, it does not capture the absolute differences between their respective scores. The formula for $\rho$ is defined as follows:
\begin{equation}
    \rho=1-\frac{6\sum_{i=1}^n(R_i-\widehat{R_i})^2}{n(n^2-1)}
\end{equation}
where $R_i$ and $\widehat{R}_i$ represent the ground truth and predicted rankings of the $i$-th sample, respectively, and $n$ denotes the total number of samples. The value of $\rho$ ranges within $[-1, 1]$, with values approaching $1$ indicating superior predictive performance.

To address this limitation, Yu et al. \cite{yu2021group} proposed the Relative L2-Distance ($R-l2$), a metric based on the L2 norm that, unlike $\rho$, emphasizes the numerical discrepancy between the ground-truth and predicted scores. Furthermore, normalizing the score ranges across different categories of actions, $R-l2$ facilitates cross-category training in a manner that the traditional L2 distance cannot. The formula $R-l2$ is defined as follows:
\begin{equation}
    R-\ell_2=\frac{1}{n}\sum_{i=1}^n(\frac{\vert s_i-\widehat{s_i} \vert}{s_{max}-s_{min}})^2
\end{equation}
where $s_i$ and $\widehat{s}_i$ represent the ground truth and predicted scores of the $i$-th sample, respectively. The variables $s_{max}$ and $s_{min}$ represent the maximum and minimum scores within the respective category of action, and $n$ represents the total number of samples. The value of $R-l2$ ranges within $[0, 1]$, with values approaching $0$ indicating superior predictive performance.

\subsubsection{Results}

All the results are summarized in \autoref{table_exp_aqa_avg}. Performance on individual actions across various splits are reported in \autoref{table_exp_van_detail}, \autoref{table_exp_cs_detail},  \autoref{table_exp_cv_detail}, and \autoref{table_exp_mv_detail}. Qualitative comparative examples are provided in \autoref{table_exp_cubist}.\\ 

\begin{table}[htbp]
\centering
\caption{\textbf{Average performance of various AQA models across various protocols.} $R-\ell_2(\times100)$. Prepended 'B' indicates best model in a particular modality. Prepended S and M indicate single and multiple views, respectively. All the multimodal and multiview models have been developed by us. Notice the improvement as modalities and views are incorporated. CUBIST paradigm achieves the new SOTA.}

\label{table_exp_aqa_avg}
\setlength{\tabcolsep}{3.5pt}

\begin{tabular}{@{}>{\raggedright\arraybackslash}p{1.3cm}>{\raggedright\arraybackslash}p{4.2cm}*{2}{>{\centering\arraybackslash}p{1.1cm}}@{}}
\toprule [1.5pt]
\textbf{Protocol} & \textbf{Model} & \textbf{$\rho \uparrow$} & \textbf{$R-\ell_2 \downarrow$} \\
\midrule [1.0pt]
\multirow{24}{*}{\textbf{Vanilla}}
& I3D-MLP \cite{tang2020uncertainty}   & 0.4837 & 3.6586 \\
& CoRe \cite{yu2021group}              & 0.6856 & 2.2941 \\
& TPT \cite{bai2022action}             & 0.5832 & 3.1595 \\
& HGCN \cite{zhou2023hierarchical}     & 0.7144 & 2.6387 \\
& MCoRe \cite{an2024multi}             & 0.6936 & 2.7761 \\
& T2CR \cite{ke2024two}                & 0.4602 & 3.8081 \\
& DAE \cite{zhang2024auto}             & 0.7431 & 2.4180 \\
& \textbf{Ours CUBIST}                 & \textbf{0.7829} & \textbf{2.1542} \\ \cmidrule{2-4}

& MS-G3D \cite{liu2020disentangling}   & 0.5512 & 3.4048 \\
& CTR-GCN \cite{chen2021channel}       & 0.5873 & 3.1140 \\
& ST-GCN++ \cite{duan2022pyskl}        & 0.6072 & 2.8277 \\
& SkateFormer \cite{do2024skateformer} & 0.6508 & 2.7112 \\ \cmidrule{2-4}

& EMG                                  & 0.2046 & 4.9312 \\ \cmidrule{2-4}

& BSV+BPose                            & 0.7954 & 1.4689 \\
& BSV+EMG                              & 0.7607 & 2.3050 \\
& BPose+EMG                            & 0.6692 & 2.4187 \\
& BSV+BPOSE+EMG                        & 0.8297 & 1.1403 \\
& Ours CUBIST (SV+Pose)                     & 0.8296 & 1.0528 \\
& \makecell[l]{\textbf{Ours CUBIST} \\ \textbf{(SV+Pose+EMG)}}                & \textbf{0.8455} & \textbf{1.0198} \\ \cmidrule{2-4}

& BMV                                  & 0.7861 & 1.8593 \\
& BMV+BPose                            & 0.8498 & 0.9915 \\
& BMV+EMG                              & 0.8016 & 1.1588 \\
& BMV+BPose+EMG                        & 0.8564 & 0.8865 \\ 
& \makecell[l]{\textbf{Ours CUBIST} \\ \textbf{(MV+Pose+EMG)}}                 & \textbf{0.8836} & \textbf{0.7418} \\\midrule[1.0pt]

\multirow{5}{*}{\makecell[l]{\textbf{Cross}\\ \textbf{Subject}}}
& BSV                                  & 0.3484 & 4.1568 \\
& BPose                                & 0.3417 & 4.1753 \\
& EMG                                  & 0.1375 & 7.7435 \\
& BSV+BPose+EMG                        & 0.3853 & 4.0956 \\
& BMV+BPose+EMG                        & \textbf{0.4323} & \textbf{3.9717} \\\midrule[1.0pt]

\multirow{7}{*}{\makecell[l]{\textbf{Cross} \\ \textbf{View}}}
& I3D-MLP \cite{tang2020uncertainty}   & 0.3136 & 4.4885 \\
& CoRe \cite{yu2021group}              & 0.3981 & 4.0688 \\
& TPT \cite{bai2022action}             & 0.3538 & 4.1956 \\
& HGCN \cite{zhou2023hierarchical}     & 0.4372 & 3.9519 \\
& MCoRe \cite{an2024multi}             & 0.4127 & 3.9628 \\
& T2CR \cite{ke2024two}                & 0.2984 & 4.3975 \\
& DAE \cite{zhang2024auto}             & \textbf{0.4663} & \textbf{3.7596} \\
\hdashline
\multirow{7}{*}{\makecell[l]{\textbf{Mix} \\ \textbf{View}}}
& I3D-MLP \cite{tang2020uncertainty}   & 0.3461 & 4.1619 \\
& CoRe \cite{yu2021group}              & 0.4409 & 3.8987 \\
& TPT \cite{bai2022action}             & 0.4089 & 4.0452 \\
& HGCN \cite{zhou2023hierarchical}     & 0.4876 & 3.6371 \\
& MCoRe \cite{an2024multi}             & 0.4593 & 3.8137 \\
& T2CR \cite{ke2024two}                & 0.3158 & 4.2100 \\
& DAE \cite{zhang2024auto}             & \textbf{0.5111} & \textbf{3.5171} \\
\bottomrule [1.5pt]
\end{tabular}
\end{table}

\begin{table*}[htbp]
\centering
\caption{\textbf{Performance of AQA models on each action in the Vanilla split ($\rho \uparrow$).}}
\label{table_exp_van_detail}
\setlength{\tabcolsep}{3.5pt}

\vspace*{0.5em}
\noindent\textbf{Panel A: A01--A10}\par
\vspace*{0.3em}

\begin{tabular}{@{}>{\raggedright\arraybackslash}p{1.5cm}>{\raggedright\arraybackslash}p{3.2cm}*{10}{>{\centering\arraybackslash}p{1cm}}@{}}
\toprule [1.5pt]
\textbf{Modality} & \textbf{Model} & \textbf{A01} & \textbf{A02} & \textbf{A03} & \textbf{A04} & \textbf{A05} & \textbf{A06} & \textbf{A07} & \textbf{A08} & \textbf{A09} & \textbf{A10} \\
\midrule [1.0pt]
\multirow{8}{*}{\textbf{Vision}} 
& I3D-MLP & 0.4382 & 0.3686 & 0.4612 & 0.3893 & 0.5475 & 0.4928 & 0.5038 & 0.5004 & 0.4888 & 0.4532 \\
& CoRe & 0.6662 & 0.5615 & 0.6660 & 0.6213 & 0.7106 & 0.7397 & 0.7392 & 0.7669 & 0.6480 & 0.6498 \\
& TPT & 0.5136 & 0.7467 & 0.5170 & 0.5717 & 0.5505 & 0.7080 & 0.5406 & 0.4851 & 0.5954 & 0.5363 \\
& HGCN & 0.6430 & 0.6766 & 0.7624 & 0.6680 & 0.6216 & 0.6743 & 0.7373 & 0.7436 & 0.6951 & 0.6994 \\
& MCoRe & 0.7035 & 0.6315 & 0.6647 & 0.4431 & 0.7789 & 0.4804 & 0.5750 & 0.9705 & 0.8133 & 0.6891 \\
& T2CR & 0.3255 & 0.4058 & 0.5474 & 0.3832 & 0.5493 & 0.5217 & 0.3748 & 0.5599 & 0.4180 & 0.3778 \\
& DAE & 0.5870 & 0.7096 & 0.7264 & 0.6832 & 0.6828 & 0.7407 & 0.7818 & 0.7802 & 0.7443 & 0.7696 \\
& CUBIST & 0.8112 & 0.7884 & 0.7820 & 0.7875 & 0.7914 & 0.7288 & 0.8375 & 0.8913 & 0.7372 & 0.7720 \\
\cmidrule{2-12}

\multirow{4}{*}{\textbf{Pose}} 
& MS-G3D & 0.3317 & 0.6422 & 0.4745 & 0.4605 & 0.5464 & 0.4689 & 0.5952 & 0.5442 & 0.5384 & 0.4906 \\
& CTR-GCN & 0.3703 & 0.4913 & 0.5656 & 0.4655 & 0.6550 & 0.5569 & 0.5080 & 0.6111 & 0.6640 & 0.5312 \\
& ST-GCN++ & 0.5355 & 0.5309 & 0.5830 & 0.4493 & 0.4700 & 0.5774 & 0.5579 & 0.6272 & 0.5331 & 0.3796 \\
& SkateFormer & 0.6530 & 0.6913 & 0.6126 & 0.5446 & 0.6948 & 0.4915 & 0.4650 & 0.6823 & 0.6754 & 0.5453 \\\cmidrule{2-12}

\multirow{1}{*}{\textbf{EMG}} 
& DAE & 0.1354 & 0.1426 & 0.1599 & 0.1486 & 0.3120 & 0.1865 & 0.2713 & 0.0598 & 0.3522 & 0.2538 \\ \cmidrule{2-12}

\multirow{6}{*}{\makecell[l]{\textbf{Single}\\ \textbf{View}}}
& BSV+BPOSE & 0.6409 & 0.7787 & 0.7948 & 0.7340 & 0.7497 & 0.7826 & 0.8275 & 0.8302 & 0.7913 & 0.8201 \\
& BSV+EMG & 0.6027 & 0.7266 & 0.7462 & 0.6987 & 0.7052 & 0.7597 & 0.8013 & 0.8002 & 0.7631 & 0.7868 \\
& BPOSE+EMG & 0.6405 & 0.6989 & 0.6619 & 0.5831 & 0.7278 & 0.5486 & 0.5073 & 0.7079 & 0.7304 & 0.5786 \\
& BSV+BPOSE+EMG & 0.6774 & 0.8132 & 0.8304 & 0.7682 & 0.7855 & 0.8203 & 0.8642 & 0.8652 & 0.8281 & 0.8566 \\
& CUBIST (SV+P) & 0.7960 & 0.7965 & 0.8484 & 0.7918 & 0.8236 & 0.8779 & 0.8572 & 0.8568 & 0.8664 & 0.7781  \\
& CUBIST (SV+P+E) & 0.8150 & 0.8780 & 0.7970 & 0.8315 & 0.8983 & 0.8600 & 0.9127 & 0.7703 & 0.9198 & 0.7924  \\ \cmidrule{2-12}

\multirow{5}{*}{\makecell[l]{\textbf{Multi}\\ \textbf{View}}}
& BMV & 0.6447 & 0.7581 & 0.7781 & 0.6875 & 0.7545 & 0.8036 & 0.8317 & 0.8111 & 0.7954 & 0.7946 \\
& BMV+BPOSE & 0.7154 & 0.8302 & 0.8418 & 0.7628 & 0.8327 & 0.8621 & 0.8872 & 0.8754 & 0.8680 & 0.8590 \\
& BMV+EMG & 0.6616 & 0.7732 & 0.7952 & 0.7019 & 0.7715 & 0.8186 & 0.8476 & 0.8255 & 0.8094 & 0.8110 \\
& BMV+BPOSE+EMG & 0.7296 & 0.8375 & 0.8472 & 0.7737 & 0.8380 & 0.8688 & 0.8951 & 0.8815 & 0.8721 & 0.8653 \\
& CUBIST (MV+P+E) & 0.8678 & 0.8757 & 0.8460 & 0.8792 & 0.9321 & 0.9000 & 0.9087 & 0.8144 & 0.9763 & 0.8349 \\
\bottomrule [1.5pt]
\end{tabular}

\vspace*{1.0em}
\noindent\textbf{Panel B: A11--A20}\par
\vspace*{0.3em}

\begin{tabular}{@{}>{\raggedright\arraybackslash}p{1.5cm}>{\raggedright\arraybackslash}p{3.2cm}*{10}{>{\centering\arraybackslash}p{1cm}}@{}}
\toprule [1.5pt]
\textbf{Modality} & \textbf{Model} & \textbf{A11} & \textbf{A12} & \textbf{A13} & \textbf{A14} & \textbf{A15} & \textbf{A16} & \textbf{A17} & \textbf{A18} & \textbf{A19} & \textbf{A20} \\
\midrule [1.0pt]
\multirow{8}{*}{\textbf{Vision}} 
& I3D-MLP & 0.4217 & 0.4335 & 0.4207 & 0.4983 & 0.5951 & 0.5047 & 0.5225 & 0.5282 & 0.5090 & 0.5966 \\
& CoRe & 0.6750 & 0.6317 & 0.4320 & 0.6989 & 0.7919 & 0.6963 & 0.7007 & 0.7658 & 0.7795 & 0.7706 \\
& TPT & 0.5163 & 0.6552 & 0.5347 & 0.4918 & 0.6586 & 0.5231 & 0.6016 & 0.7907 & 0.5763 & 0.5509 \\
& HGCN & 0.6391 & 0.6938 & 0.6327 & 0.7547 & 0.8387 & 0.7569 & 0.7126 & 0.7672 & 0.7592 & 0.8110 \\
& MCoRe & 0.5939 & 0.4725 & 0.7020 & 0.7656 & 0.9631 & 0.5307 & 0.8679 & 0.7999 & 0.6097 & 0.8173 \\
& T2CR & 0.3408 & 0.4802 & 0.2732 & 0.4981 & 0.6894 & 0.5402 & 0.5394 & 0.4073 & 0.5005 & 0.4707 \\
& DAE & 0.7336 & 0.6881 & 0.7210 & 0.7466 & 0.8181 & 0.7307 & 0.8233 & 0.8145 & 0.7434 & 0.8368 \\
& CUBIST & 0.8060 & 0.7415 & 0.7470 & 0.7371 & 0.7192 & 0.8233 & 0.7199 & 0.8649 & 0.7967 & 0.7753 \\ 
\cmidrule{2-12}

\multirow{4}{*}{\textbf{Pose}} 
& MS-G3D & 0.4732 & 0.6695 & 0.5766 & 0.4620 & 0.7012 & 0.4005 & 0.6730 & 0.6097 & 0.6439 & 0.7212 \\
& CTR-GCN & 0.4938 & 0.6460 & 0.6606 & 0.5035 & 0.7478 & 0.5864 & 0.6696 & 0.6913 & 0.6376 & 0.6910 \\
& ST-GCN++ & 0.5959 & 0.6803 & 0.6339 & 0.7207 & 0.9037 & 0.5173 & 0.6334 & 0.6899 & 0.7265 & 0.7983 \\
& SkateFormer & 0.5725 & 0.6060 & 0.6454 & 0.6615 & 0.8505 & 0.6474 & 0.7191 & 0.7717 & 0.6844 & 0.8010 \\ \cmidrule{2-12}

\multirow{1}{*}{\textbf{EMG}} 
& DAE & 0.1271 & 0.2641 & 0.2308 & 0.1386 & 0.3753 & 0.1630 & 0.1877 & 0.1592 & 0.1054 & 0.3188 \\ \cmidrule{2-12}

\multirow{6}{*}{\makecell[l]{\textbf{Single}\\ \textbf{View}}}
& BSV+BPOSE & 0.7885 & 0.7436 & 0.7838 & 0.7967 & 0.8721 & 0.7922 & 0.8597 & 0.8613 & 0.7751 & 0.8851 \\
& BSV+EMG & 0.7502 & 0.7089 & 0.7449 & 0.7602 & 0.8345 & 0.7457 & 0.8378 & 0.8333 & 0.7549 & 0.8524 \\
& BPOSE+EMG & 0.6137 & 0.6155 & 0.6704 & 0.6936 & 0.8825 & 0.6514 & 0.6461 & 0.7328 & 0.6932 & 0.7997 \\
& BSV+BPOSE+EMG & 0.8231 & 0.7703 & 0.8152 & 0.8332 & 0.9048 & 0.8236 & 0.8936 & 0.8903 & 0.8115 & 0.9195 \\
& CUBIST (SV+P) & 0.8425 & 0.8606 & 0.7689 & 0.8543 & 0.8308 & 0.7681 & 0.8568 & 0.8497 & 0.8469 & 0.8211\\
& CUBIST (SV+P+E) & 0.8538 & 0.8846 & 0.7745 & 0.8297 & 0.8420 & 0.7888 & 0.8612 & 0.8949 & 0.8513 & 0.8531\\ \cmidrule{2-12}

\multirow{5}{*}{\makecell[l]{\textbf{Multi}\\ \textbf{View}}}
& BMV & 0.7430 & 0.7377 & 0.7815 & 0.7810 & 0.8591 & 0.7795 & 0.8525 & 0.8547 & 0.7992 & 0.8748 \\
& BMV+BPOSE & 0.8102 & 0.7910 & 0.8412 & 0.8362 & 0.9229 & 0.8482 & 0.9091 & 0.9112 & 0.8501 & 0.9408 \\
& BMV+EMG & 0.7559 & 0.7517 & 0.7980 & 0.7969 & 0.8742 & 0.7939 & 0.8698 & 0.8703 & 0.8156 & 0.8907 \\
& BMV+BPOSE+EMG & 0.8181 & 0.7952 & 0.8455 & 0.8443 & 0.9294 & 0.8553 & 0.9145 & 0.9146 & 0.8569 & 0.9460 \\
& CUBIST (MV+P+E) & 0.9111 & 0.9183 & 0.8083 & 0.8681 & 0.8928 & 0.8424 & 0.9046 & 0.9483 & 0.8588 & 0.8838 \\
\bottomrule [1.5pt]
\end{tabular}
\end{table*}

\begin{table*}[htbp]
\centering
\caption{\textbf{Performance of AQA models on each action in the Cross-Subject splits ($\rho \uparrow$).}}
\label{table_exp_cs_detail}
\setlength{\tabcolsep}{3.5pt}

\noindent\textbf{Panel A: A01--A10}\par

\begin{tabular}{@{}>{\raggedright\arraybackslash}p{2.95cm}*{10}{>{\centering\arraybackslash}p{1.2cm}}@{}}
\toprule [1.5pt]
\textbf{Model} & \textbf{A01} & \textbf{A02} & \textbf{A03} & \textbf{A04} & \textbf{A05} & \textbf{A06} & \textbf{A07} & \textbf{A08} & \textbf{A09} & \textbf{A10} \\
\midrule [1.0pt]

Single-View & 0.0589 & 0.3558 & 0.0983 & 0.3018 & 0.4618 & 0.1686 & 0.3436 & 0.6652 & 0.4201 & 0.6200 \\
SkateFormer & 0.2408 & 0.1325 & 0.3272 & 0.5297 & 0.5823 & 0.4260 & 0.3946 & 0.1378 & 0.1367 & 0.2333 \\
EMG & 0.1672 & 0.0530 & 0.0863 & 0.0065 & 0.2597 & 0.1546 & 0.0638 & 0.2548 & 0.1408 & 0.1573 \\
BSV+BPOSE+EMG & 0.1005 & 0.3654 & 0.2141 & 0.3327 & 0.5814 & 0.3791 & 0.1193 & 0.5170 & 0.3195 & 0.5290 \\
BMV+BPOSE+EMG & 0.2038 & 0.3866 & 0.2660 & 0.5499 & 0.6963 & 0.3466 & 0.2488 & 0.5141 & 0.2726 & 0.4432 \\
\bottomrule [1.5pt]
\end{tabular}

\noindent\textbf{Panel B: A11--A20}\par

\begin{tabular}{@{}>{\raggedright\arraybackslash}p{2.95cm}*{10}{>{\centering\arraybackslash}p{1.2cm}}@{}}
\toprule [1.5pt]
\textbf{Model} & \textbf{A11} & \textbf{A12} & \textbf{A13} & \textbf{A14} & \textbf{A15} & \textbf{A16} & \textbf{A17} & \textbf{A18} & \textbf{A19} & \textbf{A20} \\
\midrule [1.0pt]

Single-View & 0.2153 & 0.2303 & 0.2659 & 0.1105 & 0.6509 & 0.1143 & 0.5191 & 0.7660 & 0.1464 & 0.4551 \\
SkateFormer & 0.1382 & 0.1870 & 0.1767 & 0.1668 & 0.5928 & 0.4020 & 0.6447 & 0.5708 & 0.1151 & 0.6995 \\
EMG & 0.1206 & 0.0087 & 0.0980 & 0.0638 & 0.3484 & 0.1895 & 0.1078 & 0.1794 & 0.1195 & 0.1710 \\
BSV+BPOSE+EMG & 0.1802 & 0.2152 & 0.2492 & 0.2425 & 0.6221 & 0.5023 & 0.6967 & 0.6300 & 0.1816 & 0.7284 \\
BMV+BPOSE+EMG & 0.12540 & 0.1887 & 0.3512 & 0.3124 & 0.5769 & 0.5971 & 0.7216 & 0.7207 & 0.2373 & 0.7589 \\
\bottomrule [1.5pt]
\end{tabular}
\end{table*}

\begin{table*}[htbp]
\centering
\vspace{-0.25cm}
\caption{\textbf{Performance of AQA models on each action in the Cross-View splits ($\rho \uparrow$).}}
\label{table_exp_cv_detail}
\setlength{\tabcolsep}{3.5pt}

\noindent\textbf{Panel A: A01--A10}\par

\begin{tabular}{@{}>{\raggedright\arraybackslash}p{2.95cm}*{10}{>{\centering\arraybackslash}p{1.2cm}}@{}}
\toprule [1.5pt]
\textbf{Model} & \textbf{A01} & \textbf{A02} & \textbf{A03} & \textbf{A04} & \textbf{A05} & \textbf{A06} & \textbf{A07} & \textbf{A08} & \textbf{A09} & \textbf{A10} \\
\midrule [1.0pt]

I3D-MLP & 0.3264 & 0.5510 & 0.3799 & 0.2694 & 0.4329 & 0.2806 & 0.0578 & 0.5087 & 0.3341 & 0.1246 \\
CoRe & 0.3803 & 0.3152 & 0.3530 & 0.3071 & 0.3885 & 0.4110 & 0.4908 & 0.4874 & 0.4202 & 0.3733 \\
TPT & 0.3672 & 0.4536 & 0.3101 & 0.2869 & 0.3743 & 0.4392 & 0.2848 & 0.3718 & 0.2885 & 0.3244 \\
HGCN & 0.3953 & 0.7166 & 0.4500 & 0.4006 & 0.3513 & 0.3893 & 0.5169 & 0.5520 & 0.3250 & 0.4726 \\
MCoRe & 0.3865 & 0.4577 & 0.4453 & 0.3874 & 0.4486 & 0.4275 & 0.3339 & 0.4585 & 0.4267 & 0.3307 \\
T2CR & 0.2308 & 0.4112 & 0.2417 & 0.3496 & 0.1773 & 0.1727 & 0.1229 & 0.5618 & 0.3270 & 0.3148 \\
DAE & 0.5775 & 0.5368 & 0.1881 & 0.3463 & 0.6717 & 0.4268 & 0.4497 & 0.5542 & 0.5690 & 0.4500 \\
\bottomrule [1.5pt]
\end{tabular}

\noindent\textbf{Panel B: A11--A20}\par

\begin{tabular}{@{}>{\raggedright\arraybackslash}p{2.95cm}*{10}{>{\centering\arraybackslash}p{1.2cm}}@{}}
\toprule [1.5pt]
\textbf{Model} & \textbf{A11} & \textbf{A12} & \textbf{A13} & \textbf{A14} & \textbf{A15} & \textbf{A16} & \textbf{A17} & \textbf{A18} & \textbf{A19} & \textbf{A20} \\
\midrule [1.0pt]

I3D-MLP & 0.2688 & 0.4579 & 0.2580 & 0.2131 & 0.6398 & 0.3691 & 0.3720 & 0.0545 & 0.1795 & 0.1944 \\
CoRe & 0.3834 & 0.3381 & 0.1273 & 0.3674 & 0.4837 & 0.4138 & 0.3634 & 0.5101 & 0.5460 & 0.5012 \\
TPT & 0.3974 & 0.3207 & 0.3415 & 0.3881 & 0.3018 & 0.3590 & 0.3384 & 0.4903 & 0.3013 & 0.3364 \\
HGCN & 0.4124 & 0.4051 & 0.1801 & 0.5447 & 0.7801 & 0.4496 & 0.4557 & 0.4039 & 0.1461 & 0.3973 \\
MCoRe & 0.3346 & 0.3499 & 0.3363 & 0.5396 & 0.4426 & 0.4119 & 0.4486 & 0.4071 & 0.4815 & 0.3989 \\
T2CR & 0.2469 & 0.1724 & 0.2398 & 0.1384 & 0.1229 & 0.3739 & 0.6558 & 0.2508 & 0.4074 & 0.4497 \\
DAE & 0.3223 & 0.4635 & 0.1952 & 0.3131 & 0.7174 & 0.5240 & 0.5742 & 0.5639 & 0.5674 & 0.3153 \\
\bottomrule [1.5pt]
\end{tabular}
\end{table*}

\begin{table*}[htbp]
\centering
\vspace{-0.25cm}
\caption{\textbf{Performance of AQA models on each action in the Mix-View splits ($\rho \uparrow$).}}
\label{table_exp_mv_detail}
\setlength{\tabcolsep}{3.5pt}

\noindent\textbf{Panel A: A01--A10}\par

\begin{tabular}{@{}>{\raggedright\arraybackslash}p{2.95cm}*{10}{>{\centering\arraybackslash}p{1.2cm}}@{}}
\toprule [1.5pt]
\textbf{Model} & \textbf{A01} & \textbf{A02} & \textbf{A03} & \textbf{A04} & \textbf{A05} & \textbf{A06} & \textbf{A07} & \textbf{A08} & \textbf{A09} & \textbf{A10} \\
\midrule [1.0pt]

I3D-MLP & 0.3006 & 0.2310 & 0.3236 & 0.2517 & 0.4099 & 0.3552 & 0.3662 & 0.3628 & 0.3512 & 0.3156 \\
CoRe & 0.4215 & 0.3168 & 0.4213 & 0.3766 & 0.4659 & 0.4950 & 0.4945 & 0.5222 & 0.4033 & 0.4051 \\
TPT & 0.3393 & 0.5724 & 0.3427 & 0.3974 & 0.3762 & 0.5337 & 0.3663 & 0.3108 & 0.4211 & 0.3620 \\
HGCN & 0.4162 & 0.4498 & 0.5356 & 0.4412 & 0.3948 & 0.4475 & 0.5105 & 0.5168 & 0.4683 & 0.4726 \\
MCoRe & 0.4692 & 0.3972 & 0.4304 & 0.2088 & 0.5446 & 0.2461 & 0.3407 & 0.7362 & 0.5790 & 0.4548 \\
T2CR & 0.1811 & 0.2614 & 0.4030 & 0.2388 & 0.4049 & 0.3773 & 0.2304 & 0.4155 & 0.2736 & 0.2334 \\
DAE & 0.3550 & 0.4776 & 0.4944 & 0.4512 & 0.4508 & 0.5087 & 0.5498 & 0.5482 & 0.5123 & 0.5376 \\
\bottomrule [1.5pt]
\end{tabular}

\noindent\textbf{Panel B: A11--A20}\par

\begin{tabular}{@{}>{\raggedright\arraybackslash}p{2.95cm}*{10}{>{\centering\arraybackslash}p{1.2cm}}@{}}
\toprule [1.5pt]
\textbf{Model} & \textbf{A11} & \textbf{A12} & \textbf{A13} & \textbf{A14} & \textbf{A15} & \textbf{A16} & \textbf{A17} & \textbf{A18} & \textbf{A19} & \textbf{A20} \\
\midrule [1.0pt]

I3D-MLP & 0.2841 & 0.2959 & 0.2831 & 0.3607 & 0.4575 & 0.3671 & 0.3849 & 0.3906 & 0.3714 & 0.4590 \\
CoRe & 0.4303 & 0.3870 & 0.1873 & 0.4542 & 0.5472 & 0.4516 & 0.4560 & 0.5211 & 0.5348 & 0.5259 \\
TPT & 0.3420 & 0.4809 & 0.3604 & 0.3175 & 0.4843 & 0.3488 & 0.4273 & 0.6164 & 0.4020 & 0.3766 \\
HGCN & 0.4123 & 0.4670 & 0.4059 & 0.5279 & 0.6119 & 0.5301 & 0.4858 & 0.5404 & 0.5324 & 0.5842 \\
MCoRe & 0.3596 & 0.2382 & 0.4677 & 0.5313 & 0.7288 & 0.2964 & 0.6336 & 0.5656 & 0.3754 & 0.5830 \\
T2CR & 0.1964 & 0.3358 & 0.1288 & 0.3537 & 0.5450 & 0.3958 & 0.3950 & 0.2629 & 0.3561 & 0.3263 \\
DAE & 0.5016 & 0.4561 & 0.4890 & 0.5146 & 0.5861 & 0.4987 & 0.5913 & 0.5825 & 0.5114 & 0.6048 \\
\bottomrule [1.5pt]
\end{tabular}
\end{table*}

\begin{table*}[htbp]
\centering
\caption{\textbf{Qualitative comparison of outputs from CUBIST and SOTA AQA Model.} Compared to DAE, which outputs only a single holistic score, CUBIST not only achieves higher prediction accuracy but also introduces interpretable mechanisms for error detection and compositional scoring, thereby significantly enhancing the transparency of the assessment. Furthermore, by leveraging the FKG, CUBIST provides actionable, expert-level feedback corresponding to specific errors.} 

\label{table_exp_cubist}
\renewcommand{\arraystretch}{1.4}
\renewcommand{\tabularxcolumn}[1]{m{#1}}
\begin{tabularx}{\textwidth}{@{} l l >{\centering\arraybackslash}X >{\centering\arraybackslash}X @{}} 
\toprule [1.5pt]
\textbf{Component} & \textbf{Model} & \textbf{\textcolor{CustomOrange}{Decline Barbell Bench Press}} & \textbf{\textcolor{RoyalBlue}{Squat Exercise}} \\ \midrule [1.5pt]

\multicolumn{2}{@{}l}{\textbf{True Score}} & 70.23 & 72.39 \\ \midrule

\multirow{2}{*}{\textbf{Predicted Score}} 
& DAE & \textcolor{red}{61.24} & \textcolor{red}{78.41} \\
& \textbf{CUBIST} & \textcolor{OliveGreen}{\textbf{63.36}} & \textcolor{OliveGreen}{\textbf{68.66}} \\ \midrule

\multirow{2}{*}{\textbf{\makecell[l]{Interpretable\\Error Detection}}} 
& DAE & \textcolor{red}{\xmark} & \textcolor{red}{\xmark} \\ 
& \textbf{CUBIST} & 
\vspace{0.4em} \textcolor{OliveGreen}{\cmark} \par \vspace{0.3em} \includegraphics[width=0.75\linewidth]{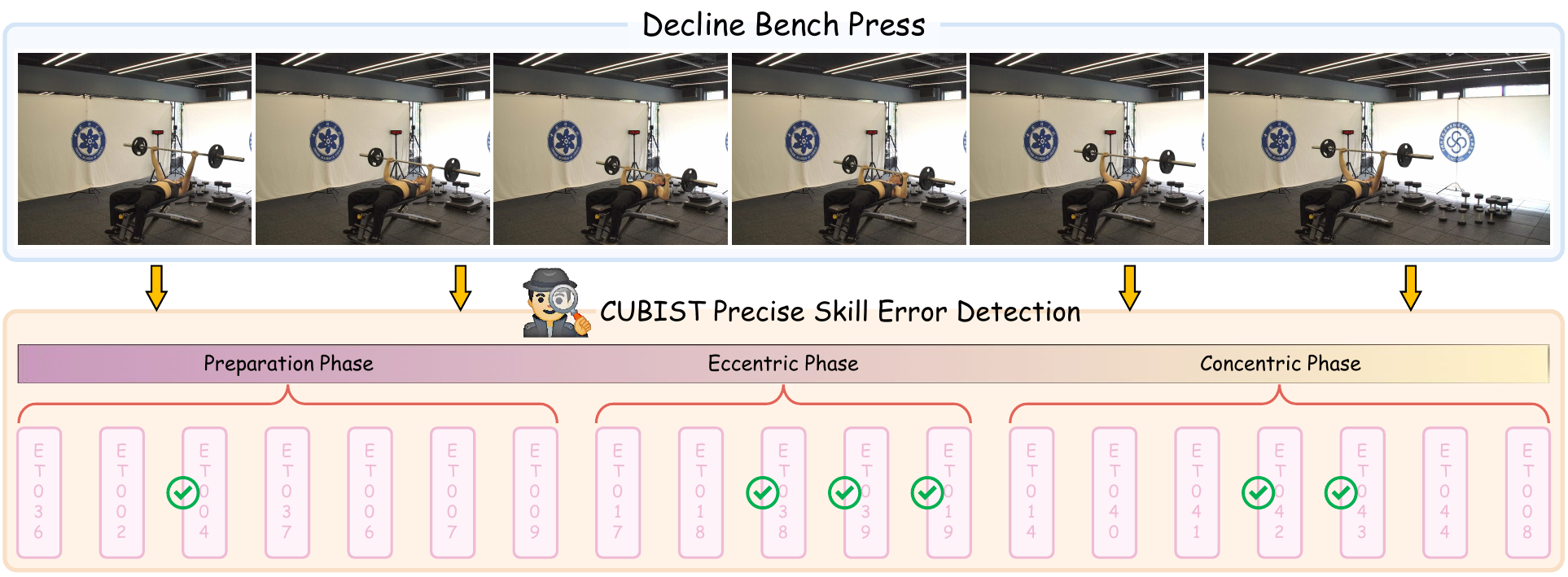} \vspace{0.4em} & 
\vspace{0.4em} \textcolor{OliveGreen}{\cmark} \par \vspace{0.3em} \includegraphics[width=0.75\linewidth]{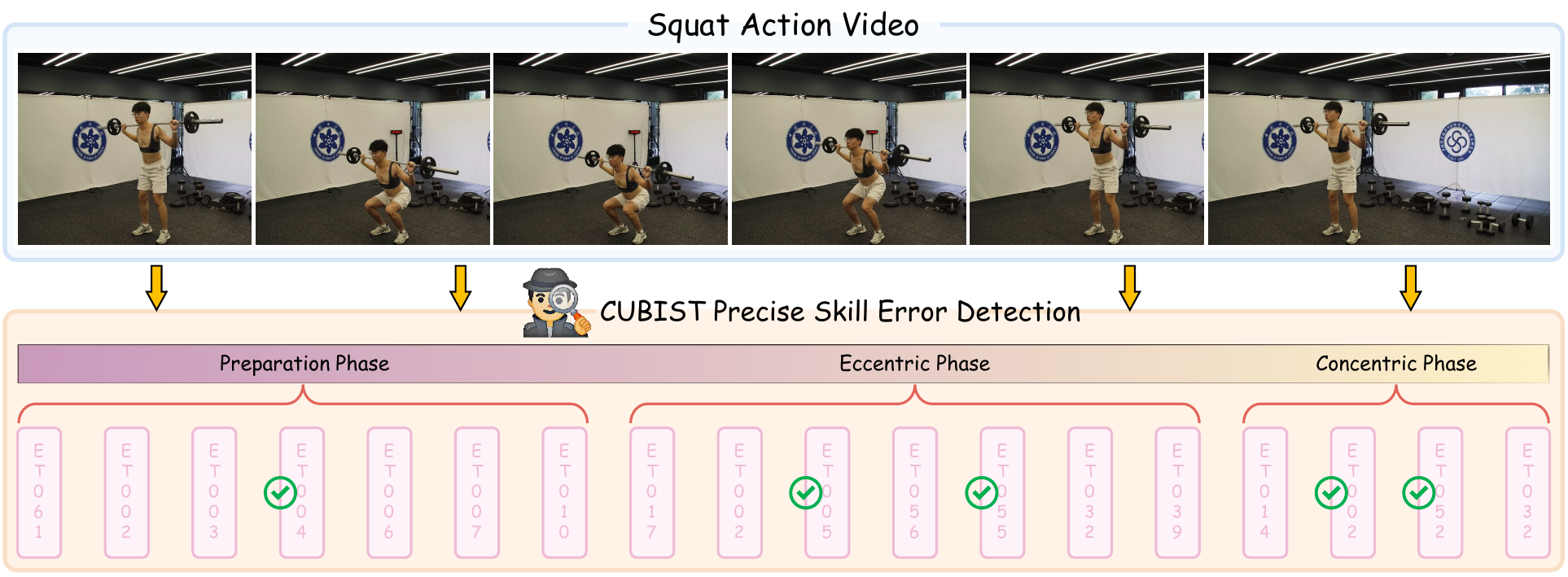} \vspace{0.4em} \\

\midrule

\multirow{2}{*}{\textbf{\makecell[l]{Compositional\\Scoring}}} 
& DAE & \textcolor{red}{\textbf{\xmark}} & \textcolor{red}{\textbf{\xmark}} \\
& \textbf{CUBIST} & \textcolor{OliveGreen}{\textbf{[1-(8+8+...+9+9)/131]$\times$100=63.36}} & \textcolor{OliveGreen}{\textbf{[1-(8+5+...+9+10)/134]$\times$100=68.66}} \\ \midrule

\multirow{14}{*}{\textbf{Feedback}} 
& DAE & \textcolor{red}{\textbf{\xmark}} & \textcolor{red}{\textbf{\xmark}} \\
\cmidrule{2-4}
& \textbf{CUBIST} & 
\begin{tabitemize}
    \item Actively depress the shoulders. Imagine pulling the scapulae down toward the hips, tightening them back and downward to prevent the shoulders from elevating toward the ears.
    \item Maintain an angle of about $45^\circ$-$60^\circ$ between the upper arms and torso, forming an arrow shape.
    \item Lower slowly until the target muscles are adequately stretched. Reduce the load if necessary.
    \item Adjust the barbell to an appropriate finishing position.
    \item Actively retract and depress the scapulae against the bench, maintaining a chest up, shoulders down posture. As you press, visualize compressing the target muscles.
    \item Imagine bending the barbell, maintaining symmetrical force output in both hands.
\end{tabitemize} & 
\begin{tabitemize}
    \item Ensure the knees always track over the toes, avoiding inward collapse.
    \item Slightly turn the toes outward and distribute your weight evenly across the entire foot. Reduce the load if necessary.
    \item Sit the hips back while the knees move forward slightly to maintain balance but avoid excessive forward knee travel. Keep the heels firmly on the ground, spine straight, and the bar’s center of gravity between the forefoot and heel.
    \item Engage the abdominal muscles and slightly contract the glutes, maintaining the lumbar spine in a neutral position.
    \item Rise by extending the hips and knees together, lifting both the hip and shoulder joints simultaneously while keeping the back stable.
\end{tabitemize} \\

\bottomrule  [1.5pt]
\end{tabularx}
\end{table*}

\noindent\textbf{Performance of Uni-Modal Models.}
\blackcircled{1}
We conduct unimodal comparison experiments on the Vanilla split. We first train and evaluate 7 baseline models from the vision branch alongside the CUBIST model under a unified experimental configuration. The results in \autoref{table_exp_aqa_avg} show clear hierarchical differences in performance among vision baseline models with diverse topological architectures.
\blackcircled{2}
The performance of foundational models such as I3D-MLP remains relatively limited. Conversely, temporal modeling or contrastive learning networks explicitly tailored for the AQA task, such as CoRe and TPT, achieve significant performance gains. Further analysis demonstrates that DAE, which relies on fine-grained feature representations, attains highly competitive results ($\rho = 0.7431$). Notably, our proposed CUBIST model outperforms all baselines -- including highly competitive ones -- achieving new SOTA performance ($\rho = 0.7829$, $R-\ell_2 = 2.1542$), thereby validating the advantages of our paradigm and the FKG structure.
\blackcircled{3}
For the pose modality, we evaluated multiple classical architectures based on skeletal keypoints. Experimental results indicate that models employing advanced graph convolutional networks (GCNs) and Transformer architectures demonstrate superior representation capabilities. Specifically, SkateFormer achieved a performance of $\rho = 0.6508$, significantly outperforming MS-G3D ($0.5512$), CTR-GCN ($0.5873$), and ST-GCN++ ($0.6072$), highlighting its advantage in capturing pure kinematic topological features of the human body.
\blackcircled{4}
Furthermore, we conducted an independent evaluation to characterize the performance of the sEMG modality ($\rho = 0.2046$). As expected, the standalone performance of the sEMG modality appears relatively limited from a purely numerical perspective. Nevertheless, the valuable physiological insights it offers extend beyond visual or kinematic modalities. Specifically, sEMG provides direct and objective information about the activation patterns of targeted muscle groups, including co-contraction, fatigue signatures, force imbalances, and compensatory muscle recruitment. Beyond its numerical contribution to overall prediction accuracy, this physiological information deserves deeper consideration for the complementary perspective it provides within a multimodal understanding framework.\\

\noindent\textbf{Effectiveness of Multi-Modal Fusion.}
\blackcircled{1}
Based on the quantitative findings from the unimodal experiments, we selected DAE \cite{zhang2024auto} as the visual backbone for our subsequent multimodal baseline models (BSV/BMV), and SkateFormer \cite{do2024skateformer} as the skeletal backbone (BPose). We then systematically ablate the multimodal configurations by progressively combining different modalities and evaluating their performance. In parallel, we conduct the same multimodal evaluations using our CUBIST model.
\blackcircled{2}
Under single-view (SV) conditions, purely visual unimodal networks are inherently limited by the absence of explicit depth information and their susceptibility to self-occlusion. When skeletal information (BSV+BPOSE, $7.04\%$ gain) or EMG signals (BSV+EMG, $2.37\%$ gain) are incorporated into BSV (DAE), the model achieves clear performance improvements. Furthermore, tri-modal fusion (BSV + BPOSE + EMG) raises performance to $\rho=0.8297$, corresponding to an $11.65\%$ improvement over the existing unimodal SOTA, DAE. More importantly, CUBIST demonstrates strong multimodal modeling capability, achieving $\rho=0.8296$ with only bimodal inputs. After integrating all three modalities in CUBIST(SV+POSE+EMG), performance further improves to $\rho=0.8455$, representing a $7.99\%$ gain over unimodal CUBIST, while $R-\ell_2$ decreases to $1.0198$.
\blackcircled{3}
To alleviate the spatial ambiguity inherent in single-view inputs, we further extend the evaluation to multi-view (BMV) settings. The multi-view input itself ($\rho=0.7861$) yields a $5.7\%$ improvement over the single-view SOTA. The benefits of multimodal fusion remain evident: relative to DAE, incorporating skeletal information (BMV+BPOSE) and all modalities (BMV+BPOSE+EMG) improves $\rho$ by $14.36\%$ and $15.25\%$, respectively. Ultimately, CUBIST(MV + POSE + EMG) achieves the best overall performance, with $\rho=0.8836$ and $R-\ell_2=0.7418$. Compared with unimodal CUBIST, this corresponds to a $12.86\%$ improvement in $\rho$; compared with DAE, the improvement reaches $18.91\%$. These progressive ablation results validate the SOTA performance of CUBIST and highlight the importance of complementary information sources for AQA. Accurate action quality assessment benefits not only from external kinematic observations, such as multi-view appearance and skeletal structure, but also from internal physiological cues provided by sEMG. By complementing visual representations with muscle-activation information, sEMG helps the model capture aspects of action execution that are not directly observable from appearance alone.
\blackcircled{4}
In summary, the multimodal ablation experiments show that progressively introducing complementary modalities leads to consistent performance gains on the AQA task. Across the evaluated settings, skeleton keypoints (POSE) and multiple views (MV) contribute notable improvements, with $\rho$ increases of $7.76\%$ and $6.32\%$, respectively, while sEMG provides a stable additional gain of $2.27\%$. Although the absolute improvement from EMG is smaller, its consistent contribution supports its complementary role in multimodal AQA. Overall, these results demonstrate a clear cross-modal synergy between external kinematic representations and internal physiological signals.\\

\noindent\textbf{Performance on Generalization-oriented Splits.}
\blackcircled{1}
To thoroughly investigate the model's generalization capability under data distribution shifts, we conduct benchmark testing across three generalization-oriented data-splitting protocols: cross-subject, cross-view, and mix-view.
\blackcircled{2}
As shown in \autoref{table_exp_aqa_avg}, all single-modality baseline models exhibit substantial performance degradation when encountering unseen subjects or novel camera views. For example, in the highly challenging cross-subject split, where subject-specific differences in movement patterns become especially pronounced, the $\rho$ value of the single-view visual model drops to $0.3484$, while SkateFormer similarly declines to $0.3417$. These results highlight the limitations of unimodal models in generalizing across subject-dependent appearance and motion variations.
\blackcircled{3}
In contrast, the multimodal fusion paradigm provides improved robustness under such distribution shifts. By incorporating structural pose cues (POSE) and appearance-independent physiological signals (EMG), the multimodal configuration helps alleviate the sharp decline in performance. Notably, in the cross-subject setting, the combined BMV+BPOSE+EMG model raises the correlation metric to $0.4323$, significantly outperforming each single-modality baseline. Furthermore, comparing the quantitative results of cross-view and mix-view reveals better overall performance under the mix-view setting. This improvement is likely attributable to the availability of video streams from multiple views during mix-view training, which enables the network to learn more robust view-invariant representations. In contrast, the cross-view protocol requires inference on entirely unseen viewpoints, making it a substantially more demanding generalization setting.\\

\noindent\textbf{Summary of the Results.}
Our research presents the following four core findings:
\begin{itemize}
\item CUBIST achieves the best performance: In pure vision unimodal evaluations, CUBIST demonstrates strong spatiotemporal modeling capability and achie-ves new SOTA performance. Under multimodal fusion settings, it further exhibits effective cross-modal feature integration and again achieves new SOTA performance, highlighting the effectiveness of MyoMechanix's FKG design, multimodal information, and the CUBIST paradigm.

\item Multimodal data delivers clear performance gains: Experimental results demonstrate that the transition from single-view to multi-view inputs, together with the incorporation of complementary modalities spanning external kinematics (vision and skeleton) and internal physiology (surface electromyography), yields substantial synergistic benefits. These findings highlight the value of multidimensional information for accurate AQA.

\item Limited generalization under challenging splits emphasizes the dataset difficulty: When confronted with substantial data distribution shifts, such as those across subjects and viewpoints, advanced models exhibit noticeable performance degradation. This finding underscores the challenge of disentangling action quality from subject- and view-specific variations, while also motivating future research toward more robust real-world AQA systems.

\item Considerable room for improvement remains across all splits: The results indicate substantial opportunities for future research in model design, multimodal fusion, generalization, interpretability, and related directions.
\end{itemize}

\subsection{Video Question Answering}
\label{Sec52-AQU}

\begin{table*}[htbp]
\caption{\textbf{Performance of VLMs on MyoMechanix-VideoQA dataset.} $R-\ell_2(\times100)$}
\label{table_exp_aqu}
\centering
\begin{tabularx}{\textwidth}{@{}l>{\centering\arraybackslash}X>{\centering\arraybackslash}X>{\centering\arraybackslash}X>{\centering\arraybackslash}X>{\centering\arraybackslash}X>{\centering\arraybackslash}X>{\centering\arraybackslash}X@{}}
    \toprule [1.5pt]
    \multirow{2}{*}{\textbf{Metrics}} & \multicolumn{3}{c}{\textbf{Pretrained}} & \multicolumn{3}{c}{\textbf{Prompt}} & \textbf{SFT}  \\ \cmidrule(lr){2-4} \cmidrule(lr){5-7} \cmidrule(lr){8-8}
                                         &InternVL& MiniCPM& Qwen   &InternVL& MiniCPM& Qwen   & Qwen \\ \midrule [1.0pt]
    \textbf{BLEU} $\uparrow$             & 0.0003 & 0.0154 & 0.0388 & 0.0007 & 0.0312 & 0.0415 & \textbf{0.1970} \\
    \textbf{ROUGE-L} $\uparrow$          & 0.1689 & 0.2902 & 0.2012 & 0.2002 & 0.3102 & 0.2128 & \textbf{0.4010}      \\
    \textbf{METEOR} $\uparrow$           & 0.1405 & 0.3024 & 0.3049 & 0.1736 & 0.3345 & 0.3206 & \textbf{0.4688}     \\
    \textbf{CHRF++} $\uparrow$           & 8.7228 &20.4180 &36.1340 &17.3078 &28.4865 &39.5308 & \textbf{52.0005}    \\
    \textbf{BERTScore F1} $\uparrow$     & 0.1017 & 0.2535 & 0.1664 & 0.1550 & 0.2743 & 0.1666 & \textbf{0.4390}     \\
    \textbf{$R-\ell_2$}$\downarrow$      & 3.9900 &10.5539 & 3.1183 & 3.2757 & 8.6448 & 3.0942 & \textbf{3.0592}     \\
    \bottomrule [1.5pt]
\end{tabularx}
\end{table*}
\begin{table*}[h]
\small
\centering
\caption{\textbf{Qualitative comparison of models in natural language feedback and error analysis generation.}}
\label{table_exp_aqu_qc}
\resizebox{\textwidth}{!}{%
\rowcolors{2}{gray!15}{white}
\begin{tabular}{@{}l|c|c|c|c@{}}
    \toprule [1.5pt]
    \textbf{Dimension} & \textbf{Qwen (SFT)} & \textbf{Qwen (Prompt)} & \textbf{MiniCPM (Prompt)} & \textbf{InternVL (Prompt)} \\ \midrule [1.0pt]
    Coarse Action Recognition           & \checkmark         & \checkmark          & \checkmark  & \checkmark \\
    Fine-grained Analysis               & \textbf{Better}    & Good                & Limited     & Limited \\
    Error Diagnosis                     & 5 faults flagged   & 2 faults            & None        & None \\
    Key-point Coverage (5 total)        & 5                  & 4                   & 3           & 3 \\
    Wording                             & Concise, technical & Verbose, exhaustive & Moderate    & Direct, brief \\
    \bottomrule [1.5pt]
\end{tabular}}
\end{table*}

\begin{table*}[h]
\centering
\caption{\textbf{Qualitative comparison of models in score prediction.}}
\label{table_exp_aqu_score}
\resizebox{\textwidth}{!}{
\rowcolors{2}{gray!15}{white}
\begin{tabular}{@{}l|c|c|c|c|c@{}}
    \toprule [1.5pt]
    \textbf{Sample ID} & \textbf{GT Score} & \textbf{Qwen (SFT)} & \textbf{Qwen (Prompt)} & \textbf{MiniCPM (Prompt)} & \textbf{InternVL (Prompt)} \\ \midrule [1.0pt]
    P08-A09-09 & 100.00   & 65 & 65 & 95 & 85 \\
    N04-A15-01 & 22.16    & 65 & 65 & 95 & 75 \\
    A05-A20-04 & 68.15    & 55 & 65 & 75 & 75 \\
    \bottomrule [1.5pt]
\end{tabular}}
\end{table*}

In this experiment, we use MyoMechanix-VideoQA data-set (\autoref{Sec342-VideoQA_Dataset}) to assess whether domain-specific fine-tuning can equip current vision-language models (VLMs) and multimodal large language models (MLLMs) with query-conditioned, fine-grained biomechanical action understanding in fitness scenarios.

\subsubsection{Baseline Methods}

We select representative state-of-the-art open-source VLMs as baseline models, including MiniCPM-O-2.6-8B \cite{yao2024minicpm}, InternVL3-2B \cite{zhu2025internvl3}, and Qwen2.5-VL-3B \cite{bai2025qwen2}. To comprehensively and objectively evaluate model performance, we consider three distinct testing configurations: \textbf{1) Off-the-shelf mode}: directly using the original pre-trained weights for inference; \textbf{2) Rules-based prompting mode}: incorporating standardized, structured action execution criteria as contextual guidance during evaluation; and \textbf{3) Supervised fine-tuning mode}: applying LoRA-based fine-tuning to adapt the pre-trained models to our dataset and task.

\subsubsection{Implementation Details}

We performed supervised fine-tuning of Qwen2.5-VL using the LLaMA-Factory framework \cite{zheng2024llamafactory}. Training was conducted in bfloat16 precision with Low-Rank Adaptation (LoRA), using a rank of 8, an alpha of 16, and a dropout rate of 0.05. We optimized the model with AdamW, using an initial learning rate of $5 \times 10^{-5}$, cosine learning rate decay, and gradient clipping with a maximum norm of 1.0. The per-device batch size was set to 2, with gradient accumulation over 4 steps. Training took $\sim$12 hours on a single NVIDIA H20 GPU.

\subsubsection{Performance Metrics}

To rigorously evaluate the quality of the open-ended answers against the ground-truth annotations provided by experts, we employ a multi-granular evaluation framework. This framework spans from surface-level lexical fidelity (BLEU \cite{papineni2002bleu}, ROUGE-L \cite{lin2004automatic}) and morphological robustness (METEOR \cite{banerjee2005meteor}, CHRF++ \cite{popovic2017chrf++}) to deep semantic consistency (BERTScore-F1 \cite{zhangbertscore}).

Specifically, BLEU and ROUGE-L assess lexical agreement between generated responses and reference answers, including the accurate use of task-relevant anatomical terms such as body-part mentions  via the precision and recall of n-grams. METEOR further accounts for stemming and synonym matching, while CHRF++ captures character- and word-level n-gram similarity, making it less sensitive to minor surface-form variations. To complement these overlap-based measures, BERTScore-F1 leverages pre-trained contextual embeddings to estimate semantic similarity between generated and reference answers. Collectively, these metrics provide a more comprehensive assessment of both phrasing fidelity and meaning preservation. For all metrics, higher values indicate better performance.

\subsubsection{Results}

\noindent\textbf{Overall.} 
\blackcircled{1}
As shown in \autoref{table_exp_aqu}, the three testing strategies and three evaluated models exhibit a clear hierarchy in performance. Pre-trained models evaluated in a zero-shot setting provide a basic baseline; however, they generally struggle to produce the structured diagnostic responses required by our MyoMechanix-VideoQA benchmark. Importantly, this suggests that our dataset requires perception and reasoning not covered under existing datasets used for training foundational models. Thus, incorporating our dataset would help develop more powerful and holistic foundational models. Incorporating domain-specific scoring rubrics through prompting yields moderate yet consistent improvements across most metrics, suggesting that inference-time guidance helps align model outputs with task expectations. This also suggests that our dataset supports novel tasks, not yet covered by existing foundational datasets. Supervised fine-tuning (SFT) further enhances task-specific generation quality. Compared with the prompt-only setting, SFT on the Qwen model produces substantial gains in text generation metrics, including BLEU, which increases from 0.0415 to 0.1970, and CHRF++, which rises from 39.5308 to 52.0005. Despite these improvements in language generation, the $R-\ell_2$ error decreases by less than 2\% across adaptation stages. This suggests that while weight adaptation is more effective than explicit textual prompting for internalizing task-specific response patterns, its impact on score prediction accuracy remains limited.
\blackcircled{2}
Beyond the trends across adaptation strategies, the comparison of individual models highlights the importance of video-oriented pretraining for fine-grained biomechanical action understanding enabled by our MyoMechanix-VideoQA dataset. Across all baseline configurations, Qwen2.5-VL-3B consistently achieves the strongest overall performance. Despite its relatively small parameter count, Qwen appears to benefit from extensive multi-source video pretraining, which may provide stronger motion priors and contribute to its balanced performance across both lexical and semantic metrics. 
In contrast, MiniCPM-O-2.6-8B benefits from a stronger language backbone, enabling competitive performance on semantic-similarity metrics after prompting. However, its weaker results on stricter lexical matching metrics suggest that limitations in video-specific temporal understanding may reduce its fine-grained diagnostic precision. InternVL3-2B, which is designed for more lightweight deployment, shows lower overall performance, potentially reflecting combined constraints in visual perception and language modeling capacity. 
For score prediction, as measured by $R-\ell_2$, Qwen consistently achieves the lowest error, MiniCPM exhibits the highest error, and InternVL falls in between.\\

\noindent\textbf{Qualitative Text Comparison.}
To provide a detailed view of VLMs' performance, we select a representative example (P07-A17-01) from the test set to showcase text generation quality (see \autoref{table_exp_aqu_qc}).
\blackcircled{1}
For sample P07-A17-01, all evaluated VLMs correctly categorize the action as a barbell bent-over row. This indicates that coarse-grained action recognition is already robustly handled by current models. However, when the task shifts to the distinctive requirements of MyoMechanix-VideoQA -- joint-level motion diagnosis, temporal reasoning, and execution scoring -- the capabilities of these models sharply diverge. The SFT Qwen delivers the most diagnostically valuable narrative. It precisely pinpoints multiple biomechanical faults, such as spinal curvature, shoulder elevation, and improper eccentric tempo. Furthermore, it expresses these observations with succinct, discipline-specific diction, successfully blending technical precision with error-focused depth. 
\blackcircled{2}
In contrast, the same model under a prompt-only configuration achieves alignment with the reference checklist but adopts a largely affirmative stance devoid of critical feedback. This exhaustive yet uncurated recitation inflates the textual volume without yielding actionable coaching insights. Similarly, the MiniCPM and InternVL baselines reproduce core setup cues but verge on generating generic rehearsal manuals. They confine themselves to superficial form validation and entirely omit the identification of faults, which limits their utility for quantitative evaluation.
\blackcircled{3}
Ultimately, this contrast underscores the core challenge and novelty of the MyoMechanix-VideoQA benchmark. While current VLMs have largely mastered the classification of action types, the modeling of kinematic nuances and the delivery of fine-grained quality assessment remain significant open challenges.\\

\noindent\textbf{Stagnant Score Prediction.}
During the experiments (see \autoref{table_exp_aqu} and \autoref{table_exp_aqu_score}), we observed that although text generation quality improved markedly, the accuracy of scoring did not exhibit a corresponding improvement, which lagged behind specialized AQA models. We believe this can be primarily attributed to the inherent limitations of large pretrained language models in handling numerical data. 
During pretraining, these models rely on autoregressive prediction over discrete tokens -- splitting numbers into subword units and treating them as standard vocabulary. Consequently, they lack distance constraints on the continuous numerical axis and do not employ specialized regression losses to reinforce gradient signals for magnitude differences. Furthermore, the training objective of these models is to maximize likelihood rather than to achieve precise counting or numerical regression, with the optimization process favoring overall syntactic and semantic coherence. As a result, these models struggle to represent fine-grained visual details or distinctions in scores, yielding approximate rather than accurately calibrated scoring outputs.
To improve on numerical task, we suggest future work could begin with a unified modeling framework that integrates discrete generation and continuous regression. By leveraging numerically aware representation learning and multi-scale supervisory signals, numerical values would be mapped to continuous vectors endowed with dimensional and ordinal constraints. An explicit regression head or score-distance regularization could then be employed to simultaneously minimize language reconstruction error and numerical deviation during autoregressive generation. At the same time, introducing cross-modal alignment through pose, temporal, or physical constraints would enforce a consistent metric structure among vision, language, and numeric modalities in the latent space, thereby enhancing the model’s sensitivity to fine-grained quantitative differences and its generalization ability.\\

\noindent\textbf{Summary.} 
Our experimental results demonstrate that the MyoMechanix-VideoQA dataset presents a significant challenge to SOTA VLMs. Although supervised fine-tuning substantially improves model performance and establishes a promising baseline for future research, the multimodal understanding capability remains far from fully realized. In the text generation task, existing VLMs still have approximately 50\% room for performance improvement. In the action quality score prediction task, the performance of current VLMs still lags significantly behind that of domain-specific AQA algorithms by up to 80\%.

\subsection{Novel Technology---Video2EMG}
\label{Sec53-V2E}

\begin{figure}[ht]
    \centering
    \includegraphics[width=\linewidth]{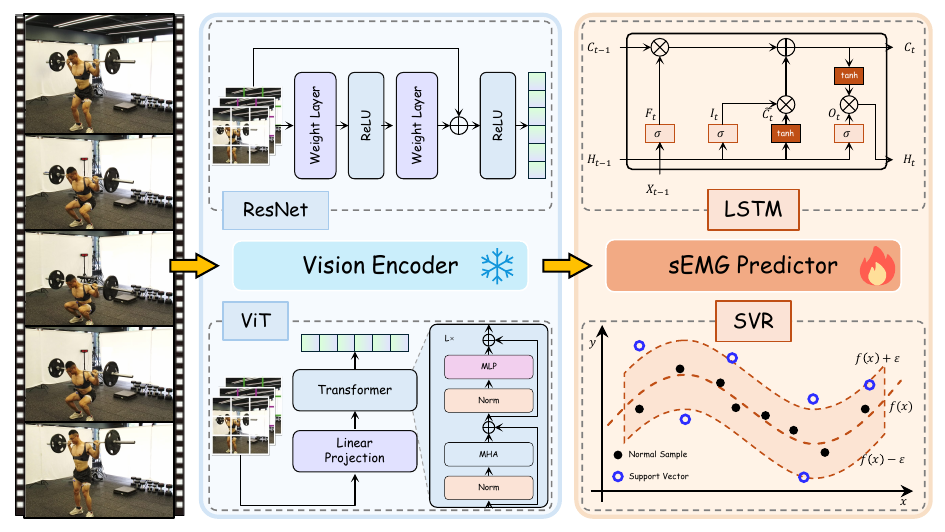}
    \caption{\textbf{Illustration of proposed baseline models' architectures for the Video2EMG task.}}
    \label{fig_exp_v2e_str}
\end{figure}

\begin{table*}[htbp]
\centering
\caption{\textbf{Performance of Video2EMG models.}}
\label{table_exp_v2e}
\resizebox{\textwidth}{!}{
\begin{tabular}{@{}l
    >{\centering\arraybackslash}p{1.2cm}
    >{\centering\arraybackslash}p{1.2cm}
    >{\centering\arraybackslash}p{1.4cm}
    >{\centering\arraybackslash}p{1.2cm}
    >{\centering\arraybackslash}p{1.2cm}
    >{\centering\arraybackslash}p{1.4cm}
    >{\centering\arraybackslash}p{1.2cm}
    >{\centering\arraybackslash}p{1.2cm}
    >{\centering\arraybackslash}p{1.4cm}
@{}}

\toprule [1.5pt]
\multirow{2}{*}{\textbf{Model}} & \multicolumn{3}{c}{\textbf{Vanilla}} & \multicolumn{3}{c}{\textbf{Cross-View}} & \multicolumn{3}{c}{\textbf{Cross-Subject}} \\ \cmidrule(lr){2-4} \cmidrule(lr){5-7} \cmidrule(lr){8-10}
& \textbf{CCP $\uparrow$} & \textbf{MAE $\downarrow$} & \textbf{RMSE $\downarrow$} 
& \textbf{CCP $\uparrow$} & \textbf{MAE $\downarrow$} & \textbf{RMSE $\downarrow$}
& \textbf{CCP $\uparrow$} & \textbf{MAE $\downarrow$} & \textbf{RMSE $\downarrow$} \\
\midrule [1.0pt]
ResNet+LSTM 
& 0.3706 & 0.1655 & 0.2133 
& 0.2723 & 0.2550 & 0.3054
& 0.2872 & 0.2934 & 0.3442 \\

ResNet+SVR 
& 0.4174 & 0.1466 & 0.1878 
& 0.2793 & 0.1508 & 0.2002
& 0.3062 & 0.1705 & 0.2170 \\

ViT+LSTM
& 0.4040 & 0.1648 & 0.2069
& 0.3104 & 0.2079 & 0.2619
& 0.3086 & 0.2148 & 0.2655 \\

ViT+SVR
& 0.4345 & 0.1465 & 0.1882
& 0.3311 & 0.1483 & 0.1985
& 0.3311 & 0.1631 & 0.2104 \\
\bottomrule [1.5pt]
\end{tabular}}
\end{table*}

As introduced earlier, while surface electromyography (sEMG) signals provide indispensable, precise evidence of muscle functional states, their acquisition relies on costly and obtrusive wearable sensors. To mitigate this limitation, we envision a future where fine-grained, muscle-level feedback can be estimated directly from ordinary videos. To pave the way for this novel cost-effective technology, we implement and evaluate the proposed Video2EMG task. By leveraging the strictly synchronized multimodal recordings within the MyoMechanix dataset, this benchmark challenges models to predict continuous sEMG sequences solely from visual observations. In this section, we establish the foundational baselines for this novel cross-modal translation task, demonstrating the feasibility of inferring internal, latent physiological dynamics from observable kinematic cues.

\subsubsection{Our Video2EMG Baseline Design}
Given the novelty of the Video2EMG task, established models are currently unavailable for direct usage or adoption. Thus, we design sound baseline models for the Video2EMG task. For this, we draw upon the core principles of time-series waveform analysis and relevant studies such as emg2pose \cite{salter2024emg2pose} to construct streamlined end-to-end spatiotemporal, visual regression models to predict EMG signals from videos (see \autoref{fig_exp_v2e_str}). Our models integrate visual encoders (ResNet \cite{he2016deep} or ViT \cite{dosovitskiy2020image}) with an sEMG sequence predictor (LSTM \cite{hochreiter1997long} or SVR \cite{drucker1996support}). LSTM processes features sequentially, while for SVR, features are concatenated along the temporal direction (a parallel implementation). We have provided further modeling and implementation details in the following.

\subsubsection{Video2EMG Modeling Details}
For LSTM-based models, the frame-level features extracted by the visual encoders were processed by a single-layer LSTM configured with a hidden size of 256. Subsequently, the final hidden state was passed through fully connected layers for the final prediction. In the case of kernel-based SVR machines, features from 16 frames were concatenated into high-dimensional vectors and were regressed to the space of the EMG signals utilizing SVRs with RBF kernels. The models were trained and evaluated on 18 muscles according to the collected sEMG signals. The training process utilized the mean squared error loss function alongside the Adam optimizer, which was configured with an initial learning rate of $1\times10^{-3}$. Furthermore, a batch size of 32 was employed. This process required approximately 6 hours on a single NVIDIA RTX4090 graphics processing unit. Beyond the vailla split, we also introduce two rigorous evaluation protocols: cross-view and cross-subject configurations. In total, the experiments for the Video2EMG task consumed approximately 72 hours of computation.

\subsubsection{Task Standardization and Performance Metrics}
Since the task is novel, we establish a standard practice for it and propose multiple suitable metrics to measure the performance on the Video2EMG task. In accordance with standard evaluation protocols in time-series waveform analysis, we adopt the Mean Absolute Error (MAE), Root Mean Square Error (RMSE), and Cross-Correlation Peak (CCP) as the primary evaluation metrics for this task. The mean absolute error (MAE) and the root mean square error (RMSE), which quantify the average error and the variability of the error, respectively, are reported. Furthermore, the cross-correlation peak (CCP) is utilized to quantify the morphological similarity between the predicted and the target waveforms. The CCP is formally defined as the maximum value of the normalized cross-correlation function:
\begin{equation}
\resizebox{0.9\columnwidth}{!}{
$CCP = \max_{m} \left( \frac{\sum_{n} [y(n) - \bar{y}] \cdot [\hat{y}(n-m) - \bar{\hat{y}}]}{\sqrt{\sum_{n} [y(n) - \bar{y}]^2 \cdot \sum_{n} [\hat{y}(n-m) - \bar{\hat{y}}]^2}} \right)$
}
\end{equation}
where $y(n)$ and $\hat{y}(n)$ represent the discrete target and predicted waveforms, respectively, $\bar{y}$ and $\bar{\hat{y}}$ denote their respective mean values; and $m$ indicates the time lag. The value of the CCP is bounded between $-1$ and $1$, with larger positive values indicating a stronger alignment of the respective waveforms. Specifically, a value of CCP approaching $1$ demonstrates the successful capture of the essential morphology of the target sequence by the computational model, even in the presence of temporal shifts.

\subsubsection{Results}

\begin{figure}[ht]
    \centering
    \includegraphics[width=\linewidth]{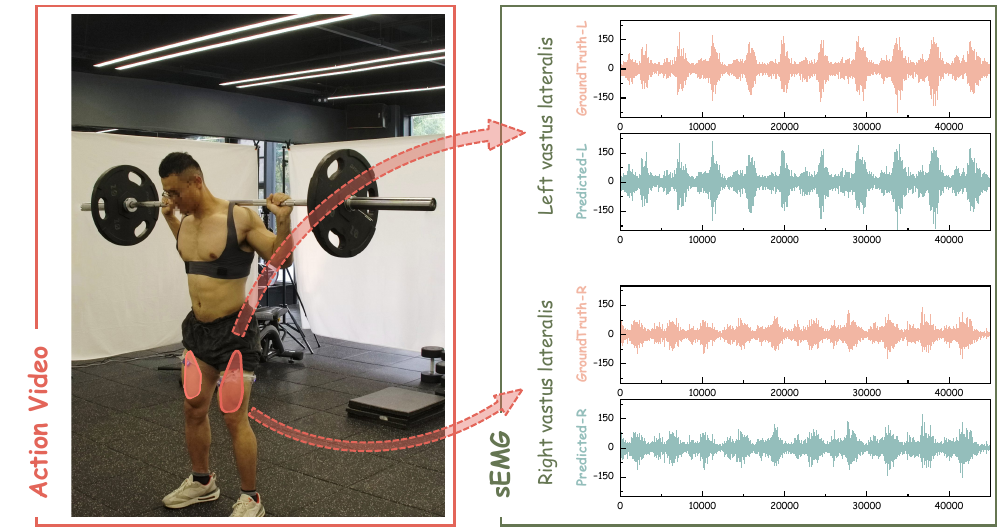}
    \caption{\textbf{Qualitative result of the Video2EMG task based on the ViT+SVR.}}
    \label{fig_exp_v2e_com}
\end{figure}

\noindent\textbf{Overall.}
Following 150 epochs under 3 evaluation protocols, the baselines for the Video2EMG task achieve promising performance on the vanilla split. As demonstrated in \autoref{table_exp_v2e}, the best configuration reaches an MAE of $0.1465$, an RMSE of $0.1882$, and a CCP of $0.4345$. We have shown qualitative results of the Video2EMG task in \autoref{fig_exp_v2e_com}. To provide a deeper understanding of the model's behavior, we present a detailed analysis of the network architectures and their robustness to distribution shifts below. \\

\noindent\textbf{Architectural Insights.}
Notably, the results in \autoref{table_exp_v2e} reveal only a marginal performance gap between the two visual backbones. This similarity is largely due to our specific experimental setup: by utilizing frozen, medium-scale models solely as feature extractors, both the ResNet-50 and ViT-S/16 architectures provide comparably mature and stable high-level semantic representations. Given identical training data and supervisory signals, the quality of the extracted features results in highly similar MAE outcomes.
In contrast, the choice of the regression head exerts a much more pronounced effect on the final predictions. Under the exact same visual representations, the SVR-based temporal modeling consistently outperforms the LSTM, achieving a relative MAE reduction of approximately $10\%$. This disparity clearly indicates that visual feature extraction is not the primary limiting factor; rather, the complex cross-modal mapping from visual features to continuous physiological signals remains the major bottleneck in the current pipeline.\\

\noindent\textbf{Robustness Analysis.}
To comprehensively measure the morphological alignment of the predicted EMG waveforms, we introduce the cross-correlation peak (CCP) as an intuitive evaluation metric. In practical EMG analysis, signals are typically nonstationary and noisy. Values above 0.7 generally indicate strong morphological similarity, 0.3 to 0.7 represent moderate similarity, and values below 0.3 suggest poor alignment or mismatched morphology.
As reported in \autoref{table_exp_v2e}, while the baseline methods achieve moderate waveform alignment on the vanilla split (CCP = 0.4345), their performance degrades noticeably under the disjoint cross-subject and cross-view protocols. This sharp decline reveals that the Video2EMG task on the MyoMechanix dataset is highly sensitive to individual physiological variations and viewpoint shifts.\\

\noindent\textbf{Summary.} These findings demonstrate that while estimating muscle-level feedback from videos is feasible, the task is highly challenging and far from being fully solved. The current baselines establish a promising foundation but leave considerable room (about 56\%, because CCP has an upper bound of 1.0) for future research to cover.
\section{Conclusion}
\label{Sec6-Con}

In this work, we introduced MyoMechanix, a biomechanically grounded multimodal ecosystem for skilled activity understanding and coaching, and established a compositional paradigm for action quality assessment that moves beyond purely visual, monolithic action modeling.
To support this direction, we presented MyoMechanix, a large-scale multimodal ecosystem for weight-loaded actions that synchronizes multiview RGB video, 3D pose, sEMG, and complementary physiological signals under strong internal–external alignment. Together with expert annotations, MyoMechanix provides a rich foundation for studying fine-grained, physically meaningful action understanding.
Building on this sensing framework, we developed the Fitness Knowledge Graph (FKG) to encode structured relationships among actions, phases, execution steps, errors, and corrective feedback. This enables a more interpretable and semantically grounded formulation of action quality assessment. We further proposed CUBIST, a compositional reasoning framework that decomposes actions into structured units, analyzes their quality, and recomposes them for holistic scoring, error attribution, and feedback generation. Across the newly established MyoMechanix-AQA, MyoMechanix-VideoQA, and MyoMechanix-Video2EMG benchmarks, extensive experiments demonstrate that multimodal sensing, structured representations, and compositional reasoning jointly improve performance, interpretability, and diagnostic capability, with our CUBIST model achieving state-of-the-art results.
Our findings highlight three broader implications. First, effective action quality assessment benefits substantially from integrating latent biomechanical signals with visible motion cues. Second, language-grounded reasoning over structured action knowledge can enhance fine-grained understanding in vision–language models. Third, the promising results on Video2EMG suggest a path toward estimating internal muscle dynamics from accessible visual observations, potentially reducing dependence on specialized sensing hardware. 
Despite these advances, MyoMechanix remains a challenging benchmark, and several directions merit further exploration, including more robust cross-subject generalization, richer causal modeling of biomechanics, broader action coverage, and stronger integration with embodied and interactive systems. Overall, this work establishes a new foundation for action understanding that is multimodal, biomechanically informed, and compositionally interpretable, opening opportunities for next-generation Physical AI in fitness, rehabilitation, healthcare, and representation learning.

\vspace{1em}
\noindent\textbf{Acknowledgement}
This work is supported by the Suzhou Basic Scientific Research Project under the Grant SSD2024013, the Doctoral Student Program of the Young S\&T Talents Cultivation Project, CAST. 

\vspace{1em}
\noindent\textbf{Data Availability}
The full dataset and codebase are released under the CC BY-NC-SA 4.0 license and are publicly available at the \href{https://github.com/HaoYin116/MyoMechanix}{Github}.

\bibliographystyle{unsrt}
\bibliography{sn-bibliography}

\end{document}